\pdfoutput=1
\documentclass[12pt,onecolumn]{article}
\ifdefined\pdfobjcompresslevel
\fi
\usepackage[utf8]{inputenc}
\usepackage[left=1.5cm, right=1.5cm, top=1.5cm, bottom=1.5cm]{geometry}
\usepackage{amsmath,amsfonts,amssymb}
\usepackage{graphicx}
\usepackage{float}
\usepackage{placeins}
\usepackage{afterpage}
\usepackage[numbers,round,sort&compress]{natbib}

\let\citenospace\cite
\renewcommand{\cite}[1]{\unskip~\citenospace{#1}}
\usepackage{xurl}
\usepackage{setspace}
\usepackage{times}
\usepackage{titlesec}
\usepackage[colorlinks=true,linkcolor=blue,citecolor=blue,urlcolor=blue,hypertexnames=false]{hyperref}
\usepackage{caption}
\usepackage[textsize=small]{todonotes}
\usepackage{tikz}
\usepackage{fontawesome5}
\renewcommand{\figurename}{Fig.}
\DeclareCaptionLabelSeparator{tabletitlebar}{\enspace\textbar\enspace}
\usepackage{tabularx}
\usepackage{booktabs}
\usepackage[table]{xcolor}
\definecolor{figurerefblue}{RGB}{26,13,172}
\colorlet{replacedred}{red!55}
\usepackage{longtable}
\usepackage{makecell}

\AtBeginDocument{%
  \hypersetup{%
    pdfborder={0 0 1},%
    pdfborderstyle={/S/U/W 0.8},%
    linkcolor=figurerefblue,%
    citecolor=figurerefblue,%
    linkbordercolor=figurerefblue,%
    citebordercolor=figurerefblue,%
    urlbordercolor=blue%
  }%
}

\definecolor{dtwocline}{HTML}{3A4046}
\newcommand{\dtwocmark}[1][1.2em]{\raisebox{-0.24em}{\resizebox{!}{#1}{%
  \begin{tikzpicture}[x=1pt,y=-1pt,line join=round,line width=6.618pt,draw=dtwocline]
    \filldraw[fill={rgb,255:red,125;green,116;blue,197}] (93.220,44.814)--(39.697,137.519)--(146.744,137.519)--cycle;
    \filldraw[fill={rgb,255:red,233;green,193;blue,202}] (93.220,44.814)--(39.697,137.519)--(6.618,56.618)--cycle;
    \filldraw[fill={rgb,255:red,242;green,223;blue,183}] (93.220,44.814)--(146.744,137.519)--(179.823,56.618)--cycle;
    \filldraw[fill={rgb,255:red,161;green,211;blue,207}] (39.697,137.519)--(146.744,137.519)--(93.220,206.618)--cycle;
    \filldraw[fill={rgb,255:red,228;green,231;blue,234}] (93.220,44.814)--(93.220,6.618)--(6.618,56.618)--cycle;
    \filldraw[fill={rgb,255:red,244;green,245;blue,246}] (39.697,137.519)--(6.618,156.618)--(93.220,206.618)--cycle;
    \filldraw[fill={rgb,255:red,228;green,231;blue,234}] (146.744,137.519)--(179.823,156.618)--(179.823,56.618)--cycle;
    \filldraw[fill={rgb,255:red,244;green,245;blue,246}] (93.220,44.814)--(93.220,6.618)--(179.823,56.618)--cycle;
    \filldraw[fill={rgb,255:red,228;green,231;blue,234}] (39.697,137.519)--(6.618,156.618)--(6.618,56.618)--cycle;
    \filldraw[fill={rgb,255:red,244;green,245;blue,246}] (146.744,137.519)--(179.823,156.618)--(93.220,206.618)--cycle;
  \end{tikzpicture}}}}

\newcolumntype{L}[1]{>{\raggedright\arraybackslash}p{#1}}
\newcolumntype{C}[1]{>{\centering\arraybackslash}p{#1}}
\newcommand{\figref}[2][]{\mbox{\hyperref[#2]{\textbf{Fig.~\ref*{#2}#1}}}}
\newcommand{\Figref}[2][]{\mbox{\hyperref[#2]{\textbf{Figure~\ref*{#2}#1}}}}
\newcommand{\suppfigref}[2][]{\mbox{\hyperref[#2]{\textbf{fig.~\ref*{#2}#1}}}}
\newcommand{\tabref}[2][]{\mbox{\hyperref[#2]{\textbf{Table~\ref*{#2}#1}}}}
\newcommand{\supptabref}[2][]{\mbox{\hyperref[#2]{\textbf{table~\ref*{#2}#1}}}}
\newcommand{\supptabrefs}[2]{\mbox{\hyperref[#1]{\textbf{tables~\ref*{#1} and~}}\hyperref[#2]{\textbf{\ref*{#2}}}}}
\newcommand{\methodsref}[1]{\hyperref[#1]{materials and methods}}
\newcommand{\methodsanchor}[1]{\phantomsection\label{#1}}

\newcommand{\legendnextpage}[2][]{%
  \afterpage{%
    \vspace*{-0.8\topskip}%
    {\setlength{\parskip}{0pt}\captionsetup{type=figure,hypcap=false,skip=0pt,#1}\caption{#2}}%
    \vspace{0em}%
    \noindent\rule{\textwidth}{0.4pt}%
    \vspace{0em}}}

\makeatletter
\newcount\supp@mainrefs
\newcount\supp@allrefs
\newif\ifsupp@started
\def\supp@strip#1{\zap@space#1 \@empty}
\renewcommand{\citation}[1]{%
  \@for\supp@key:=#1\do{%
    \edef\supp@key{\expandafter\supp@strip\expandafter{\supp@key}}%
    \@ifundefined{supp@cited@\supp@key}{%
      \global\@namedef{supp@cited@\supp@key}{}%
      \global\advance\supp@allrefs\@ne
      \ifsupp@started\else\global\advance\supp@mainrefs\@ne\fi}{}}}
\newcommand{\supp@begin}{\global\supp@startedtrue}
\newcommand{\supp@counts}[2]{\gdef\suppfigures{#1}\gdef\supptables{#2}}
\newcommand{\suppfigures}{0}
\newcommand{\supptables}{0}
\newcommand{\supprefrange}{\the\numexpr\supp@mainrefs+1\relax--\the\supp@allrefs}
\newcommand{\supplementstart}{\immediate\write\@auxout{\string\supp@begin}}
\newcommand{\supplementend}{\immediate\write\@auxout{\string\supp@counts{\arabic{figure}}{\arabic{table}}}}
\makeatother

\titleformat*{\section}{\Large\bfseries}
\titlespacing*{\section}{0pt}{3em}{0.2em}
\titleformat*{\subsection}{\large\bfseries}
\titlespacing*{\subsection}{0pt}{1.5em}{-1.0em}

\begin{document}
\fontsize{13}{15.6}\selectfont
\setstretch{1.0}
\thispagestyle{plain}

\newcommand{\papertitle}{Auditable Clinical Support with Decision Agents}
\newcommand{\paperauthors}{%
Tiancheng Bao\textsuperscript{1,$\dagger$},
Riga Wu\textsuperscript{1,$\dagger$},
Jan Wienhold\textsuperscript{2,3,$\dagger$},
Sebastian Billig\textsuperscript{2,3},
Patrick Winnersbach\textsuperscript{2,3},
Hsuan-Chih Chen\textsuperscript{1},
Akis Giannoukos\textsuperscript{1},
Jiayang Li\textsuperscript{1},
Li Shen\textsuperscript{1,4,5},
Gary Weissman\textsuperscript{4,6},
Eduardo~J.~Mortani~Barbosa~Jr.\textsuperscript{1},
Dyke Ferber\textsuperscript{7,8},
Jakob Nikolas Kather\textsuperscript{7,8,9},
Christos Davatzikos\textsuperscript{1,5},
Daniel Truhn\textsuperscript{10,$\ddagger$},
\& Tianyu Han\textsuperscript{1,5,*,$\ddagger$}}
\newcommand{\correspondingauthor}{Tianyu Han (tianyu.han@pennmedicine.upenn.edu)}

\noindent{\fontsize{20}{24}\selectfont\sffamily\bfseries\raggedright \papertitle\par}
\vspace{1.1em}
{\setstretch{1.15}\fontsize{13.5}{16.5}\selectfont\raggedright
\noindent
\paperauthors
\par}
\vspace{1.1em}

{\setlength{\parskip}{0pt}%
\fontsize{12}{14.2}\selectfont
\itshape
\raggedright
\noindent
\textsuperscript{1}Department of Radiology, Perelman School of Medicine, University of Pennsylvania, Philadelphia, PA, USA.\\[0.05em]
\textsuperscript{2}Department of Anesthesiology, Medical Faculty, RWTH Aachen University, Aachen, Germany.\\[0.05em]
\textsuperscript{3}Department of Intensive Care and Intermediate Care, Medical Faculty, RWTH Aachen University, Aachen, Germany.\\[0.05em]
\textsuperscript{4}Department of Biostatistics, Epidemiology and Informatics, Perelman School of Medicine, University of Pennsylvania, Philadelphia, PA, USA.\\[0.05em]
\textsuperscript{5}Center for Artificial Intelligence and Data Science for Integrated Diagnostics (AI2D), Perelman School of Medicine, University of Pennsylvania, Philadelphia, PA, USA.\\[0.05em]
\textsuperscript{6}Pulmonary, Allergy, and Critical Care Division, Department of Medicine, Perelman School of Medicine, University of Pennsylvania, Philadelphia, PA, USA.\\[0.05em]
\textsuperscript{7}Else Kr\"oner Fresenius Center for Digital Health, Faculty of Medicine and University Hospital Carl Gustav Carus, TUD Dresden University of Technology, Dresden, Germany.\\[0.05em]
\textsuperscript{8}Department of Medical Oncology, National Center for Tumor Diseases (NCT), University Hospital Heidelberg, Heidelberg, Germany.\\[0.05em]
\textsuperscript{9}Institute for Medical AI (IMAI), Heidelberg University Hospital and Medical Faculty of Heidelberg University, Heidelberg, Germany.\\[0.05em]
\textsuperscript{10}Department of Diagnostic and Interventional Radiology, University Hospital RWTH Aachen, Aachen, Germany.\par

\vspace{0.8em}
\noindent\textsuperscript{$\dagger$}These authors contributed equally to this work. \textsuperscript{$\ddagger$}These authors contributed equally to this work.\\[0.4em]
\noindent\textsuperscript{*}\textbf{Corresponding author:} \correspondingauthor
\par
}

\vspace{1.75em}
{\centering\fontsize{12.5}{15}\selectfont
\href{https://github.com/peterhan91/Decision2Care}{\textcolor{black}{\faGithub}\hspace{0.45em}Open-source code repository}%
\hspace{2.8em}%
\href{https://peterhan91.github.io/Decision2Care/}{\dtwocmark\hspace{0.45em}Project website}
\par}
\vspace{1.9em}
\enlargethispage{\baselineskip}
{\fontsize{12}{14}\selectfont\bfseries\noindent
Emergency care requires auditable decisions, but a language-model agent just outputs text.
We introduce Decision2Care, a decision agent that generates a probability for every option at each step.
It diagnosed 546 emergency admissions across 12 diseases with 86.5\% accuracy in a median of 2.2~s; language-model agents reached 86.1 to 91.7\% but took more than 90~s.
Its drug orders were judged guideline-adherent more often than documented first-day care and with about half as many records, it predicted death as well as three ICU clinicians.
In the four cases where it was confident but wrong, we traced each decision to one imaging report.
Our study suggests that an agent's work-up can be reviewed like a clinician's, motivating prospective trials.
\par}

\newpage
\begingroup
\setlength{\parskip}{0.75em}
\renewcommand{\topfraction}{0.95}
\renewcommand{\bottomfraction}{0.95}
\renewcommand{\textfraction}{0.05}
\renewcommand{\floatpagefraction}{0.85}
\titlespacing*{\section}{0pt}{1.5em}{0.2em}
\titlespacing*{\subsection}{0pt}{1em}{0pt}
\section*{Introduction}

At morbidity and mortality conferences, clinicians review missed diagnoses by asking what was known at each step and why a test was or was not done\cite{orlander2002morbidity,pierluissi2003discussion}.
A patient with right upper abdominal pain shortly after cholecystectomy may first be investigated for an abdominal complication. 
The cause may instead be pulmonary embolism (PE), one of the most commonly missed diagnoses\cite{schiff2009diagnostic}, which can be overlooked if the investigation stops at the first plausible diagnosis\cite {graber2005diagnostic}.
To review such a case, clinicians must know which options were considered and which results changed the assessment.

Evaluations of large language models (LLMs) in medicine have moved from clinical vignettes\cite{han2024comparative} to diagnostic conversations\cite{tu2025conversational}, work-ups on patient records\cite{hager2024cdm,ferber2026mira,zhang2026onprem} and actions in the electronic health record\cite{jiang2025medagentbench}.
In a work-up scenario, an LLM agent can request the patient's examination findings, laboratory results, and imaging before reaching a diagnosis and treatment plan.
Ferber et al.\cite{ferber2026mira} built one such agent, MIRA (Medical Intelligence for Reasoning and Action), which questions a simulated patient and orders tests and treatments in a virtual electronic health record.
MIRA was at least as good as a physician at diagnosing emergency cases given the same information.
Zhang et al.\cite{zhang2026onprem} adapted it to run on a hospital's own servers, and we refer to their agent as MIRA-v2.
In sequential diagnosis, where the next question or test is chosen at each step, an LLM system that simulates a panel of physicians was more accurate than generalist physicians on published case records and on emergency department encounters from MIMIC-IV\cite{nori2026sequential}.
An LLM also outperformed physician baselines across clinical reasoning tasks, including second opinions on 76 emergency room cases at Beth Israel Deaconess Medical Center\cite{brodeur2026reasoning}.
If such agents are to order tests in a real clinical setting, their workups must be auditable.

Current LLM agents\cite{hager2024cdm,ferber2026mira,zhang2026onprem} produce their reasoning and tool calls as text\cite{yao2023react}.
The transcript shows what the agent did, but its explanation may omit the findings that led to a particular decision\cite{turpin2023unfaithful}.
Confidence measures are also costly.  
In MIRA-v2\cite{zhang2026onprem}, the best-performing measure requires five work-ups of the same case.
DeepRare \cite{zhao2026deeprare} links its diagnostic reasoning to medical sources, but source attribution alone does not establish how individual patient findings affect a decision.
Earlier reinforcement learning agents selected from a pre-defined set of actions for symptom inquiry\cite{kao2018symptom} and sepsis treatment\cite{komorowski2018clinician}.
Each of those was trained on a single task using structured input.
None of these agents record, at each step, the options they weighed in a form that can be quantified and re-run.

Here, we present Decision2Care, a clinical agent for the patient in the emergency department (ED) and intensive care unit (ICU) that generates no decision as text.
It does the work-up via typed decisions (e.g.\ choosing a diagnosis from a list, deciding whether to order a test).
In each case, the decision model returns a probability for every option. 
A controller then applies fixed rules to those probabilities: It executes the most probable action, and orders a test when its probability reaches 0.5.
We call an agent constructed this way a \textbf{decision agent}.
The same architecture is used to select treatment and to estimate mortality risk, and the controller can use different decision models.
Decision2Care records the probabilities of all the options at each step, so a reviewer can see which options were considered. 
The work-up takes a few seconds, so we can remove or replace a result, re-run the work-up and check which result changed the decision.

To test whether a decision agent can still do the work of an LLM agent, we ran Decision2Care on 546 emergency admissions to Beth Israel Deaconess Medical Center spanning 12 diseases, taken from MIMIC-IV.
Decision2Care reached a mean diagnostic accuracy of 86.5\%, within the range of MIRA-v2 with the three LLMs, while a work-up took a median of 2.2~s instead of 92--100~s and cost less than a fifteenth as much.
Its work-ups could be followed step by step. 
For example, in the patient described above, the probability of PE in the differential diagnosis was 0.45 after abdominal computed tomography (CT) and 0.96 after chest CT.
When we left out the lipase and amylase results and repeated the diagnosis, accuracy in pancreatitis admissions decreased by 54 percentage points, whereas removing the imaging reports or the routine blood tests left it unchanged.
The confidence score derived from a single work-up distinguished correct from incorrect diagnoses about as well as the agreement among five repeated MIRA-v2 work-ups, and the 49.3\% of admissions in which Decision2Care was most confident were diagnosed with 98.5\% accuracy.
The drug orders of Decision2Care were judged guideline-adherent more often than those of MIRA-v2 and the documented first-day care.
Using about half of the available records, it predicted death in emergency patients as well as three ICU clinicians who had all of them.
In the ICU, its predictions from 12~h of records were as good as or better than three severity scores computed at 24~h.
In records from Stanford Health Care, the prediction reached areas under the receiver operating characteristic curve (AUROCs) of 0.81 in 555 patients with PE and 0.77 in 78 with pneumonia, similar to those at Beth Israel Deaconess Medical Center.
These results indicate that a decision agent can perform a clinical work-up, including treatment and prognosis, without generating text, and that each step is auditable.

\FloatBarrier
\section*{Results}

\subsection*{Decision2Care acts on typed decisions}

\afterpage{%
\begin{figure}[H]
    \centering
    \includegraphics[width=\textwidth]{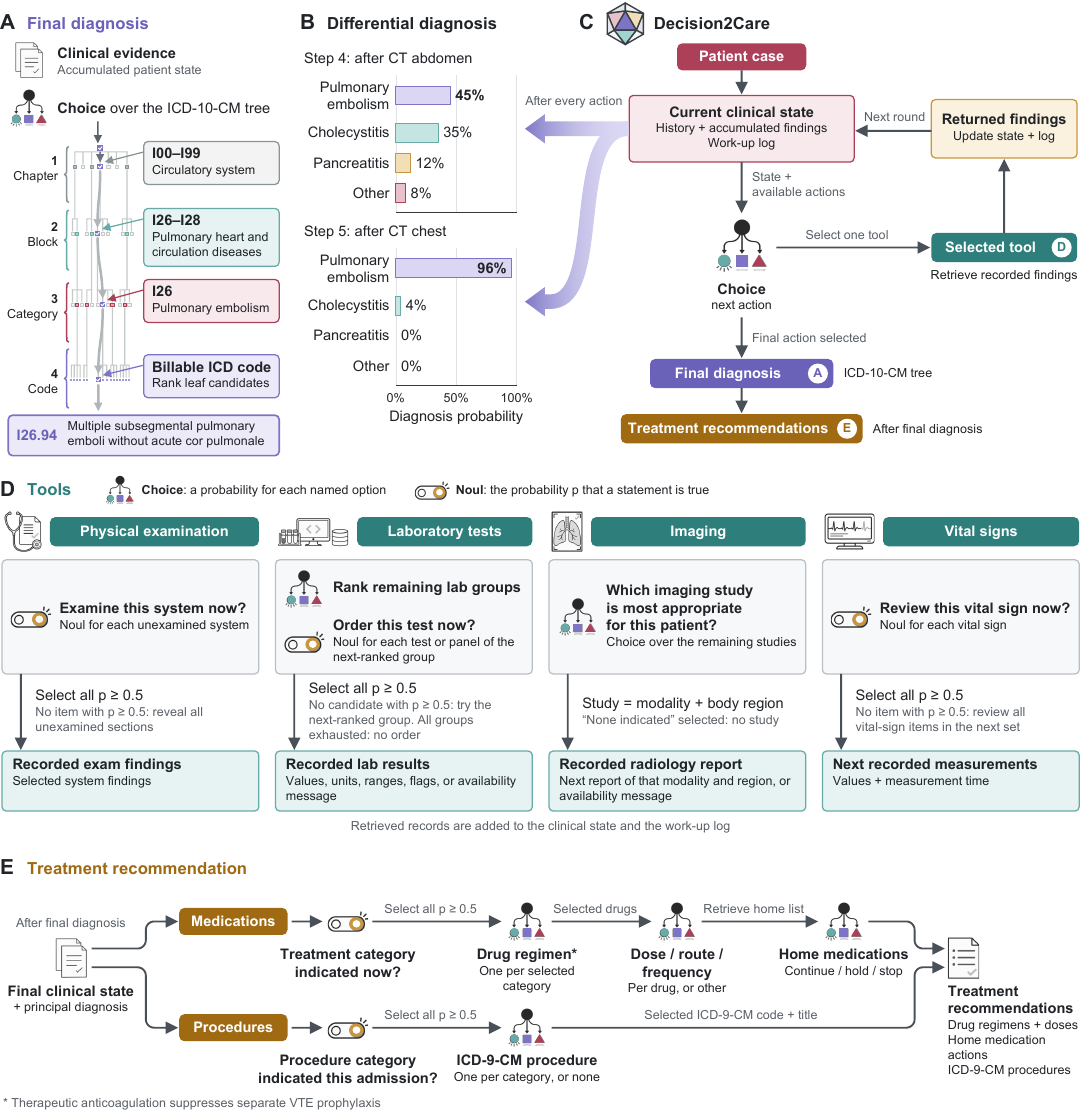}
\end{figure}
\legendnextpage[font=small]{\textbf{Overview of Decision2Care.}
    Decision2Care puts 2 kinds of typed questions into its decision model: for a Choice, the model returns a probability for each named option, and for a Noul, a yes/no question, it returns the probability that a statement is true.
    With these questions, Decision2Care codes the final diagnosis (\textbf{A}), follows the differential diagnosis (\textbf{B}), runs the work-up (\textbf{C}) with tools (\textbf{D}) and recommends treatment (\textbf{E}).
    \textbf{A}, The final diagnosis.
    At the end of the work-up, the decision model answers a Choice at each level of the ICD-10-CM tree and then one among the billable codes, which here gives I26.94.
    \textbf{B}, The differential diagnosis.
    After every action, a Choice over the diseases of the evaluation and `other' gives the probability of each diagnosis without changing the work-up.
    In the patient shown, the probability of pulmonary embolism was 0.45 after the abdominal CT and 0.96 after the chest CT.
    \textbf{C}, The work-up loop.
    In each round, the decision model reads the current clinical state and answers a Choice over the available actions.
    The controller then runs the selected tool, which retrieves the recorded findings and adds them to the state.
    The loop continues until the Choice selects the final diagnosis.
    \textbf{D}, Four of the 5 tools (the review of the home medication list is not shown).
    For the physical examination and the vital signs, a Noul is asked for each item, and the tool retrieves every item with a probability of at least 0.5.
    For laboratory tests, a Choice first ranks the groups of tests and a Noul then selects specific tests within a group.
    For imaging, a Choice selects one study, defined by its modality and body region, or the option that no further imaging is indicated.
    \textbf{E}, Treatment recommendation.
    After the final diagnosis, a Noul for each medication and procedure category selects the indicated options.
    Choices then select the drug regimen with its dose, route and frequency, the action on each home medication and the ICD-9-CM procedure.
    \label{fig:overview}}%
}

Decision2Care divides a clinical work-up between a decision model and a deterministic controller (\figref{fig:overview}).
The decision model reads the current clinical state and answers typed questions with probabilities rather than text.
We call these answers typed decisions.
The controller turns them into actions using fixed rules.
One kind of typed question, a Choice, asks for a probability for each of a set of named options, such as the codes from which the final diagnosis is selected (\figref[A]{fig:overview}) or the diseases of the differential diagnosis (\figref[B]{fig:overview}).
The other kind is a yes/no question, called a Noul.
It asks for the probability that a statement is true, for example that a given test should be ordered now (\methodsref{meth:architecture}).
Unless stated otherwise, the decision model is Jev\cite{almeida2026jev}, developed by TypeSafe AI.
It can be replaced by another decision model without changing the controller or the questions (\methodsref{meth:models}).

A diagnostic work-up proceeds in rounds (\figref[C]{fig:overview}).
In each round, the model is given a Choice over the remaining actions, and the controller executes the most probable one, either a tool or the final diagnosis.
Each tool asks further questions to decide what to retrieve.
The physical examination, for example, asks a Noul for each body system and adds the recorded findings of every system that reaches a probability of 0.5 to the clinical state (\figref[D]{fig:overview}).
After every action, the model can also be asked for the differential diagnosis, which is recorded without changing the course of the work-up.
The final diagnosis ends the rounds and is coded by Choices that descend the ICD-10-CM tree from chapter to billable code.
The same 2 kinds of questions then select the treatment (\figref[E]{fig:overview}), and in the mortality cohorts Choices select the records to retrieve and then give the probability of death (\methodsref{meth:prognosis}).
Decision2Care logs each question, the clinical state used to answer it, the probabilities returned, and the rule applied.
A reviewer can then go through the work-up as a sequence of typed decisions and re-ask any question using the state that was recorded at that point.

\FloatBarrier
\subsection*{Diagnostic accuracy and cost per case compared with MIRA-v2}

We first asked whether typed decisions alone are enough to complete a diagnostic work-up, and how the agent's accuracy, time and cost compare with those of an LLM agent.
The comparator was MIRA-v2\cite{zhang2026onprem}, run with OpenAI's GPT-OSS-120B\cite{agarwal2025gpt}, Z.ai's GLM-5 (754B)\cite{zeng2026glm} and Anthropic's Claude Sonnet 5.
Both systems worked up 546 MIMIC-IV admissions spanning 12 diseases, 437 from the MIRA-v2 benchmark and 109 with 5 additional diseases.
Admissions whose history disclosed the diagnosis had been excluded (\tabref{tab:diagnostic-efficiency}; \methodsref{meth:cases}).
For this comparison, we ran Decision2Care with Jev and each MIRA-v2 configuration 5 times per case.
We scored every answer against the diagnosis recorded for the admission.

\afterpage{\afterpage{%
\begin{table}[!b]
    \centering
    \captionsetup{position=top,font={small,sf,bf},labelfont=bf,skip=8pt}
    \caption{Comparison of computational efficiency and disease-specific diagnostic accuracy between Decision2Care and MIRA-v2}
    \label{tab:diagnostic-efficiency}
    \begingroup
\definecolor{jevfill}{HTML}{E7F0EC}
\definecolor{jevhead}{HTML}{C8DDD4}
\newcommand{\J}[1]{\cellcolor{jevfill}#1}
\newcolumntype{Y}{>{\centering\arraybackslash}X}
\begin{minipage}{\linewidth}
\sffamily\raggedright
\setlength{\parskip}{0pt}
\setlength{\parindent}{0pt}
\fontsize{8.6}{10.4}\selectfont
\setlength{\tabcolsep}{4pt}
\renewcommand{\arraystretch}{1.48}
\setlength{\aboverulesep}{0pt}
\setlength{\belowrulesep}{0pt}
\begin{tabularx}{\linewidth}{>{\raggedright\arraybackslash}p{50mm}>{\centering\arraybackslash}p{7mm}YYYY}
\toprule[0.8pt]
\textbf{Measure} & \textbf{\textit{n}} & \cellcolor{jevhead}\textbf{Decision2Care} & \multicolumn{3}{c}{\textbf{MIRA-v2 workflow}} \\
 & & \cellcolor{jevhead}\textbf{This work} & \textbf{GPT-OSS-120B} & \textbf{GLM-5} & \textbf{Claude Sonnet 5} \\
\midrule[0.6pt]
\multicolumn{6}{l}{\rule{0pt}{15pt}\textbf{Computational efficiency}} \\[2pt]
Median time per work-up, s* & \textemdash & \J{\textbf{2.2\,{\scriptsize (1.7--2.7)}}} & 92.7\,{\scriptsize (74.3--116.6)} & 92.4\,{\scriptsize (75.5--121.8)} & 100.1\,{\scriptsize (87.5--119.1)} \\
Cost per case, 1 work-up, US\$\textdagger{} & 546 & \J{\textbf{0.0024}} & 0.0376 & 0.1266 & 0.1460 \\
Total cost, 1 work-up, US\$\textdagger{} & 546 & \J{\textbf{1.31}} & 20.5 & 69.1 & 79.7 \\
Total cost, 5 work-ups, US\$\textdaggerdbl{} & 546 & \J{\textemdash} & 102.6 & 345.5 & 398.5 \\
\midrule
\multicolumn{6}{l}{\rule{0pt}{15pt}\textbf{Diagnostic accuracy, \%}} \\[2pt]
Appendicitis & 148 & \J{99.9\,{\scriptsize (99.3--100.0)}} & 98.6\,{\scriptsize (98.0--99.3)} & 98.5\,{\scriptsize (98.0--98.6)} & 99.2\,{\scriptsize (98.6--99.3)} \\
Cholecystitis & 129 & \J{77.7\,{\scriptsize (77.5--78.3)}} & 81.2\,{\scriptsize (79.8--82.9)} & 85.7\,{\scriptsize (84.5--86.8)} & 91.0\,{\scriptsize (88.4--93.0)} \\
Diverticulitis & 54 & \J{88.1\,{\scriptsize (85.2--90.7)}} & 90.7\,{\scriptsize (88.9--92.6)} & 93.7\,{\scriptsize (92.6--94.4)} & 95.6\,{\scriptsize (94.4--96.3)} \\
Pancreatitis & 52 & \J{91.5\,{\scriptsize (88.5--92.3)}} & 94.2\,{\scriptsize (90.4--98.1)} & 98.5\,{\scriptsize (98.1--100.0)} & 98.1\,{\scriptsize (96.2--100.0)} \\
Pneumonia & 21 & \J{45.7\,{\scriptsize (42.9--47.6)}} & 37.1\,{\scriptsize (33.3--42.9)} & 52.4\,{\scriptsize (47.6--61.9)} & 53.3\,{\scriptsize (52.4--57.1)} \\
Pulmonary embolism & 15 & \J{92.0\,{\scriptsize (86.7--93.3)}} & 70.7\,{\scriptsize (60.0--80.0)} & 88.0\,{\scriptsize (80.0--93.3)} & 84.0\,{\scriptsize (80.0--93.3)} \\
UTI & 18 & \J{82.2\,{\scriptsize (77.8--83.3)}} & 60.0\,{\scriptsize (50.0--72.2)} & 78.9\,{\scriptsize (72.2--83.3)} & 82.2\,{\scriptsize (72.2--94.4)} \\
Bowel obstruction & 67 & \J{91.6\,{\scriptsize (91.0--92.5)}} & 90.4\,{\scriptsize (88.1--94.0)} & 94.9\,{\scriptsize (92.5--97.0)} & 94.3\,{\scriptsize (92.5--97.0)} \\
Heart failure & 13 & \J{64.6\,{\scriptsize (61.5--69.2)}} & 81.5\,{\scriptsize (76.9--84.6)} & 78.5\,{\scriptsize (69.2--84.6)} & 81.5\,{\scriptsize (76.9--84.6)} \\
Subarachnoid haemorrhage & 7 & \J{42.9\,{\scriptsize (42.9--42.9)}} & 22.9\,{\scriptsize (14.3--42.9)} & 34.3\,{\scriptsize (28.6--42.9)} & 40.0\,{\scriptsize (28.6--42.9)} \\
Myocardial infarction & 13 & \J{69.2\,{\scriptsize (69.2--69.2)}} & 86.2\,{\scriptsize (84.6--92.3)} & 83.1\,{\scriptsize (76.9--92.3)} & 86.2\,{\scriptsize (84.6--92.3)} \\
Meningitis & 9 & \J{100.0\,{\scriptsize (100.0--100.0)}} & 88.9\,{\scriptsize (77.8--100.0)} & 80.0\,{\scriptsize (66.7--88.9)} & 82.2\,{\scriptsize (66.7--100.0)} \\
\bottomrule[0.8pt]
\end{tabularx}\par
\vspace{5pt}
\fontsize{7.5}{9.5}\selectfont
Accuracy is the mean of 5 runs, with the minimum and maximum in parentheses.
*Median (interquartile range) per work-up; Decision2Care's excludes the differential and confidence readings made alongside it.
\textdagger{}Application programming interface list prices, 1 work-up per admission; MIRA-v2's exclude its judge, Decision2Care's include the confidence reading.
\textdaggerdbl{}The 5 runs that MIRA-v2's confidence signal requires.
Bold, the lowest time and cost; \textit{n}, admissions per disease; UTI, urinary tract infection; dash, not applicable (Decision2Care's confidence needs 1 work-up).\par
\end{minipage}
\endgroup

\end{table}}}

Mean diagnostic accuracy was 86.5\% for Decision2Care (95\% confidence interval [CI] 83.6--89.2\%), within the range of MIRA-v2 with GPT-OSS-120B, GLM-5 and Claude Sonnet 5 (86.1\%, 89.9\% and 91.7\%, respectively).
With a case counted as correct when at least 3 of its 5 runs were correct\cite{zhang2026onprem}, accuracy did not differ between Decision2Care and GPT-OSS-120B (two-sided exact McNemar test, $p = 0.78$) and was lower for Decision2Care than for GLM-5 and Claude Sonnet 5 ($p = 0.023$ and $p < 0.001$).
More than half of the difference in mean accuracy from these 2 configurations came from a single disease, cholecystitis (77.7\%, compared with 81.2--91.0\% for the 3 comparators).
Most errors in cholecystitis were coded as gallstones without cholecystitis (103 of the 144 incorrect answers across 5 runs; \supptabref{tab:supp-cholecystitis-errors}).
When we counted a match among the first 5 entries of the final differential, Decision2Care's accuracy was 85.7--100\% in 11 of the 12 diseases (\suppfigref[, A to~L]{fig:supp-ddx-topn}).
In the other 11 diseases, Decision2Care's accuracy did not differ from any comparator after correction for multiple comparisons, and it was numerically higher than all 3 comparators in appendicitis, PE, subarachnoid haemorrhage (SAH), and meningitis (\tabref{tab:diagnostic-efficiency}).
In the MIRA study, 4 board-certified physicians worked up a random sample of 290 admissions of the MIRA-v2 benchmark and diagnosed 77.6\% of them correctly (225 of 290; 95\% CI 72.4--82.0\%)\cite{ferber2026mira}.
Weighted to the same mix of diseases, Decision2Care's accuracy was 84.6\% (95\% CI 80.2--88.7\%; \supptabref{tab:supp-physicians}; \methodsref{meth:physicians}).

Accuracy varied little between runs, from 85.9\% to 87.0\% for Decision2Care, with a standard deviation of 0.4 percentage points, compared with 0.4--0.7 points for the comparators.
Accuracy also changed little when the options were given in another order, the questions were reworded or the record was presented in another format ($-2.1$ to $+1.3$ percentage points in 14 variants, 1 of which differed after correction for multiple comparisons; \supptabref{tab:supp-robustness-variants}; \methodsref{meth:variants}).
Similar accuracies were obtained when 2 other decision models drove the same controller, tools and typed questions in place of Jev, 85.9\% with PPLX decider v1.2\cite{sileo-2024-tasksource,perplexity2026decider} and 87.2\% with Qwen3.8-Flash-Next\cite{nvidia2026qwen}, a general LLM accessed through its candidate answer-token probabilities (\supptabref{tab:supp-decision-models}; \methodsref{meth:models}).
The paired differences from Jev were $-0.6$ percentage points for PPLX (95\% CI $-2.9$ to $+1.8$) and $+0.7$ for Qwen ($-1.7$ to $+3.1$).
In cholecystitis, the 2 models reached 80.6\% and 87.8\%, suggesting that the shortfall in this disease is not inherent to typed decisions.

Median time per case was 2.2 s for Decision2Care (interquartile range 1.7--2.7 s), compared with 92.7 s, 92.4 s and 100.1 s for MIRA-v2 with GPT-OSS-120B, GLM-5 and Claude Sonnet 5, respectively (\tabref{tab:diagnostic-efficiency}).
Jev answered a mean of 233 typed questions in 14 requests per case.
Working up all 546 admissions cost US\$1.31 for Decision2Care at list prices, compared with US\$20.5, US\$69.1 and US\$79.7 for 1 run of MIRA-v2 with GPT-OSS-120B, GLM-5 and Claude Sonnet 5 (\tabref{tab:diagnostic-efficiency}).
A Decision2Care work-up was thus more than 40 times faster than 1 run of MIRA-v2, at a cost 16 to 61 times lower.

Decision2Care work-ups followed a disease-specific path and led to the test on which the disease's guideline bases the diagnosis (\suppfigref[, A to~L]{fig:supp-workup-paths}).
The test was CT for appendicitis\cite{disaverio2020appendicitis}, diverticulitis\cite{hall2020diverticulitis}, bowel obstruction\cite{amara2021sbo} and PE\cite{konstantinides2020pe}, and head CT for SAH\cite{nice2022ng228}.
It was ultrasound for cholecystitis\cite{pisano2020cholecystitis}, urine tests for urinary tract infection (UTI)\cite{bonkat2024eau}, microbiology with cerebrospinal fluid (CSF) for meningitis\cite{vandebeek2016meningitis}, and laboratory tests for pancreatitis\cite{leppaniemi2019pancreatitis}, myocardial infarction (MI)\cite{nice2010cg95} and heart failure\cite{nice2014cg187} (\supptabref{tab:supp-guideline-tests}).
In 11 of the 12 diseases, the most frequent path ran from the physical examination to this test, and 49--99\% of a disease's work-ups followed it in order.
The guideline's test itself was obtained in 71--99\% of case-runs in 9 of these 11 diseases (\supptabref{tab:supp-decisive-steps}).
Most work-ups also began with the physical examination (95\% of case-runs) and ended with the agent choosing the final diagnosis, rather than running out of options (97\%; the tests requested are compared with those on record in \supptabref{tab:supp-test-selection}).
In short, typed decisions alone were enough to carry out the diagnostic work-up, at a fraction of the time and cost of the LLM agent.

\FloatBarrier

\subsection*{Monitoring the differential diagnosis at every step}

We next asked whether the record shows which options the agent was considering at each step and which result changed its assessment. 
We queried the agent's state for the differential and for the diagnosis it would code at each point.
The differential gave a probability for each tested disease and a probability that none were present.
We ran both queries separately from the controller's work-up: First with just the history, then after each action.
The decision model is fast enough to run the same queries during a work-up.
We did this in all 5 runs with Jev for each of the 546 admissions.
\Figref[, A to~C]{fig:fig3}, shows the differential for 3 work-ups alongside the probabilities the controller used at each step.

\begin{figure}[!htbp]
    \centering
    \captionsetup{font=small}
    \includegraphics[width=\textwidth,height=\textheight,keepaspectratio]{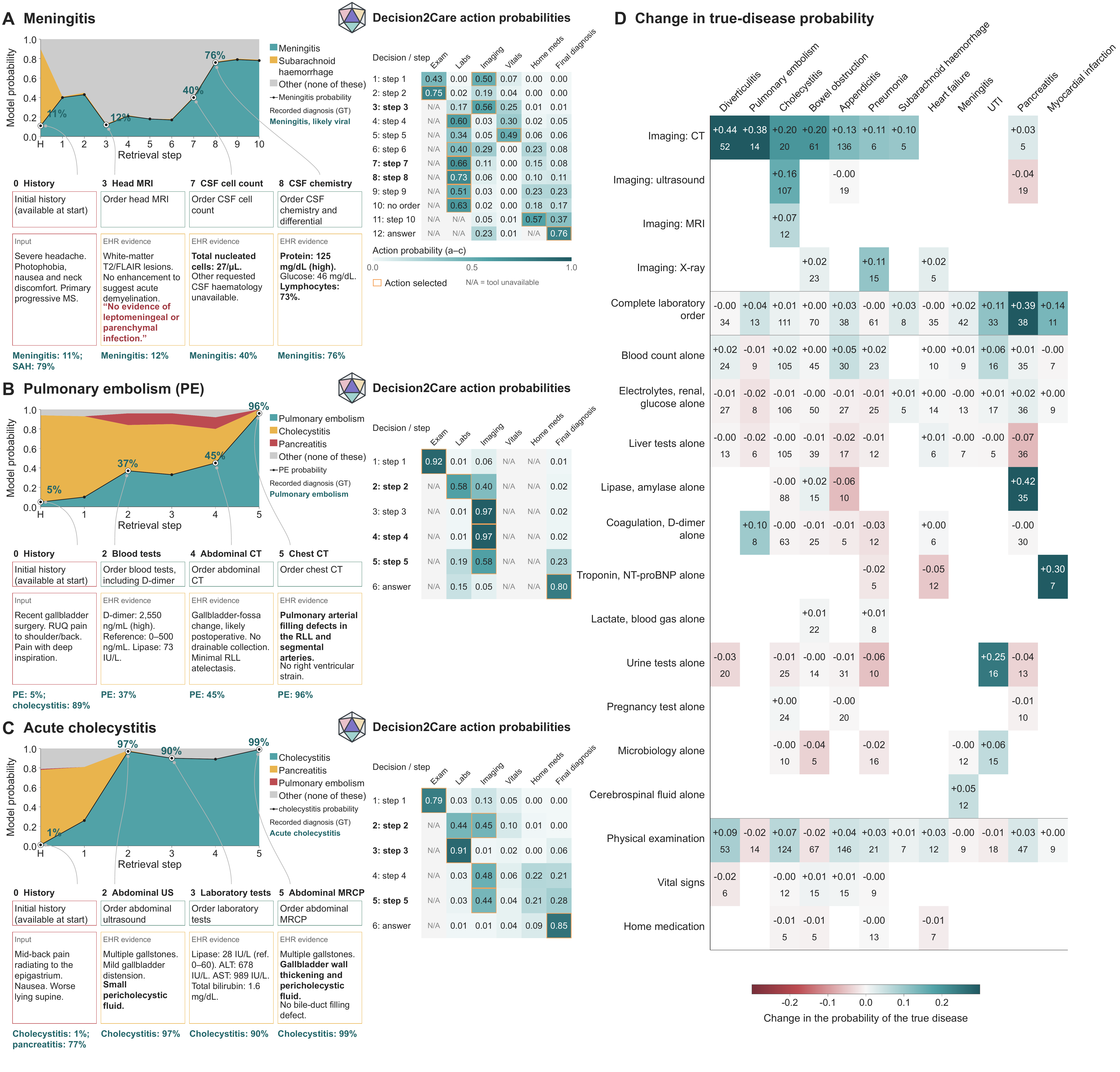}
    \caption{\textbf{The differential diagnosis after every action.}
    \textbf{A}--\textbf{C}, The differential diagnosis during 3 work-ups of the baseline run.
    The stacked areas show the probability of each disease and of none of the tested diseases (grey) on the history alone (H) and after each step.
    The true disease is shown in teal, and the line traces its probability.
    The cards give, for the history and the circled steps, the action, the findings returned and the resulting probability of the true disease.
    The table shows the probability assigned to each action at each decision, with the action taken outlined in orange and the circled steps in bold.
    \textbf{A}, A work-up of meningitis.
    \textbf{B}, The work-up of pulmonary embolism shown in \figref{fig:overview}.
    \textbf{C}, A work-up of acute cholecystitis.
    \textbf{D}, The mean change in the probability of the true disease per action, by disease, over 5 runs on the 546 paired admissions.
    }
    \label{fig:fig3}
\end{figure}

These cases began with an incorrect top-ranked disease at a probability of 0.50 or more.
They changed to the correct disease at a guideline test (\supptabref{tab:supp-guideline-tests}; \methodsref{meth:showcase}).
In the first, a patient with multiple sclerosis and a severe headache (\figref[A]{fig:fig3}), a normal head CT changed the top-ranked disease from SAH (0.79) to none of the 12 diseases.
Magnetic resonance imaging (MRI) without meningeal enhancement then lowered the probability of meningitis from 0.43 to 0.12, and the CSF chemistry raised it to 0.76, where it remained top-ranked until the end of the work-up.
In the second case (\figref{fig:overview}), the differential on the history alone was led by cholecystitis (0.89) although the gallbladder had been removed, and PE stood at 0.05.
The chest CT raised the probability of PE from 0.45 to 0.96 (\figref[B]{fig:fig3}).
In the third case, an abdominal ultrasound raised the probability of cholecystitis from 0.26 to 0.97 in a patient whose history suggested pancreatitis (\figref[C]{fig:fig3}).
In each instance the probability of a disease went down when a result spoke against it.
The agent kept ordering tests as it deemed the evidence insufficient.
Where the leading diagnosis in the differential changed, the coded diagnosis changed at the same step or the next.
It arrived at the correct diagnosis in all 5 runs for each of the 3 cases.

Across the 546 admissions the true disease was top of the differential in 71.4\% of case-runs on the history alone and 93.3\% after the work-up (\suppfigref[A]{fig:supp-differential}).
The mean probability of the true disease rose from 0.68 to 0.91.
The coded diagnosis read at the same steps was correct in 59.3\% and 86.5\% of cases.
The evidence changed the top-ranked disease from incorrect to correct in 123.8 cases per run (22.7\%; range 122--125) and from correct to incorrect in 4.0 (0.7\%; range 3--5; \suppfigref[B]{fig:supp-differential}), indicating that the results obtained seldom misled the differential.

The probability of true disease increased by an average of 0.21 after steps that returned a CT imaging report (95\% CI 0.17--0.25; 303 reports per run).
The mean increase was 0.11 after an ultrasound report (95\% CI 0.07--0.15; 146 reports per run), 0.05 after a laboratory order (95\% CI 0.03--0.07) and 0.04 after the physical examination (95\% CI 0.02--0.05).

In 10 of the 12 diseases, the largest mean increase followed a test the guideline uses to make the diagnosis (\figref[D]{fig:fig3}).
CT in diverticulitis ($+0.44$), PE ($+0.38$) and the laboratory order in pancreatitis ($+0.39$) are all examples of this.
When we looked at the laboratory tests individually (\methodsref{meth:labsplit}), the largest increases were again seen for the tests that are named in the guideline:
The probability increased by 0.42 with lipase and amylase in pancreatitis, 0.30 with troponin in MI, 0.25 with urine tests in UTI and 0.10 with the coagulation tests (including D-dimer) in PE.
For all the other individual laboratory test groups, the mean change was within 0.07 of zero.

Of the 2,730 case-runs (546 cases times 5 repeats), 1,009 had at least one step which increased the probability of the true disease by 0.20 or more.
In 760 of those (75.3\%) the greatest increase in probability occurred after a guideline test.
For an action selected at random from the same case, the expected proportion was about 30\% (\methodsref{meth:guideline}; \suppfigref[C]{fig:supp-differential} and \supptabrefs{tab:supp-guideline-tests}{tab:supp-decisive-steps}).

To test whether the diagnosis rested on these steps, we rebuilt the final chart of each case in the baseline run without one source at a time. 
We repeated diagnostic scoring using the remaining evidence (\methodsref{meth:leaveout}).
Removing the CT reports lowered accuracy by 24.4 points among the 303 cases with an available CT report (exact McNemar test, $p < 0.001$), the ultrasound reports by 16.4 points (146 cases; $p < 0.001$) and lipase and amylase by 9.9 points (152 cases; $p = 0.004$).
By contrast, removing routine blood tests, the examination, or vital signs changed accuracy by 1.3 points or less (\suppfigref[B]{fig:supp-ablation}).
In pancreatitis, removing lipase and amylase from the 37 cases that had them lowered accuracy by 54.1 points, whereas removing any other source left it unchanged (\suppfigref[A]{fig:supp-ablation}).
The stepwise differential thus showed which diagnoses the agent considered and how each result changed its assessment, while withholding evidence tested whether the final diagnosis depended on it.
\hyperref[fig:supp-diagnosis-trajectories]{\textbf{Figures~\ref*{fig:supp-diagnosis-trajectories} to~\ref*{fig:supp-diagnosis-trajectories-3}}} show one work-up for each of the remaining nine diseases.

\FloatBarrier
\subsection*{Deferring uncertain diagnoses to a clinician after a single work-up}

\begin{figure}[!htbp]
    \centering
    \captionsetup{font=small,skip=4pt}
    \includegraphics[width=\textwidth]{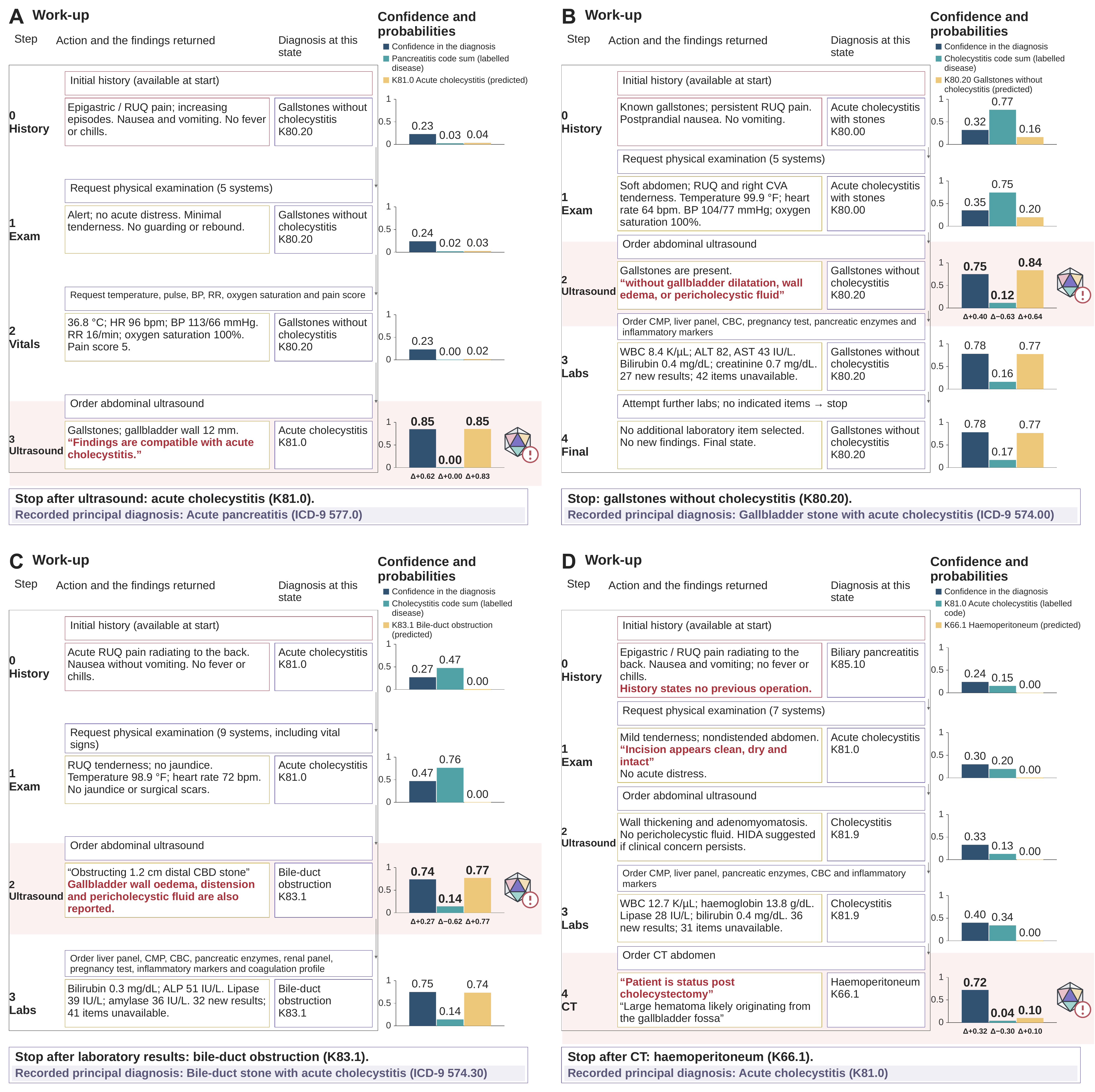}
    \caption{\textbf{Audit of the 4 high-confidence diagnostic errors.}
    \textbf{A}--\textbf{D}, The 4 admissions out of 546 in which Decision2Care retained an incorrect diagnosis at a confidence of at least 0.70 in the baseline run.
    Each table details the history and the condensed findings from each action, along with the diagnosis the agent would have coded at that point.
    The bars show the confidence in that diagnosis (navy), the summed probability of the ICD-10-CM codes that match the recorded disease under the benchmark's rule (teal) and the probability of the code the agent finally chose (gold).
    The pink band indicates the arrival of the implicated imaging report and delta ($\Delta$) indicates the change from the previous state.
    The red text highlights the evidence that led to the change in assessment.
    The quotation marks denote verbatim extracts.
    The agent's final diagnosis and the recorded principal diagnosis are both shown at the bottom of each work-up.
    \textbf{A}, Gallstone pancreatitis coded as acute cholecystitis.
    \textbf{B}, Cholecystitis coded as gallstones without cholecystitis.
    \textbf{C}, Cholecystitis coded as bile duct obstruction.
    \textbf{D}, Cholecystitis coded as haemoperitoneum with the single recorded code shown in teal.
    The CT report describes a postoperative haematoma.
    The benchmark uses the recorded principal diagnosis as the reference.
    }
    \label{fig:fig4}
\end{figure}

We next asked whether the agent could tell, from a single work-up, which of its diagnoses to act on and which to hand to a clinician\cite{geifman2017selective,dvijotham2023codoc}.
MIRA-v2 answers this question with the agreement of 5 repeated work-ups\cite{zhang2026onprem}.
Decision2Care's signal is its confidence in its diagnosis, one probability queried after a single work-up (\methodsref{meth:signoff}).
MIRA-v2's signal is ConsistencyDx, the agreement among the diagnoses of its 5 runs, computed with its own code\cite{zhang2026onprem}.
A case counted as correct when most of its runs were correct (\methodsref{meth:consistency}).
At list prices, these 5 runs cost US\$103--399 for 546 admissions, depending on the LLM, compared with US\$1.31 for a single Decision2Care work-up with its confidence request (\tabref{tab:diagnostic-efficiency}).
Decision2Care's confidence separated correct from incorrect diagnoses about as well as ConsistencyDx (\suppfigref{fig:supp-deferral}).
Its AUROC was 0.831 (95\% CI 0.784--0.876), compared with 0.827, 0.830, and 0.870 for ConsistencyDx with GLM-5, GPT-OSS-120B, and Claude Sonnet 5, and the bootstrap intervals of the 3 differences included zero.

We used confidence from a single work-up to flag cases for review.
A threshold was set to retain approximately half the cases (\methodsref{meth:signoff}).
In the baseline run Decision2Care retained 269 of 546 (49.3\%) and correctly diagnosed 265 of those (98.5\%; 95\% CI 96.2--99.4\%).
Of the 71 incorrect diagnoses, 67 were in cases flagged for review, and 4 were in retained cases (\suppfigref[, A and~B]{fig:supp-deferral-comparators}).
Decision2Care retained 130 of the 148 appendicitis admissions and correctly diagnosed all 130.

At the same coverage, MIRA-v2's 5 work-ups gave a similar result.
The 269 cases with the highest ConsistencyDx contained 9, 5 and 3 errors with GPT-OSS-120B, GLM-5 and Claude Sonnet 5, compared with 4 for Decision2Care (\supptabref{tab:supp-deferral}).
Each system also has its own threshold, 0.70 for Decision2Care and the published 0.90 for MIRA-v2\cite{zhang2026onprem}.
At these thresholds, Decision2Care sent 94.4\% of its errors to review, compared with 82.6--89.1\% for MIRA-v2.
Its retained cases were 98.5\% correct, compared with 96.0--98.4\%.
MIRA-v2 retained more of the cases, 54.9--62.8\% (\suppfigref[, C to~H]{fig:supp-deferral-comparators}; \supptabref{tab:supp-deferral}).
Thus, a confidence read after 1 work-up retained about half of the cases as reliably as the agreement of 5 work-ups.

For each of the 4 errors it retained (\figref[, A to~D]{fig:fig4}), the audit showed where the work-up went wrong.
All 4 were hepatobiliary, and each could be traced to a single imaging report, at which the confidence rose from 0.23--0.47 to 0.72--0.85.
Without that report, repeating diagnostic scoring yielded the correct diagnosis in 3 of the 4 admissions.
In the remaining admission (\figref[A]{fig:fig4}), a gallstone pancreatitis for which no pancreatic enzymes had been ordered, adding the record's lipase to the chart left the diagnosis unchanged but lowered the confidence to 0.62, below the threshold (\methodsref{meth:leaveout}).
The error in \figref[D]{fig:fig4} had a different cause.
The CT was taken on the day after the cholecystectomy, and the agent coded the haemoperitoneum that the report described, whereas the reference diagnosis was the cholecystitis that had led to the operation.

If the confidence follows the evidence, it should fall when evidence is withheld or misleading.
In 5 runs with the history withheld (\methodsref{meth:stress}), Decision2Care retained 35.1\% of cases (down from 49.7\% in the 5 runs on the complete record).
Overall accuracy dropped from 86.5\% to 59.4\%.
Of the cases that were retained, 97.5\% were correctly diagnosed (\supptabref{tab:supp-stress}).
In 5 further runs with the decisive results replaced by normal findings the agent retained 4.2\% of cases (20--25 per run).
Overall accuracy was 22.2\%. Of the retained cases, 84.3\% were correctly diagnosed.
These results show that Decision2Care could identify from a single work-up which diagnoses to hand to a clinician.

\FloatBarrier
\subsection*{Decision agents extend to treatment selection}

After diagnosis, Decision2Care used the same 2 types of decisions to select the treatment and recorded the steps that led to each order (\figref{fig:treatment-selection}).
The first step was to decide whether or not each medication category or ICD-9-CM procedure group was indicated.
The second step was to select the regimen or procedure, or how to manage each of the patient's home medications.
The final step was to check each order against the patient's allergies and renal function (\methodsref{meth:treatment}).
\Figref[A]{fig:treatment-selection} shows an antibiotic order for a patient with appendicitis and a listed cefaclor allergy.
The agent first selected metronidazole with ceftriaxone.
The allergy check flagged ceftriaxone on the first attempt and cefazolin on the second.
The third regimen (metronidazole with ciprofloxacin) passed both checks.
The order can thus be audited through the 3 choices and the 2 flags that led to it.

\begin{figure}[!htbp]
    \centering
    \captionsetup{font=small}
    \includegraphics[width=\textwidth]{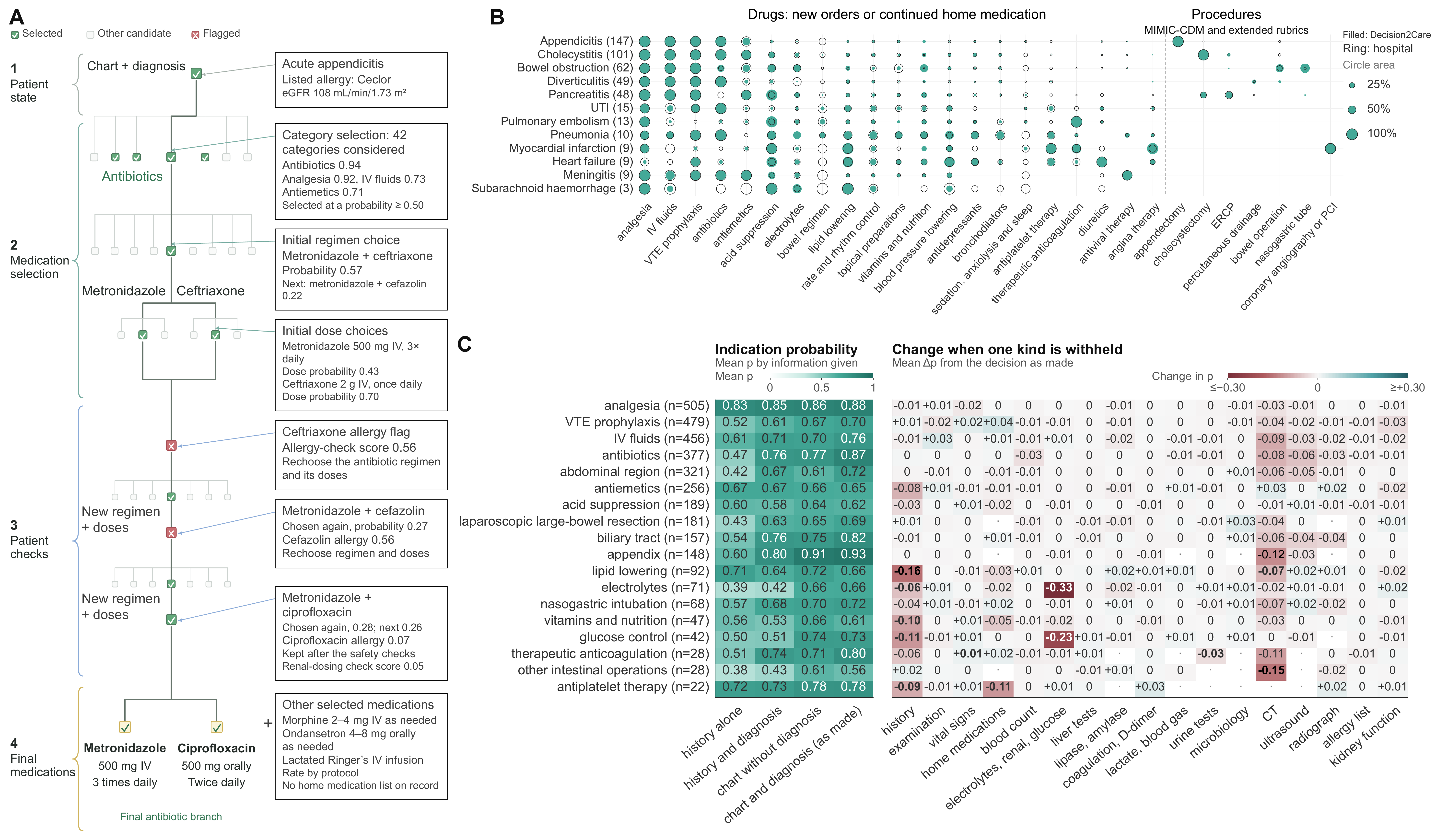}
    \caption{\textbf{Treatment selection and its dependence on patient information.}
    \textbf{A}, The medication decisions for one appendicitis admission with a listed cefaclor allergy.
    The agent selects medication categories, a regimen, and doses (2).
    It then checks each drug for a listed or cross-reacting allergy and for an excessive dose or a drug to avoid at the recorded kidney function (3).
    Here, the allergy check flagged ceftriaxone, then cefazolin; metronidazole with ciprofloxacin was kept (4).
    Categories were selected and concerns flagged at scores of at least 0.50, and regimen and dose probabilities are shared among the options offered.
    \textbf{B}, Treatment selections by disease in the 475 admissions that the agent diagnosed correctly.
    Filled circles give the share of admissions in which the agent selected each medication category or procedure, and rings the share in which the hospital recorded it in first-day orders or admission procedures.
    Circle area is proportional to the share.
    \textbf{C}, The probability that each treatment was indicated, decided again on all 546 admissions with information withheld.
    Rows are the treatments that the agent selected, with the number of admissions (n); procedure groups carry ICD-9-CM category names.
    The left heat map shows the mean probability for each set of information, with the diagnosis as the decision made.
    The right heat map gives the mean change from that decision when one kind of information is withheld and the diagnosis is kept.
    Bold marks cells in which at least 25\% of decisions fell below 0.50, and dots mark cells with fewer than 3 admissions to withhold from.
    }
    \label{fig:treatment-selection}
\end{figure}

We compared the selected treatments against the hospital records for the 475 correctly diagnosed admissions (\figref[B]{fig:treatment-selection}).
Decision2Care selected an appendectomy in all 147 of the appendicitis cases and a cholecystectomy in 100 of the 101 cholecystitis cases.
The corresponding rates in the hospital records were 92.5\% and 90.0\%.
We then repeated the indication decisions for all 546 admissions, omitting one type of information at a time (\figref[C]{fig:treatment-selection}).
Omitting electrolytes, renal, and glucose results produced the largest mean reductions in the probability of ordering electrolyte replacement and glucose control, of 0.33 and 0.23, respectively.
Of the 285 probability changes shown, 264 were within 0.05 of zero.

We then compared Decision2Care's orders with those of MIRA-v2 and with the care documented on the first day and in the ED.
This comparison used the 437 of these 475 admissions in which the majority of each comparator's 5 runs were also correct (\methodsref{meth:rubric}).
As in the frameworks whose rubrics we used\cite{hager2024cdm,ferber2026mira}, we scored treatment only after a correct diagnosis.
We first measured how many of the treatments indicated by these rubrics each system ordered.
Decision2Care ordered 97.0\% of the 1,024 indicated treatments, compared with 83.5\%, 81.5\% and 95.1\% for MIRA-v2 with GPT-OSS-120B, GLM-5 and Claude Sonnet 5 (\suppfigref[A]{fig:supp-treatment-comparison}).
It ordered every indicated treatment for 10 of the 19 items with 5 or more cases (by item in \supptabref{tab:supp-treatment-indicated}).
After correction for multiple comparisons, 7 differences remained, all with higher adherence for Decision2Care (Benjamini--Hochberg-adjusted $q < 0.01$ for each).
Six concerned supportive care, which GLM-5 ordered in 25--73\% of cases across 5 of the 6 disease groups and GPT-OSS-120B in 61\% of pancreatitis cases.
The seventh concerned cholecystectomy in cholecystitis, which GPT-OSS-120B ordered in 66.5\% of cases.
Compared with documented care, which could be scored for 14 items, only antibiotics in appendicitis remained significantly different, with Decision2Care ordering them in 99.3\% of admissions versus 84.7\% ($q < 0.001$).

We then judged the drug orders by category against the guidelines with the evaluator of Ferber et~al.\cite{ferber2026mira}, on 355 of the 437 admissions with one of its 7 diseases (\methodsref{meth:judge}).
Averaged over a patient's categories, 88.2\% of Decision2Care's decisions were judged adherent (95\% CI 86.0--90.3\%), compared with 65.6\%, 58.7\% and 79.7\% for the comparator and 81.4\% for the documented first-day care (Wilcoxon signed-rank test, $p < 0.001$ for each; \suppfigref[B]{fig:supp-treatment-comparison}).
The largest differences were in intravenous fluids, analgesia for appendicitis and cholecystitis, and antibiotics for appendicitis, with Decision2Care meeting these orders in 97.2--100\% of cases and the comparator in 4.5--98.7\%.
Of the 48 category comparisons, 13 remained significant after correction, with Decision2Care performing better in 12 (\supptabref{tab:supp-guideline-categories}).
Compared with documented first-day care, 2 of the 16 categories differed significantly after correction, both with higher adherence for Decision2Care: antibiotics in appendicitis (97.2\% versus 66.0\%) and pancreatitis (100\% versus 66.0\%; $q < 0.001$ for both).

Finally, Decision2Care's orders agreed with the drug classes given on the first day or in the ED at least as closely as those of the comparator.
Its F1 score (the harmonic mean of precision and recall) was 40.3, compared with 34.7--40.2.
With enoxaparin and heparin for prophylaxis of venous thromboembolism counted as 1 class, it was 50.7, compared with 35.4--40.3 (\suppfigref[C]{fig:supp-treatment-comparison}; \supptabref{tab:supp-treatment-measures}).
Overall, Decision2Care ordered nearly every treatment the rubrics indicated. 
Its drug orders were judged guideline-adherent more often than those of the LLM agents and the documented first-day care, and each order could be traced through the choices and checks that produced it (\figref[A]{fig:treatment-selection}).

\FloatBarrier
\subsection*{Decision agents extend to outcome prediction}

We finally asked whether the same kind of decision could also predict death.
Given the disease, the agent chose, step by step, which records to read and when to stop (\figref{fig:mortality-audit}; \methodsref{meth:prognosis}).
For a prediction made 8~h after arrival in the ED, it then gave the probability of death within 7, 30 and 90 days, and in the ICU the probability of death before discharge.
We examined this prediction in 5 ways.
We followed it record by record in individual patients, used it to rank patients at 8 h after arrival, repeated it on records from a second hospital, compared it with the predictions of ICU clinicians who were given every record, and compared it hour by hour in the ICU with 3 severity scores.

\begin{figure}[!htbp]
    \centering
    \captionsetup{font=small,skip=6pt}
    \includegraphics[width=\textwidth]{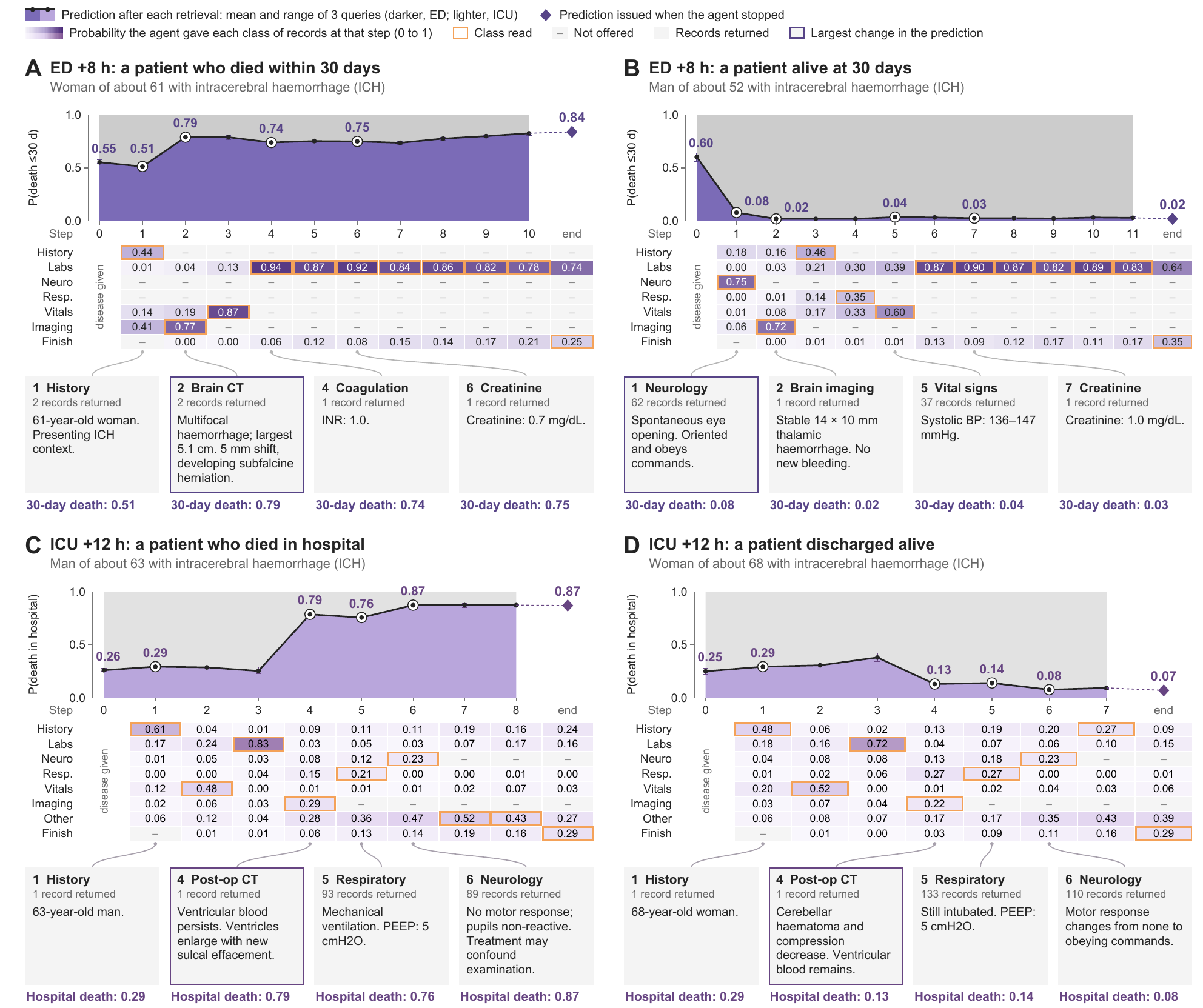}
    \caption{\textbf{Mortality predictions in intracerebral haemorrhage (ICH).}
    \textbf{A}--\textbf{D}, Predictions after each retrieval in 4 patients with ICH.
    \textbf{A}, \textbf{B}, The probability of death within 30~days, predicted 8~h after arrival in the emergency department (ED), for a patient who died within 30~days (\textbf{A}) and for one who was alive at 30~days (\textbf{B}).
    \textbf{C}, \textbf{D}, The probability of death in hospital, predicted 12~h after admission to the intensive care unit (ICU), for a patient who died in hospital (\textbf{C}) and for one who was discharged alive (\textbf{D}).
    The curve gives the prediction with the disease alone (step~0) and after each retrieval.
    Shown are the mean and the range of 3 predictions made on the records read up to that step; the diamond is the prediction that the agent issued when it stopped.
    The heatmap shows the probability the agent gave to each class of records at each step, summed over the actions for that class.
    The class that it read is outlined in orange, and a dash marks a class that was not offered.
    The cards show the records returned at 4 steps, with the prediction after each.
    \label{fig:mortality-audit}}
\end{figure}

We showcase 4 patients with intracerebral haemorrhage (ICH), 2 in the ED and 2 in the ICU.
In each setting one had a high predicted probability of death and died, and the other had a low one and survived (\figref{fig:mortality-audit}; for such a pair in each of the 8 diseases, \hyperref[fig:supp-ed8-audit-1]{\textbf{figs.~\ref*{fig:supp-ed8-audit-1} to~\ref*{fig:supp-icu12-audit-2}}}; \methodsref{meth:prognosis}).
In these patients, the prediction could be followed one record at a time, and it changed with the content of a record rather than with its type.
In the ED, with only the disease known, the probability of death within 30~days was 0.55 for a woman of about 61 and 0.60 for a man of about 52.
It rose to 0.79 when the agent read her CT report (multifocal haemorrhage, developing herniation) and fell to 0.08 when it read his neurological observations (eyes open, oriented, obeying commands).
She died within 30~days and he survived (\figref[, A and~B]{fig:mortality-audit}).
The 8 and 9 retrievals that followed in these 2 patients, mostly single laboratory values, changed it by no more than 0.05 at any step.
In the ICU, a postoperative CT report moved the probability of death in hospital from 0.25 to 0.79 in a man whose ventricles had enlarged, who died, and from 0.38 to 0.13 in a woman whose haematoma had shrunk, who was discharged alive (\figref[, C and~D]{fig:mortality-audit}).
Enlargement of the ventricles (hydrocephalus) and haematoma growth are both associated with worse outcomes after ICH\cite{greenberg2022ich}.

At 8 h after arrival, the prediction ranked the patients who died within 30 days with an AUROC of 0.871 in 8,565 entries of 8 disease cohorts from MIMIC-III and MIMIC-IV (0.849 for death within 90 days in 8,570 entries, \figref[, A and~B]{fig:fig6}; by disease in \suppfigref{fig:supp-mortality-30-disease} and \suppfigref{fig:supp-mortality-90-disease}).
By cohort the 30-day AUROC ranged from 0.72 in candidates for acute respiratory distress syndrome (ARDS) in MIMIC-III (38 patients) to 0.90 in ICH in MIMIC-IV (652; \supptabref{tab:supp-prognosis-cohorts}), indicating that the prediction ranked patients within each disease and database and not only between them.

\begin{figure}[!htbp]
    \centering
    \captionsetup{font=small,skip=6pt}
    \includegraphics[width=\textwidth]{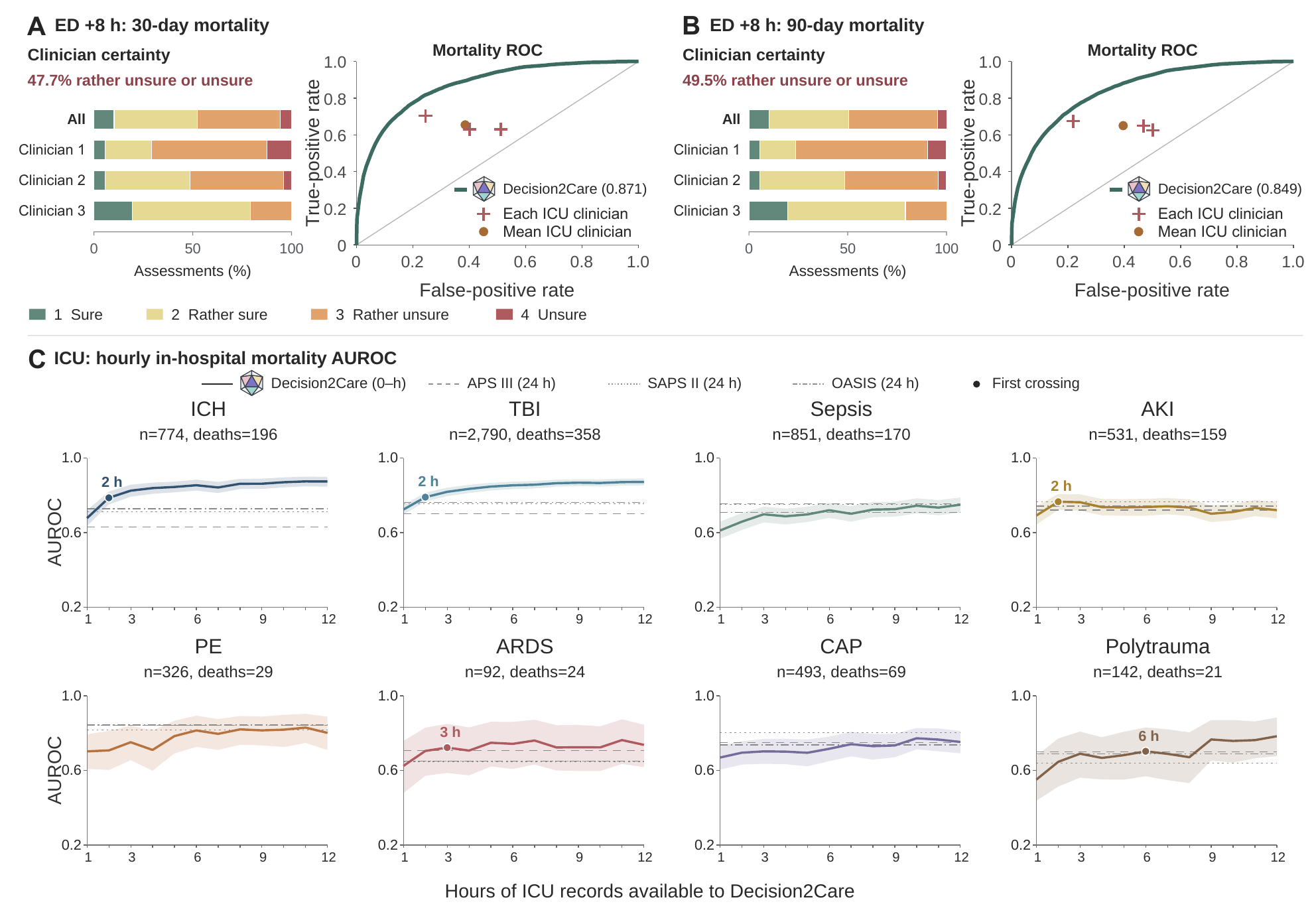}
    \caption{\textbf{Accuracy of the mortality predictions in the emergency department (ED) and intensive care unit (ICU).}
    \textbf{A}, \textbf{B}, Receiver operating characteristic (ROC) curves of the prediction made 8~h after arrival in the ED for death within 30~days (\textbf{A}) and 90~days (\textbf{B}), from the mean of 5 runs on 8 disease cohorts (8,565 and 8,570 entries; 1,673 and 2,062 deaths).
    Crosses show the predictions of 3 ICU clinicians, and the dot is their mean, on 72 cases in which wrong agent predictions were over-represented.
    The stacked bars show their confidence.
    \textbf{C}, AUROC for predicting in-hospital death from the first 1 to 12~h of ICU records, by disease.
    Bands are 95\% bootstrap intervals.
    Grey lines show the Acute Physiology Score III (APS~III), the Simplified Acute Physiology Score II (SAPS~II) and the Oxford Acute Severity of Illness Score (OASIS) at 24~h in the same patients.
    The dot marks the first hour at or above the highest score.
    \label{fig:fig6}}
\end{figure}

All of these cohorts came from Beth Israel Deaconess Medical Center, so we repeated the prediction, with the agent unchanged, on INSPECT\cite{huang2023inspect}, a data set of CT pulmonary angiography examinations at Stanford Health Care (\suppfigref{fig:supp-second-centre}; \methodsref{meth:secondcentre}).
We restricted the evaluation to pulmonary embolism (PE) and pneumonia because INSPECT provides radiology report impressions, laboratory values and vital signs, but no presenting history or clinical notes.
In 555 and 542 patients with PE (41 and 65 deaths), the prediction ranked the patients who died within 30 and 90 days with AUROCs of 0.813 and 0.800 (95\% bootstrap intervals 0.746--0.873 and 0.743--0.851).
In 78 and 76 patients with pneumonia that could not be verified as community-acquired (6 and 8 deaths), the AUROCs were 0.770 and 0.762 (0.658--0.883 and 0.646--0.871; \supptabref{tab:supp-prognosis-cohorts}).
These AUROCs did not differ from those of the MIMIC cohorts of the same 2 diseases (differences of $-0.07$ to $+0.04$ in PE and $-0.11$ to $+0.03$ in pneumonia, with intervals including zero).

To compare the prediction with those of clinicians, we asked 3 ICU clinicians to predict the outcome of, and then to review, 72 cases from the 8 MIMIC cohorts, 7--10 from each disease (\supptabref{tab:supp-readers}; \methodsref{meth:readers}).
We selected cases so the agent's wrong predictions were overrepresented (33 of the 72 predictions of 90-day mortality, 46\%, compared with 26\% in the cohorts).
The ICU clinicians were given every record offered to the agent, a median of 124 per case, of which the agent had read a median of 54\%.
The agent read nearly all the records on which the ICU clinicians relied (1,297 of the 1,397 available records they marked essential for their own predictions, 92.8\%).

In 196 of their 216 reviews (91\%) they found no essential record left unread (\supptabref{tab:supp-readers-review}; \suppfigref{fig:supp-retrieval-audit}).
In the same cases, the agent's AUROC did not differ from the ICU clinicians' mean prediction at 30 or 90 days.
It did not differ from that of a single ICU clinician in 5 of 6 comparisons (\figref[, A and~B]{fig:fig6}; \supptabref{tab:supp-readers}).
Weighted back to the cohorts from which the cases were drawn, its AUROC was 0.90 at 30 days and 0.84 at 90 days, within the range of the 3 ICU clinicians at both (0.88--0.93 and 0.82--0.89; \methodsref{meth:readers}).
Of the 33 cases in which the agent's prediction of 90-day mortality was wrong, at least 2 of the 3 ICU clinicians were also wrong in 20.
When shown the agent's prediction, the ICU clinicians rated it as at least defensible in 182 of their 216 reviews (84\%; \supptabref{tab:supp-readers-review}).
The agent's requests for a case took a median of 2.2~s, whereas an ICU clinician's own prediction took a median of 3.4~min of active time and the review that followed it 1.2~min.
Thus, from about half of the records, the agent predicted death as well as ICU clinicians who had all of them.

In the ICU, the agent predicted in-hospital death from the first 1 to 12~h of records, with a separate prediction for each hour, in 5,999 entries of the 8 disease cohorts (1,026 deaths).
We compared its AUROC with the Acute Physiology Score III (APS~III)\cite{knaus1991apache}, Simplified Acute Physiology Score II (SAPS~II)\cite{legall1993saps} and Oxford Acute Severity of Illness Score (OASIS)\cite{johnson2013oasis} at 24~h in the same patients (\figref[C]{fig:fig6}; \supptabref{tab:supp-icu-hourly}; \methodsref{meth:icu}).
In ICH and traumatic brain injury (TBI), the 12~h prediction exceeded all 3 scores computed at 24~h.
Its AUROC was 0.874 and 0.872, 0.15--0.24 and 0.11--0.17 above the scores (paired bootstrap intervals excluding zero; \supptabref{tab:supp-icu-scores}).
In ICH, it was already above the highest score with 2~h of records (AUROC 0.787, compared with 0.728 for OASIS; the interval of the difference excluded zero).
In ICH and TBI, the agent read the neurological observations newly recorded each hour in 91--99\% of the cases that had them (\suppfigref[, D and~E]{fig:supp-hourly-retrieval}).
In the other 6 cohorts, sepsis, acute kidney injury (AKI), the ARDS candidates, polytrauma, PE and community-acquired pneumonia (CAP), the prediction from 12~h of records did not differ from the scores computed at 24~h (differences of $-0.05$ to $+0.14$, with intervals including zero).
In 5 of the 8 cohorts, the AUROC of the prediction first reached or exceeded that of the highest score within 2 to 6~h of records (after 2~h in ICH, TBI and AKI, 3~h in the ARDS candidates and 6~h in polytrauma; \figref[C]{fig:fig6}; \supptabref{tab:supp-icu-hourly}).
The same kind of decision thus predicted death at 8~h after arrival and hour by hour in the ICU, and each prediction can be audited against the records read to make it (\figref{fig:mortality-audit}).

\FloatBarrier
\section*{Discussion}

Decision2Care used only typed decisions to diagnose acute illness, select treatments, and estimate mortality risk.
Each decision was a choice from a set of named options (e.g.\ ICD-10-CM codes) or a judgement about the truth of a statement (e.g.\ whether to order a particular test).
On the same 546 emergency admissions across 12 diseases, Decision2Care's diagnostic accuracy fell within the range achieved by MIRA-v2\cite{zhang2026onprem} with 3 LLMs.
Its work-ups took a fraction of the time and cost far less.
The confidence from a single Decision2Care work-up was comparable to the agreement across multiple MIRA-v2 work-ups in flagging cases for clinician review.
Decision2Care's drug orders were more often judged to be guideline-adherent than the drug orders that clinicians themselves documented on the first day.
The agent predicted mortality in emergency patients to a degree similar to 3 ICU clinicians with access to the full chart.
For 10 of the 12 diseases, the differential changed most after the agent saw the result of the guideline's confirmatory test.
Accuracy dropped when we repeated the diagnosis without the CT reports, ultrasound reports, or the lipase and amylase results.
Withholding the routine blood tests, examination findings, and vital signs had little effect.
Together, these results show that a record of typed decisions is enough both to carry out a work-up and to review it.

An LLM-based clinical agent\cite{yao2023react,hager2024cdm,ferber2026mira,zhang2026onprem} can be asked to produce a rationale for an action, but that rationale may omit the evidence on which the action was based\cite{turpin2023unfaithful}. 
By contrast, Decision2Care records the question, the model's probabilities and the clinical state at each step.
The controller applies a set of fixed rules to those probabilities so a reviewer can trace each action back to the scores used to select it.
We queried the saved states separately to read the differential after each action.
A typed question takes a fraction of a second to answer.
We could therefore re-run a decision, or a full work-up, with a result removed or replaced and see how the diagnosis and its confidence changed.
The same audit applies to drug orders and mortality predictions.
In earlier work, we inspected predictions through clinical concepts\cite{han2026clear} or counterfactual images\cite{han2024confounders}. 
Here we extend that inspection to the actions that an agent takes during a work-up.
The need is not specific to the clinic: a multi-agent system for drug development retained its analysis artifacts for review, and human experts re-implemented its core analyses to confirm them\cite{zhang2026virtualbiotech}.

The decision model scores only the options it is presented with\cite{almeida2026jev,zhang2026cs1}.
Its menus therefore cannot contain a hallucinated test. 
The menus are in plain text so the same model\cite{almeida2026jev,zhang2026cs1} can be used to score new questions without the need for retraining.
We examined 5 diseases in addition to the 7 in the MIRA-v2 benchmark and created new questions for both treatment selection and prognosis.
The same approach could be used for other diseases and for records from hospitals beyond Beth Israel Deaconess Medical Center and Stanford Health Care.
A generative model could propose options for a typed question, and the decision model could score them\cite{zhang2026cs1}.

We used the same controller, tools and questions to test 2 alternatives to Jev, one based on the answer token probabilities of a general purpose LLM.
Both achieved diagnostic accuracy comparable to that of Jev (\supptabref{tab:supp-decision-models}) and ran on a single NVIDIA DGX Spark.
A hospital could run the agent on that hardware and keep their patient records on site.

This work has limitations.
We evaluated diagnostic confidence as a ranking score\cite{vancalster2019calibration,savage2025uncertainty}. 
We would need to assess its calibration in the target population, and establish and validate thresholds for diagnostic confidence and mortality prediction there.
Confidence dropped when we withheld the history or replaced the key results with normal values and the agent flagged more cases for review in those circumstances.

Decision2Care can be confident in a wrong diagnosis.
In every instance where the agent retained a wrong diagnosis, the audit traced the increase in confidence back to a single imaging report (\figref{fig:fig4}).
In a patient whose recorded diagnosis was pancreatitis, the ultrasound report described the findings as `compatible with acute cholecystitis'.
The agent took that wording as the diagnosis.
A clinician would have balanced that report against the history, examination and laboratory results.
Repeating the diagnosis with the relevant reports excluded corrected the majority of these errors.
A similar error is visible at the start of the work-up in \figref{fig:overview}, where the history gave acute cholecystitis as the reason for a recent cholecystectomy.
Cholecystitis led the differential at 0.89 before any test, although the gallbladder had been removed, and the results that followed corrected it.
The agent read the radiologist's report and not the images.
A vision-language model now reaches expert-level diagnosis on abdominal CT\cite{zhang2026radar}, and such a model could give a second reading when a single CT report drives the diagnosis.

Early decisions to limit care are associated with death\cite{greenberg2022ich}, and a mortality prediction may partly reflect those decisions if they are recorded in the chart.
A reviewer can see how the Decision2Care estimate changes as it goes through each record (\figref{fig:mortality-audit}).

Clinicians determine what to do in response to the agent's diagnoses, suggested orders, and predictions. 
Accountability depends on what each party can know and control \cite{habli2020accountability}, and the American Medical Association has proposed sharing responsibility among the developers, the hospital using the AI, and the clinician \cite{ama2026liability}. 
Decision2Care supports clinical oversight by recording the evidence available at each stage and the probability assigned to each option, and by restricting the agent's actions to a predefined set of options. 
This creates a record that the clinician can use to approve, modify, or reject a proposed order. 
Prospective studies can then look at how clinicians use the record and whether and when it leads to a change in their decision.

The evaluation was retrospective.
Prospective studies would be needed to assess how clinicians use the agent, how much case review costs and whether there is any patient benefit.
Evaluations of LLMs on clinical reasoning tasks have called for the same trials\cite{brodeur2026reasoning}, and comparisons made in a controlled benchmark do not establish performance in clinical practice\cite{nori2026sequential}.
A direct test would be to pre-specify how a particular finding should lead to a decision and then see what the agent does\cite{wu2026act}.
Where no guideline names the decisive findings, the results can be removed from the chart to see which findings the agent was relying on.
The recorded decisions allow a reviewer to review a missed diagnosis step by step, as is done at a morbidity and mortality conference.

\FloatBarrier
\endgroup
\clearpage
\renewcommand{\refname}{References and Notes}
\bibliographystyle{science_refs}
\bibliography{references}

\section*{Acknowledgments}

\noindent\textbf{Funding:}
The authors received no specific funding for this work.

\noindent\textbf{Author contributions:}
T.H. and D.T. conceived the study and designed the research.
T.B. and R.W. developed the software, curated the datasets and performed the computational experiments.
T.B., R.W. and T.H. analysed the results.
J.W., S.B. and P.W. performed the ICU clinician assessments and contributed to the clinical interpretation of the findings.
J.W. helped to design the ICU mortality analysis.
H.C.C., A.G., J.L., L.S., G.W., E.J.M.B., D.F., C.D. and J.N.K. provided clinical and methodological advice and contributed to the interpretation of the results.
T.H., T.B., R.W. and D.T. drafted the manuscript.
D.T. and T.H. jointly supervised the work.
All authors critically reviewed and revised the manuscript and approved the final version.

\noindent\textbf{Competing interests:}
E.J.M.B. reports research grants from Siemens Healthineers and Canon Medical Systems.
D.F. holds shares of and is an employee at Synagen.
J.N.K. holds shares in and serves as scientific advisor of StratifAI, Synagen, Spira Labs, Tremont AI and Saterra AI; is Co-PI on institutional research grants from GSK and AstraZeneca, and declares honoraria or consulting fees from AstraZeneca, Bayer, Bioptimus, Daiichi Sankyo, Eisai, Janssen, Merck, MSD, Novartis, BMS, Roche, and Pfizer.
D.T. holds shares in and serves as scientific advisor of StratifAI GmbH, Germany, and Synagen GmbH, Germany, and has received honoraria for lectures from AstraZeneca, Bayer, Boehringer Ingelheim, Debiopharm, GE HealthCare, Gilead, Ipsen, medtastic, MSD, Orthomol, Pfizer, Philips, and Roche.
The remaining authors declare no competing interests.

\noindent\textbf{Data, code, and materials availability:}
This study generated no new physical materials.
The clinical data used in this study are available through PhysioNet: MIMIC-III\cite{johnson2016mimiciii} (\url{https://physionet.org/content/mimiciii/}), MIMIC-IV\cite{johnson2023mimiciv} (\url{https://physionet.org/content/mimiciv/}), MIMIC-IV-ED\cite{johnson2023mimiced} (\url{https://physionet.org/content/mimic-iv-ed/}) and MIMIC-IV-Note\cite{johnson2023mimicnote} (\url{https://physionet.org/content/mimic-iv-note/}).
Access requires PhysioNet credentialing, completion of the required research training and acceptance of the applicable data-use agreements.
Access to patient-level records and derived clinical case materials remains subject to these agreements.
INSPECT is available from the Stanford Center for Artificial Intelligence in Medicine and Imaging (\url{https://inspect.stanford.edu}) under a data use agreement for non-commercial research\cite{huang2023inspect}.
Summary results supporting the findings are included in the article and its Supplementary Materials.
Code for cohort construction and data processing will be released upon publication.
The core source code for Decision2Care, including diagnosis, treatment selection and mortality prediction, is publicly available at \url{https://github.com/peterhan91/Decision2Care}.
A project website with recorded example work-ups and the main results is available at \url{https://peterhan91.github.io/Decision2Care/}.

\clearpage
\supplementstart
\setcounter{page}{1}
\noindent{\fontsize{14}{18}\selectfont\sffamily\bfseries\raggedright Supplementary Materials for\par}
\vspace{0.4em}
\noindent{\fontsize{20}{24}\selectfont\sffamily\bfseries\raggedright \papertitle\par}
\vspace{1.1em}
{\setstretch{1.15}\fontsize{13.5}{16.5}\selectfont\raggedright
\noindent
{\renewcommand{\textsuperscript}[1]{}\paperauthors}
\par}
\vspace{0.8em}
\noindent\textbf{Corresponding author:} \correspondingauthor

\vspace{1.2em}
\noindent\textbf{This PDF file includes:}

\noindent
Materials and Methods\\
Figs.~S1 to~S\suppfigures\\
Tables~S1 to~S\supptables

\clearpage
\section*{Materials and Methods}

\subsection*{Decision2Care architecture and diagnostic execution}

The following paragraphs describe Decision2Care's components and the flow of a diagnostic work-up in more detail.
The released code includes the wording of every question, prompt, and answer option.

\methodsanchor{meth:architecture}%
\textbf{Typed questions.}
In Decision2Care, a decision model scores the options of each clinical decision and a task-specific controller executes the decision.
Decision2Care makes decisions through typed questions, which are either Choices among named options or Nouls, yes/no questions about a single proposition.
Each request consists of the current clinical state $s$ and one or more typed questions.
The decision model returns a probability for each named answer and generates no text\cite{almeida2026jev,zhang2026cs1}.
A Choice is a question $q$ with $K$ mutually exclusive named options $o_1, \dots, o_K$, for which the model returns a probability for every option,
\begin{equation}
  p_k = P(o_k \mid s, q), \qquad k = 1, \dots, K, \qquad \sum_{k=1}^{K} p_k = 1 .
\end{equation}
A Noul is a single proposition $a$, for which the model returns the probability that it is true,
\begin{equation}
  p = P(a \mid s), \qquad 0 \leq p \leq 1 .
\end{equation}
For example, the Noul that selects a laboratory test reads `Should the laboratory test [name of the test] be ordered now to work up this patient?'.

\textbf{Controller.}
The controller applies fixed rules to the returned probabilities.
If a Choice selects an action, then the controller executes the most probable option,
\begin{equation}
  k^{*} = \operatorname*{arg\,max}_{k} \; p_k .
\end{equation}
If a Choice ranks the alternatives, then the controller goes through them in order of descending probability or keeps a fixed number of the most probable.
If the question is a Noul, then the controller endorses the proposition when $p$ reaches a threshold.
We used 0.50 for the Nouls that selected items or evaluated indications.
Multiple Nouls can select multiple items for a single action.
After each action, the controller adds the returned observations to the state.
Diagnosis, treatment, and prognosis all use the same interface and differ in their action menus, execution rules, and endpoints.
Questions sharing a state were submitted together in one request, with stable identifiers linking each answer to its question.

\methodsanchor{meth:models}%
\textbf{Decision models.}
We ran Decision2Care using Jev 1.13.0\cite{almeida2026jev} (accessed via an account with no data retention) as the main decision model.
The same diagnostic work-up (same controller, same settings) was also run with two locally hosted open-source models (\supptabref{tab:supp-decision-models}).
PPLX decider v1.2 ran in bfloat16 (BF16)\cite{perplexity2026decider}.
Qwen3.8-Flash-Next ran in NVIDIA's 4-bit floating-point format (NVFP4)\cite{nvidia2026qwen}.
We ran local PPLX and Qwen inference on an NVIDIA DGX Spark with the GB10 graphics processing unit.
PPLX is a decision model with its own readout.
Qwen is a general-purpose LLM which returns a probability for the next token rather than for an option.
Its readout therefore adapted the answer-token approach of Zhang et~al.\cite{zhang2026cs1}.
We assessed each question in isolation with a prompt that included the current state, the question, and all options (each under a single-token letter code).
The log probabilities $\ell_1, \dots, \ell_K$ that the model assigned to these letter codes as the next token were normalised over the options,
\begin{equation}
  p_k = \frac{\exp(\ell_k / T)}{\sum_{j=1}^{K} \exp(\ell_j / T)} ,
\end{equation}
with $T = 1$ in all experiments.
A Noul was presented as a question with two options (false under A, true under B) and the normalised probability of B was returned.
For questions with more than 128 options, we read the log probabilities in multiple requests with the same prompt and then normalised them together.
The options kept the order in which the controller supplied them, thinking was turned off and the probabilities were uncalibrated.
For the model comparison, we ran the work-up 5 times for each of the 3 models: Jev 1.13.0, local PPLX, and Qwen3.8.
All the experiments used the existing model checkpoints and their native probability scales.

\textbf{Clinical state and tools.}
For diagnosis, the state was a structured record containing the presenting history, recorded sex, retrieved findings and a chronological work-up log.
Separate fields held the physical examination, vital signs, laboratory results, microbiology, imaging and home medications.
The primary configuration did not provide age and chief complaint because age is censored in the MIMIC notes, and the triage text can name a transfer diagnosis.
Each action other than the final diagnosis was a service that a hospital already defines: an examination, a review of the vital signs, an order for laboratory tests (including microbiology) or imaging, or a review of the home medication list.
These services were implemented as tools (the first 4 are shown in \figref[D]{fig:overview}), each of which revealed the corresponding existing record and added it to the clinical state.
Laboratory observations included values, reference ranges and abnormality flags, and imaging tools returned report text.
If a requested imaging study was absent from the record, then the tool returned an availability message.
Across 5 runs on 546 admissions, a report was on record for 72\% of CT, 63\% of ultrasound, and 15\% of MRI requests.
The tool layer tracked requested items and exhausted tools to construct the next action menu.

\textbf{Diagnostic work-up.}
At each round, a Choice first selected among the remaining tools and the final diagnosis (the routing Choice).
Tool-specific questions then selected examination systems, vital-sign items, laboratory groups, and imaging studies.
We retrieved physical examination and vital-sign items with Noul scores of at least 0.50.
Laboratory groups were considered one by one (in order of descending Choice probability), and the first group containing endorsed items supplied the order.
Panels and individual tests were selected at 0.50, with redundant panels removed and limits of 12 panels or 40 items per order.
In the primary configuration, the routing Choice was asked again after each laboratory order.
For imaging, a Choice selected one of the remaining studies or a stop-imaging option, which removed imaging from the later rounds.
The diagnostic loop ended when final diagnosis was selected, the remaining action menu contained only final diagnosis, or 30 tool actions had been executed.
Alongside the routing Choice, a Noul asked whether the documented findings sufficed to name the principal diagnosis with high confidence.
Its probability was recorded without being acted on, so the routing Choice alone determined the executed action.
Across 5 runs on 546 admissions, a laboratory action ordered a median of 6 panels or tests, and there were 1.65 imaging requests per case.

\textbf{Coding the final diagnosis.}
After these rounds, we selected the final diagnosis by hierarchical search through ICD-10-CM chapters, blocks, 3-character categories, and billable codes.
We ranked choices at each level, with beam widths of 4 chapters, 4 blocks, and 5 categories.
The score of a path was the product of the conditional probabilities along it.
For each retained category, the billable codes were compared in a flat Choice, using chunks of up to 200 where required.
After ranking the codes within each category, the best code had to be selected across categories.
The code with the highest path score became the principal diagnosis unless it lay outside the highest-scoring category or the leading code of another category reached half of its path score.
In these cases, a final Choice among the leading codes of the retained categories selected the principal diagnosis.
This comparison was needed because the probability of one disease can be divided among several codes of a category.
Remaining codes retained their path-score order.

\textbf{Audit record.}
The saved work-up record links the questions, the returned probabilities, the rule the controller applied, the executed actions, the returned observations, and the final code.
From this record, the state at each decision can be reconstructed, which the analyses below rely on.

\subsection*{Study cohorts and information availability}

\methodsanchor{meth:cases}%
The study used retrospective, de-identified records from MIMIC-III and MIMIC-IV, both derived from Beth Israel Deaconess Medical Center\cite{johnson2016mimiciii,johnson2023mimiciv}.
MIMIC-IV discharge summaries and radiology reports came from MIMIC-IV-Note\cite{johnson2023mimicnote}, and ED data came from MIMIC-IV-ED\cite{johnson2023mimiced}.
The diagnostic dataset contained 873 admissions across 12 diseases: appendicitis, cholecystitis, diverticulitis, pancreatitis, pneumonia, PE, UTI, bowel obstruction, congestive heart failure, MI, meningitis and SAH.
It combined the 551 admissions in the MIRA-v2 benchmark, previously described\cite {zhang2026onprem}, with 322 admissions for 5 additional diseases (bowel obstruction, congestive heart failure, MI, meningitis, and SAH).
We selected the additional admissions from MIMIC-IV following the selection rules and the sampling rate of MIRA\cite{ferber2026mira} and removed those we could not evaluate (for example, MI without a troponin measurement).
Following the MIMIC-IV-Ext Clinical Decision Making (MIMIC-CDM) benchmark\cite{hager2024cdm}, which excludes histories that disclose the diagnosis, we retained the 548 admissions whose history held no such mask.
We removed two because MIRA's dataset builder could not build a case from them under its own rules.
The paired analysis therefore included 546 admissions, 437 from the MIRA-v2 benchmark and 109 with the additional diseases.

In the diagnostic experiments, each system requested the parts of a retrospective record one after another.
Decision2Care and MIRA-v2 differed in terms of how the presenting history reached them.
Following MIMIC-CDM \cite{hager2024cdm}, Decision2Care received the recorded presenting history directly, whereas MIRA-v2's physician agent received history from its patient agent, whose instructions restricted disclosure of the ED course.
The history supplied to Decision2Care contained ED course information in 164 of the 546 admissions.
As released, the dataset builder of MIRA serves radiology reports in which the disease name is not masked.
We therefore ran the comparator again with the name masked in every served text (the masked reruns), matching the masked diagnostic records used by Decision2Care.
We used the reference disease label for each admission and its discharge codes to score the answers.

Prognostic experiments used 8 disease cohorts in each MIMIC database: ICH, TBI, sepsis, AKI, PE, ARDS candidates, CAP and matched polytrauma.
ARDS cohort reflected the implemented clinical screening criteria, and we matched polytrauma cohorts between databases.
Cohort sizes and horizon-specific outcomes are reported in \supptabrefs{tab:supp-prognosis-cohorts}{tab:supp-icu-hourly}.
We offered only records available at the time of prediction, 8~h after arrival in the ED or 1 to 12~h after admission to the ICU.
We included an admission in each prognostic cohort it belonged to, so cohorts are counted per entry.
The ED prediction was also run on INSPECT, a data set from Stanford Health Care\cite{huang2023inspect}, as described under outcome prediction.

\subsection*{Diagnostic accuracy, time and cost}

\methodsanchor{meth:scoring}%
We compared Decision2Care with MIRA-v2 on the same 546 admissions, using GPT-OSS-120B, GLM-5 or Claude Sonnet 5 as the underlying LLM of MIRA-v2.
We ran MIRA-v2 using its released code\cite{zhang2026onpremcode} on Amazon Bedrock, with its prompts and tools unchanged.
We ran each system 5 times under its fixed configuration because repeated requests do not return identical answers from Jev or the LLMs.
Unless stated otherwise, single-run analyses of Decision2Care used its first run (the baseline run), which also had the highest diagnostic accuracy of the 5 (87.0\%, compared with 85.9--86.6\%).
Decision2Care's principal diagnosis was scored by the strict MIMIC-CDM disease-matching rule\cite{hager2024cdm,zhang2026onprem}, applied to the title of the selected code.
MIRA-v2's free-text answer was scored by the judge of its released code, with Claude Opus 5.5 replacing its original LLM and with matching criteria added for the 5 additional diseases.
This judge agreed with the rule on all 546 answers of Decision2Care in the baseline run.
The reference was the diagnosis recorded for the admission, which can reflect information that became available later.
We reported mean diagnostic accuracy across runs.
In the 437 admissions of the MIRA-v2 benchmark and the 109 added admissions, mean accuracy was 87.3\% and 83.3\% for Decision2Care, 86.5\% and 84.4\% for MIRA-v2 with GPT-OSS-120B, 90.8\% and 86.4\% with GLM-5 and 92.8\% and 87.3\% with Claude Sonnet 5.
For the paired McNemar tests, we classified each admission as correctly diagnosed by a system if at least 3 of its 5 runs were correct, following the majority-vote rule in the released MIRA-v2 evaluation code\cite{zhang2026onpremcode}.
For differential accuracy, Decision2Care contributed 1 entry per 3-character ICD-10-CM category and MIRA-v2 supplied ranked diagnoses after the encounter, and every entry was scored under the strict rule.
Where the first 5 entries of a list held no credited disease, additional queries extended it to 10 entries (\suppfigref{fig:supp-ddx-topn}).
Test-selection analyses matched each system's requests to the same recorded tests (\supptabref{tab:supp-test-selection}).

\methodsanchor{meth:physicians}%
As a reference for the accuracy of physicians, we used the results published for the 4 board-certified physicians of the MIRA study\cite{ferber2026mira}.
Each of them had worked up a random, non-overlapping share of 311 benchmark admissions, and the judge of MIRA had scored their free-text diagnoses.
We left out the 21 admissions with pancreatic cancer, which MIRA-v2 does not include.
The admissions that the physicians read have not been published, so the 2 results could not be compared admission by admission.
We therefore weighted Decision2Care's mean accuracy in each disease by the number of admissions that the physicians had read in that disease (\supptabref{tab:supp-physicians}).
The interval for the physicians used Wilson's method, and that for Decision2Care came from 20,000 resamples of admissions within each disease.
The physicians took the history from MIRA's patient agent, whose instructions restricted disclosure of the ED course.
In the 321 benchmark admissions whose recorded history held no ED course, Decision2Care's weighted accuracy was 84.6\% (95\% CI 79.1--89.7\%).

\methodsanchor{meth:timing}%
We measured runtime from the recorded execution of the diagnostic workflow through final code selection.
We timed Decision2Care in one of its 5 runs, in which every request was sent to the model anew.
Comparator medians pooled their 5 masked runs.
Factoring in model serving and workflow steps, these times reflect each complete system as run.
We calculated token-based costs from logged token usage using the list prices recorded for the experiments.
Decision2Care costs included the post-work-up confidence request.
The comparator's cost per run was its mean over 4 uninterrupted runs per model as released, because only those logs provided complete usage accounting, and its cost for 5 runs was 5 times this mean.
The diagnostic and timing comparisons used the masked reruns.
Judge costs were tabulated separately.
\methodsanchor{meth:variants}%
For sensitivity analyses, we ran the same work-up on the same paired cohort with variants of the diagnostic instruction wording, option order, and state presentation (\supptabref{tab:supp-robustness-variants}).

\subsection*{Monitoring the differential diagnosis}

\methodsanchor{meth:traces}%
We reconstructed the states of 546 admissions across 5 runs, beginning with the initial state, in which no action had been taken, and continuing with the state after every action.
At each state, an auxiliary Choice compared the 12 benchmark diseases plus ``other'' as a reading of the differential, and the hierarchical scorer generated an ICD-10-CM diagnosis.
During the run, the controller acted on the probabilities of the routing Choice.
However, in order to show how the diagnosis developed, these readings were made separately on the reconstructed states after the run.
When a work-up ended before the last plotted step, its final state was carried forward.

\methodsanchor{meth:guideline}%
The decisive step was the action producing the largest increase in the probability of the true disease, with ties assigned to the earliest action.
Guideline alignment was assessed for increases of at least 0.20 using fixed disease-specific test mappings (\supptabrefs{tab:supp-guideline-tests}{tab:supp-decisive-steps}).
The cited source documents defined these operational mappings.
The reference for an action selected at random from the same case was the mean fraction of actions in each case run mapped to a guideline test.

\methodsanchor{meth:labsplit}%
\methodsanchor{meth:leaveout}%
For each test category in the laboratory decomposition, the new results were added separately to the same pre-order state, followed by a fresh differential query.
The source-by-disease heatmap required at least 25 contributing actions across the 5 runs per displayed cell.
Source-omission analyses applied the same diagnostic scoring but started from completed charts rebuilt after removing one source.
We calculated changes within cases containing the corresponding source, with cells requiring at least 5 cases.
For each of the 4 errors retained at a confidence of at least 0.70, we coded the diagnosis and asked the confidence question at every stage of the recorded work-up. 
The omitted source was the imaging modality of the report at which confidence rose most.
In the retained error without a laboratory order, we added the recorded pancreatic enzymes to the final chart before scoring and repeated the confidence question.

\methodsanchor{meth:showcase}%
We selected illustrative cases from trajectories that began with an incorrect top-ranked disease at model probability $\geq 0.50$ and changed to the correct disease at a guideline-mapped test.
We ranked these trajectories by the size of the change in the differential and by the length and variety of the work-up, and selected the highest-ranked case for each disease.
The selected cases had correct final diagnoses across all 5 runs.
Along with these cases, one further baseline work-up is shown for each of the 9 remaining diseases, each with at least 5 executed retrievals (\hyperref[fig:supp-diagnosis-trajectories]{\textbf{figs.~\ref*{fig:supp-diagnosis-trajectories} to~\ref*{fig:supp-diagnosis-trajectories-3}}}).

\subsection*{Confidence and deferral to a clinician}

\methodsanchor{meth:signoff}%
After the work-up, the decision model answered one more question, a Noul on whether the agent's proposed diagnosis was the admission's principal diagnosis, given its title and the retrieved chart.
Its probability measures confidence in the diagnosis, an answer-conditioned self-evaluation described previously\cite{kadavath2022know}.
We evaluated this confidence score against diagnostic correctness using the area under the receiver operating characteristic curve (AUROC) and accuracy at different retained fractions.
Coverage was the fraction of cases retained without review, and retained accuracy was the fraction of those diagnoses that were correct.
The single-run analysis used the baseline run, and the results of each of the 5 runs are given in \supptabref{tab:supp-deferral}.

\methodsanchor{meth:consistency}%
Comparator confidence was ConsistencyDx, a measure of the agreement among sentence embeddings of 5 diagnostic answers, computed with the MIRA-v2 implementation\cite{zhang2026onprem}.
Diagnostic correctness followed the majority verdict.
The same implementation was applied to the 5 answers of Decision2Care.
Threshold sweeps measured coverage, retained accuracy, errors sent to review and the number of cases sent to review (\supptabref{tab:supp-deferral}).
We simulated sending a case to review retrospectively, with correctness defined by the recorded reference diagnosis.

\methodsanchor{meth:stress}%
To test whether confidence followed the evidence, we ran the complete work-up 5 more times with the presenting history withheld and 5 more times with the confirming results of each disease made normal.
For the latter, we replaced imaging reports of the affected body region with a normal report of the same study type from a fixed template, and we replaced the confirming laboratory results (for example, lipase and amylase in pancreatitis) with normal values without an abnormality flag.
We then assessed confidence on each rerun's final chart (\supptabref{tab:supp-stress}).

\subsection*{Treatment selection and evaluation}

\methodsanchor{meth:treatment}%
The treatment controller used the baseline retrieved chart, the agent's principal diagnosis, the allergy list and kidney function.
Allergy information came from the discharge-note allergy field, and renal assessment used the first creatinine measurement and estimated glomerular filtration rate calculated with the 2021 Chronic Kidney Disease Epidemiology Collaboration creatinine equation\cite{inker2021egfr}.
First, Nouls assessed 42 medication categories and 100 ICD-9-CM procedure groups at a threshold of 0.50.
Next, for the selected medication categories, Choices selected regimens from combinations observed in first-day MIMIC-IV orders, drug-specific doses and actions on home medications.
Therapeutic anticoagulation and prophylaxis had separate dose menus.
Order-level checks then assessed allergy, renal dosing, bleeding contraindications and duplication with continued home medication.
Flagged regimens needed correction, so the controller ran up to 2 replacement rounds, removing the implicated drugs from the candidate menu.
Procedure selection ranked codes within endorsed groups, assessed individual indications and checked whether additional operations were required.

For evidence-dependence analyses, we applied the same indication questions but now after removing a chart source, the explicit diagnosis, allergy information or kidney function in separate variants.
Source-omission comparisons retained the diagnosis and used cases containing that source.
History-only variants retained allergy and kidney information and were evaluated before and after adding the final diagnosis.
Summaries included decisions originally positive at 0.50, with at least 20 positive decisions per row and 3 cases per source-omission cell.
When the same questions were asked again on unchanged states, the probability of a positive decision changed by 0.012 on average, and 73 of the 3,875 positive decisions crossed 0.50.

\methodsanchor{meth:rubric}%
\Figref[B]{fig:treatment-selection} shows the 475 admissions that Decision2Care diagnosed correctly in the baseline run.
Comparative rubric analyses used the 437 of these admissions that every system diagnosed correctly (a different set from the 437 MIRA-v2 admissions), combining the correctness of Decision2Care in the baseline run with each comparator's majority verdict.
Comparator treatment selection was averaged over its correctly diagnosed runs.
Indicated-treatment rubrics extended the MIMIC-CDM evaluators\cite{hager2024cdm} with disease-specific guideline items.
Hospital-performed procedures defined the corresponding retrospective procedure indications (\supptabref{tab:supp-treatment-indicated}).
Extension rubrics drew on the Bologna adhesive small-bowel obstruction recommendations\cite{tenbroek2018bologna}, the 2022 heart-failure guideline\cite{heidenreich2022hf}, the 2025 acute-coronary-syndrome guideline\cite{rao2025acs}, the European Society of Clinical Microbiology and Infectious Diseases (ESCMID) bacterial-meningitis guideline\cite{vandebeek2016meningitis} and the 2023 aneurysmal-subarachnoid-haemorrhage guideline\cite{hoh2023sah}.
Inherited benchmark conventions, including antibiotic indication for every diverticulitis case, were retained for comparability, although the American Society of Colon and Rectal Surgeons (ASCRS) guideline permits selected uncomplicated cases to be treated without antibiotics\cite{hall2020diverticulitis}.

\methodsanchor{meth:judge}%
The categories in \supptabref{tab:supp-guideline-categories} were assessed using the MIRA evaluation format\cite{ferber2026mira} in 355 admissions across 7 diseases.
MIRA Supplementary Data Table 52 supplied the appendicitis, cholecystitis, pancreatitis and UTI texts, the latter based on Al Lawati et~al.\cite{allawati2024uti}.
Pneumonia used the joint American Thoracic Society (ATS) and Infectious Diseases Society of America (IDSA) guideline\cite{metlay2019pneumonia}, and diverticulitis and PE used investigator-written summaries.
The evaluator received the guideline text, the categories, the principal diagnosis and the complete list of orders, and returned a verdict for each category.
Orders from every source were written in the same format with no sign of their origin.
The underlying LLM of the evaluator was Claude Opus 5.5 (run on Amazon Bedrock at low effort) in place of the GPT-4o of the original\cite{ferber2026mira}.
Drug-class agreement used first-day prescriptions and ED dispensing in 435 admissions with recorded care.
Procedure comparisons (\supptabref{tab:supp-treatment-measures}) used exact ICD-9-CM matches, 3-digit-category matches, precision, request counts and Tversky distance\cite{tversky1977similarity}, weighting unmatched requests twice and missed procedures once.
These endpoints measured rubric coverage, model-assessed guideline adherence and agreement with recorded care.

\subsection*{Outcome prediction}

\methodsanchor{meth:prognosis}%
ED predictions were made 8~h after arrival, with the cohort disease supplied as known context.
Eligible observations had event and availability times at or before the prediction cutoff, and tools grouped histories, vital-sign series, individual laboratory tests and specimens, urine observations and radiology reports.
At each selection round, a Choice ranked the remaining retrieval tools and, after the first retrieval, a finish option.
The selected tool then appended its original records to the state.
After these rounds (which ended on finish, tool exhaustion or a 20-round selection limit), a final Choice assigned probabilities to death within 7 days, days 8--30, days 31--90 or survival through 90 days, with horizons measured from the prediction time.
The 30-day score summed the first 2 probabilities and the 90-day score summed the first 3.
Outcomes were assigned from recorded mortality information, with eligibility determined separately for each horizon by outcome availability.
Each fixed cohort was evaluated over 5 runs.
Pooled analyses concatenated disease-task entries within each run, computed its AUROC and averaged the run-wise receiver operating characteristic curves on a common grid.
Reported ranges describe variation across the 5 runs (\supptabref{tab:supp-prognosis-cohorts}).

\methodsanchor{meth:secondcentre}%
We repeated the ED prediction on INSPECT, a de-identified data set of 23,248 CT pulmonary angiography examinations of 19,402 adults at Stanford Health Care between 2000 and 2021, with report impressions and structured patient records\cite {huang2023inspect}.
Eligible encounters were ED visits with the impression of a CT pulmonary angiogram dated within 8~h of arrival.
PE was identified from the impression by a text rule that accepted only a definite, present embolus.
The rule was applied without knowledge of the outcome.
Pneumonia candidates met the report rule and the diagnosis codes of the MIMIC pneumonia cohorts at the same encounter.
For each patient the first eligible encounter was used, and patients who had died by the prediction time were excluded, resulting in 590 patients with PE and 87 pneumonia candidates.
The records offered to the agent were age and sex, vital signs, laboratory and urine results and the impression, each available from its recorded time.
INSPECT holds no presenting history or clinical note.
Death was taken from the recorded date of death, and a patient counted as alive at a horizon only when follow-up was observed beyond it.
We excluded patients with unknown status at that horizon (35 and 48 with PE and 9 and 11 pneumonia candidates at 30 and 90 days).
We applied the agent, its tools, and its questions unchanged from the MIMIC cohorts, and evaluated each cohort over 5 runs.
Intervals came from 2,000 bootstrap replicates in which patients were resampled within outcome and the AUROCs of the 5 runs were averaged.
For differences between cohorts, we resampled the 2 cohorts independently.

\methodsanchor{meth:icu}%
The ICU analysis included first ICU admissions from encounters whose patients remained alive and hospitalised at 24~h, the point at which the severity scores were computed.
Across the 8 disease cohorts and 2 databases, the analysis comprised 5,999 case entries and 1,026 in-hospital deaths (\supptabref{tab:supp-icu-hourly}).
The endpoint was all-cause death before hospital discharge.
The retrieval was run independently for each cumulative window, from 1 to 12~h after ICU admission, each time starting with no dynamic observations in the initial state.
Age and recorded sex were retained as static context, and dynamic observations came from the corresponding ICU window.
The controller selected available records through its prognosis-specific tools, then returned a binary mortality probability.
The worked example also used a 24-h window.
APS III\cite{knaus1991apache}, SAPS II\cite{legall1993saps} and OASIS\cite{johnson2013oasis} were computed from the first 24~h in the ICU using the MIMIC code implementations\cite{johnson2018mimiccode}.
Within each disease, predictions from the 2 databases were concatenated before AUROC calculation.
Paired comparisons used 2,000 patient-bootstrap replicates stratified by database and outcome, with shared resamples across hours and clinical scores (\supptabref{tab:supp-icu-scores}).
The first hour reaching the highest clinical-score AUROC was recorded descriptively.
Retrieval efficiency was summarised by tool calls and the fraction of eligible records read at 12~h.

\subsection*{ICU clinician prediction and retrieval review}

\methodsanchor{meth:readers}%
Three ICU clinicians independently assessed 72 MIMIC cases in 2 stages, producing 216 assessments.
S.B. and P.W. were board-certified anaesthesiologists with 6 and 11.5 years since full medical licensure and 1.5 and 2.5 years working exclusively in the ICU, respectively.
J.W. was an anaesthesiology resident with 3.5 years since full medical licensure and 1.5 years working exclusively in the ICU.

Sampling was stratified by disease, database, 90-day outcome and correctness of the saved agent prediction.
Wrong predictions of 90-day mortality therefore made up 33 of the 72 review cases (46\%), compared with 2,247 of the 8,570 cohort entries (26\%).
The ICU clinicians were told the cohort disease and were given every record offered to the agent, that is, all records available by 8~h after ED arrival (a median of 124 per case).
The records were shown on 1 page with a timeline of values and no time limit; for MIMIC-III, which has no reference ranges, the ranges were taken from the same tests in MIMIC-IV.
The agent had read a median of 54\% of the records of a case.
In 8 cases, a formatting error caused 11 retrieved laboratory results to reach the agent without their numerical values.
The ICU clinicians saw the true values, which were marked.
In stage 1, the ICU clinicians were blinded to the agent's output and the patient's actual outcome.
They selected a survival interval, estimated 30- and 90-day mortality probabilities, rated certainty, identified essential records using a disease-specific checklist and recorded information gaps.
In stage 2, they viewed the agent's retrieval sequence, action probabilities and prediction while remaining blinded to the patient's actual outcome.
They assessed essential records left unread, unnecessary retrieval steps, the plausibility of the prediction and agreement with the predicted interval, and could record a safety concern.
Plausibility was rated from 1 to 5, with 1 labelled `implausible given the data', 3 `defensible, I would differ' and 5 `fully consistent with the data' (2 and 4 were unlabelled).
A rating of 3 or more was counted as at least defensible and a rating of 4 or 5 as plausible.

We derived sensitivity and specificity from each ICU clinician's interval prediction, and we used AUROC based on their stated probabilities.
The agent comparison used 1 saved run on the same cases.
Paired case bootstrapping with 2,000 resamples estimated differences between the agent and each ICU clinician, and between the agent and the ICU clinicians' mean probability (\supptabrefs{tab:supp-readers}{tab:supp-readers-review}).
As a sensitivity analysis, each review case was weighted by the number of cohort entries in its stratum divided by the number of review cases drawn from that stratum, and the AUROCs and their differences were computed with these weights.
The weights were concentrated (effective sample size 14.6), and 77 of the 8,570 cohort entries lay in 2 strata from which no case had been drawn.
A prediction of 90-day mortality was counted as wrong when the interval prediction did not agree with the outcome at 90 days.

\subsection*{Statistical analysis}

Bootstrap analyses resampled admissions, retaining each admission's runs and steps together.
Percentile 95\% CIs used 2,000 resamples for diagnostic accuracy and treatment, 400 for trajectories, 1,000 for confidence AUROC and 5,000 for model substitution.
For the mortality analyses, the resampling is described under outcome prediction and under the ICU clinicians' predictions.
Run ranges describe execution variability, and CIs describe sampling uncertainty conditional on saved outputs.
Coverage and retained-accuracy intervals used Wilson's method.
Paired categorical comparisons used two-sided exact McNemar tests, and continuous treatment comparisons used two-sided Wilcoxon signed-rank tests.
Holm correction\cite{holm1979multiple} covered disease-level diagnostic and robustness comparisons.
Benjamini--Hochberg correction\cite{benjamini1995fdr} covered indicated-treatment and guideline-category families, with documented-care comparisons corrected separately.
Benjamini--Hochberg-adjusted $p$ values are reported as $q$ values.
Other tests used nominal $p$ values.
Robustness differences used paired case-wise standard errors.

\subsection*{Ethics statement}

This study analysed retrospective, de-identified clinical records from MIMIC-III and MIMIC-IV, including MIMIC-IV-ED and MIMIC-IV-Note, and from INSPECT.
Patient information from MIMIC was processed in accordance with PhysioNet's `Credentialed Health Data Use Agreement' and its statement `Use of MIMIC Data with Large Language Models and Online Services'.
The institutional review boards of Beth Israel Deaconess Medical Center and the Massachusetts Institute of Technology approved the original MIMIC-III project, with a waiver of individual patient consent.
The institutional review board of Beth Israel Deaconess Medical Center approved the collection and sharing of MIMIC-IV and its associated resources and waived the requirement for informed consent.
The Stanford University Institutional Review Board approved the release of INSPECT, and the study that assembled it included a waiver of consent\cite{huang2023inspect}.
The analyses were retrospective and did not influence patient care.

\clearpage
\begingroup
\normalsize
\setstretch{1.0}
\setlength{\parskip}{0.75em}
\setlength{\parindent}{0pt}

\renewcommand{\figurename}{Fig.}
\renewcommand{\tablename}{Table}
\captionsetup[table]{labelfont=bf}
\renewcommand{\thefigure}{S\arabic{figure}}
\renewcommand{\thetable}{S\arabic{table}}
\setcounter{figure}{0}
\setcounter{table}{0}

\begin{figure}[H]
    \centering
    \includegraphics[width=\textwidth]{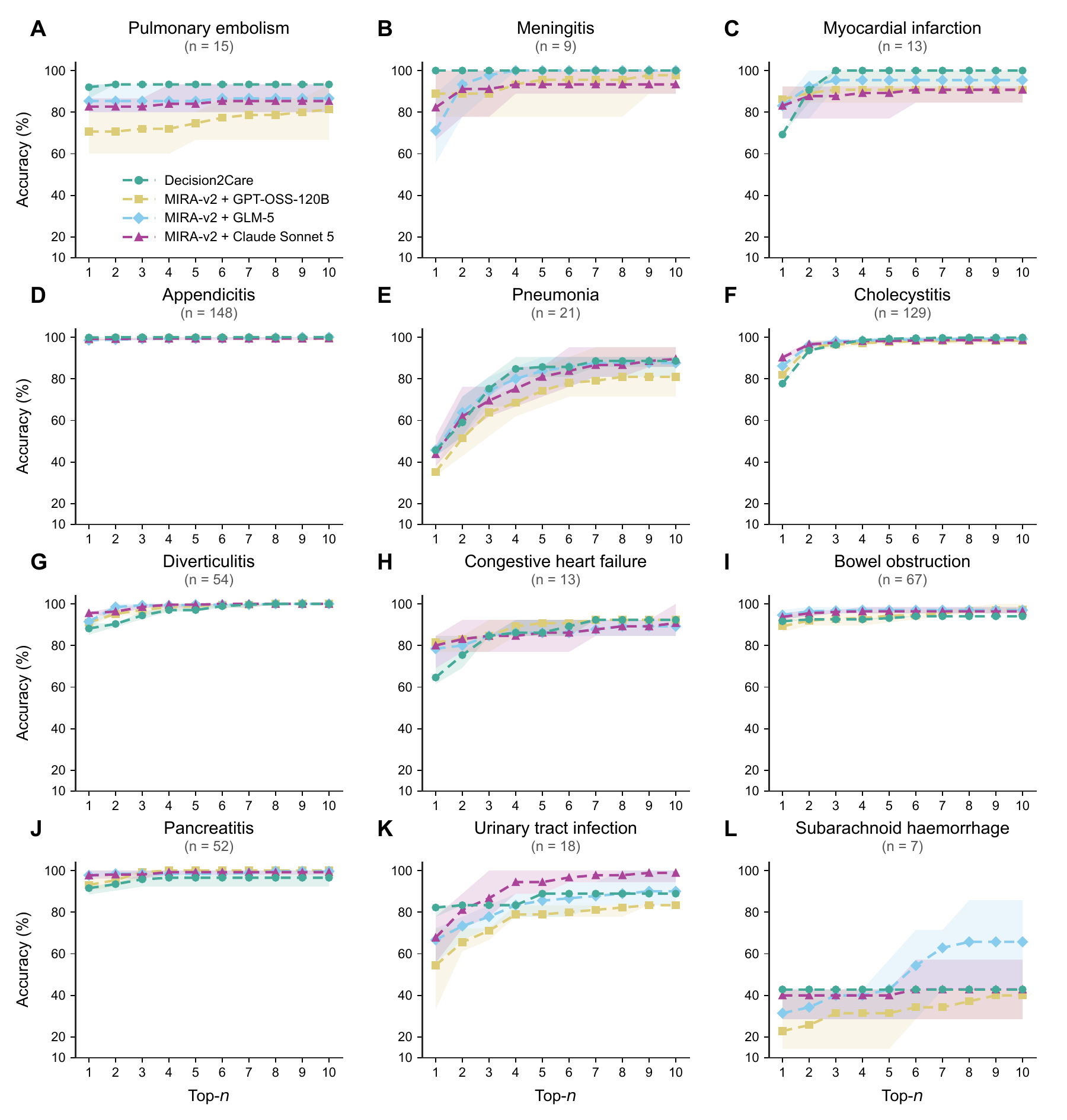}
    \caption{\textbf{Top-$n$ accuracy of the final differential by disease.}
    \textbf{A}--\textbf{L}, Strict disease-level accuracy for $n = 1$--10 on the 546 paired admissions, with the number of admissions per disease; lines, mean of 5 runs; shading, range of runs.
    Panels are ordered by Decision2Care's mean lead over the strongest comparator across $n$.
    In cholecystitis (\textbf{F}), the disease was among the first 5 entries of the differential in 139 of Decision2Care's 144 incorrect predictions (96.5\%).}
    \label{fig:supp-ddx-topn}
\end{figure}
\clearpage

\begin{figure}[p]
    \centering
    \includegraphics[width=\textwidth]{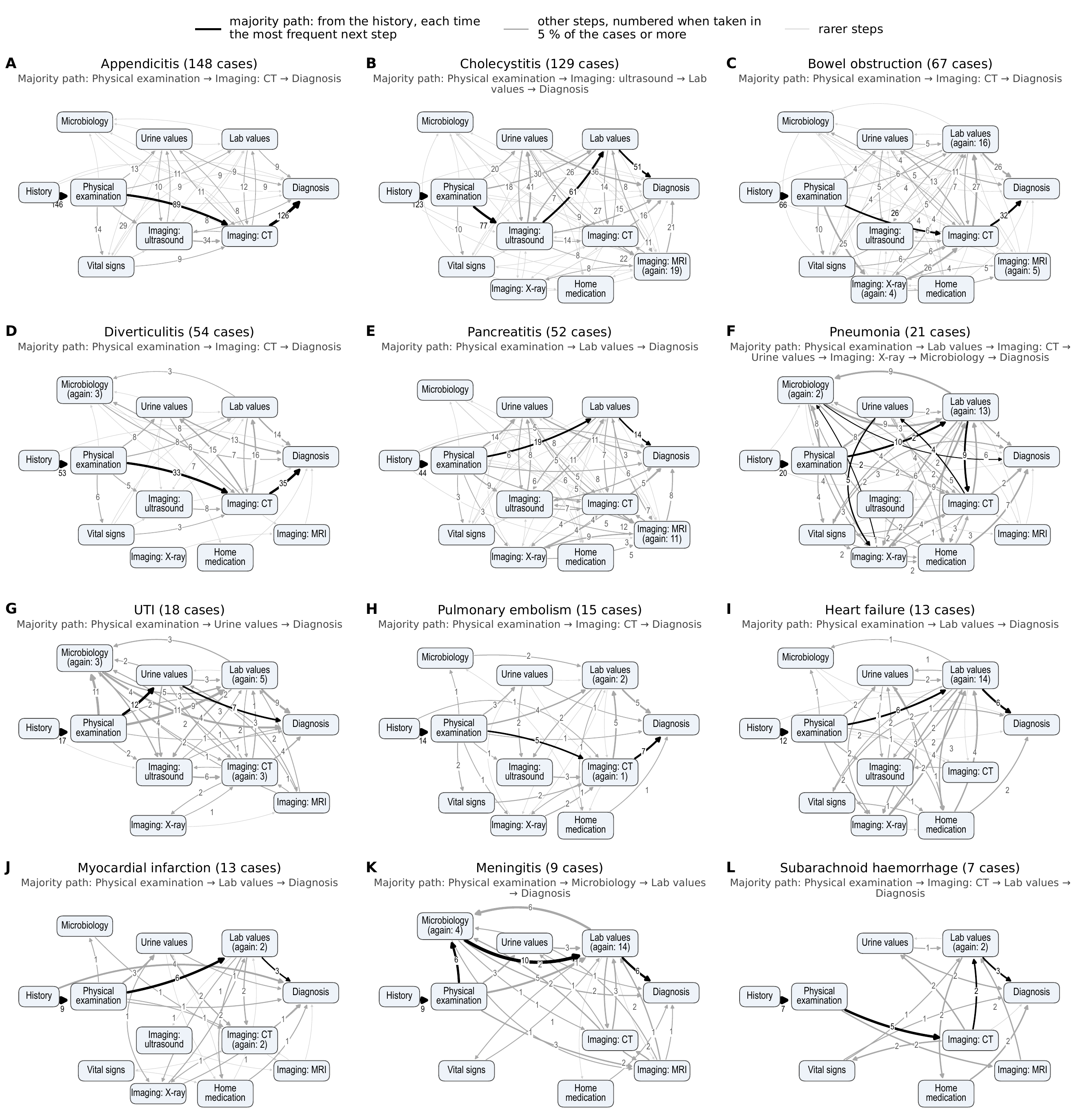}
    \caption{\textbf{Work-up paths by disease.}
    \textbf{A}--\textbf{L}, Mean transitions between information sources over Decision2Care's 5 runs on the 546 paired admissions, with the number of admissions per disease.
    Black arrows follow the most frequent next unvisited source; grey arrows mark steps taken in fewer than 5\% of cases; repeat requests are counted inside the boxes.
    }
    \label{fig:supp-workup-paths}
\end{figure}
\clearpage

\begin{figure}[H]
    \centering
    \includegraphics[width=\textwidth]{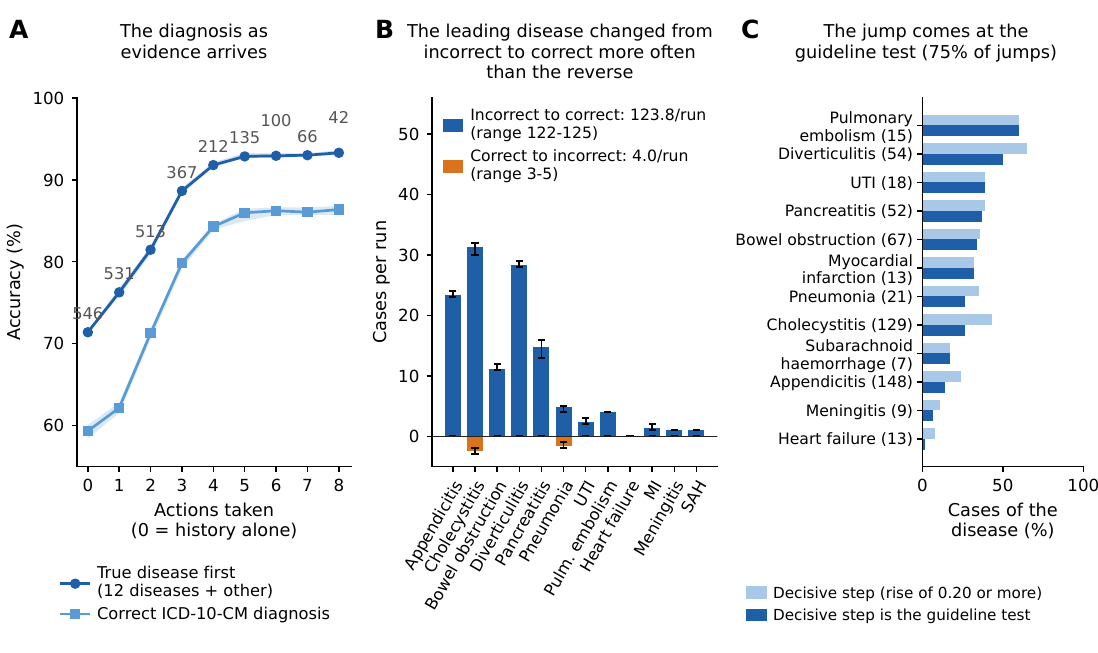}
    \caption{\textbf{The differential over the work-up.}
    Decision2Care's 5 runs on the 546 paired admissions; lines, means; bands, range of runs.
    \textbf{A}, Share of case-runs with the true disease first in the differential (circles) and with a correct coded diagnosis (squares) after each action; numbers, cases still acting.
    Where the history alone did not put the true disease first, the differential favoured none of the 12 diseases in 17.3\% of case-runs, another disease of the same organ system in 10.5\% and a disease of another organ system in 0.7\%.
    \textbf{B}, Cases per run whose top-ranked disease changed from incorrect to correct (blue) and from correct to incorrect (orange).
    \textbf{C}, By disease, the share of case-runs with a decisive step, a rise of at least 0.20 in the probability of the true disease (light), and the share whose decisive step was a guideline test (dark).
    }
    \label{fig:supp-differential}
\end{figure}
\clearpage

\begin{figure}[H]
    \centering
    \includegraphics[width=\textwidth]{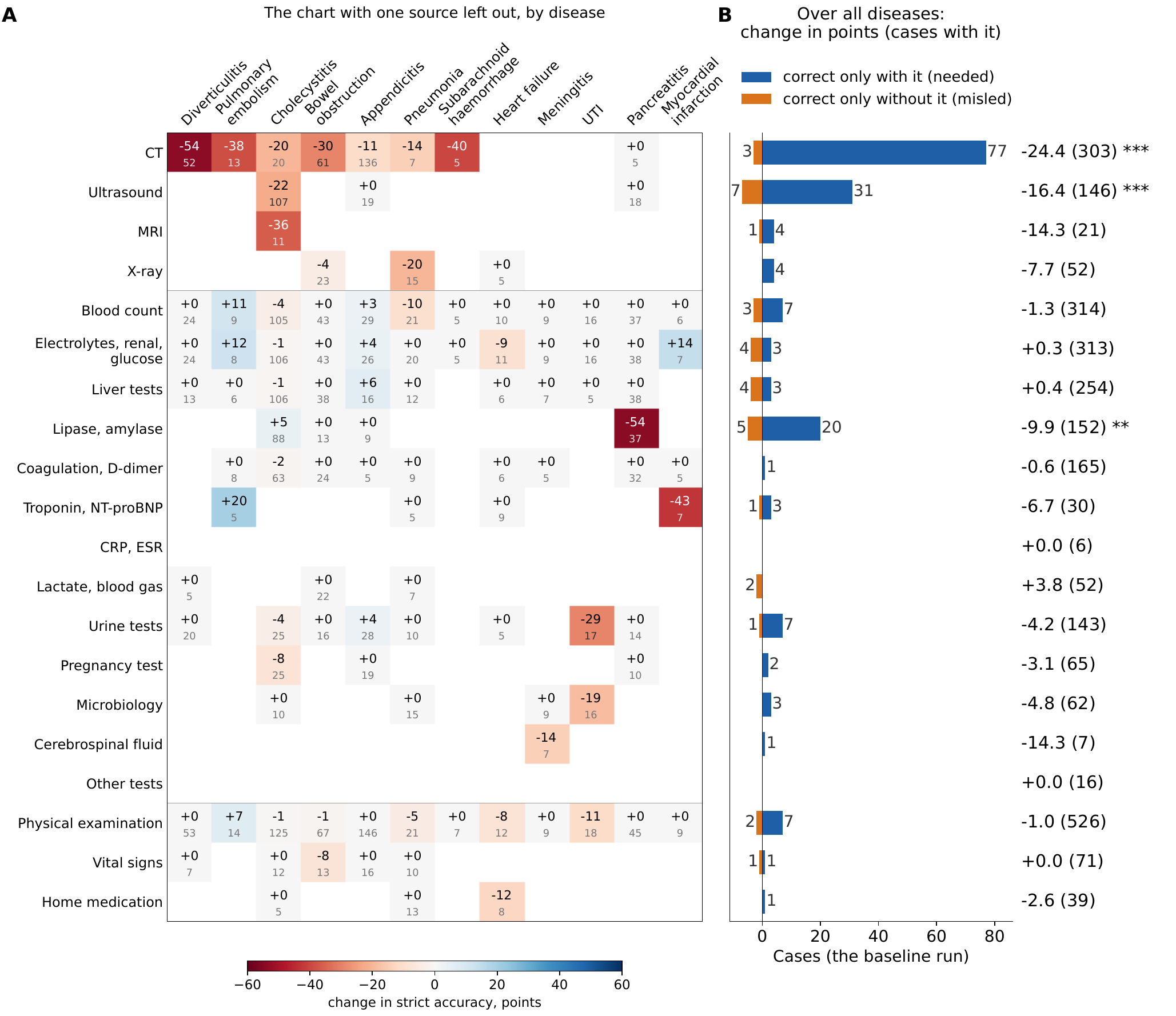}
    \caption{\textbf{The diagnosis made again with one source left out of the chart.}
    Decision2Care's baseline run on the 546 paired admissions.
    \textbf{A}, Change in strict accuracy, in points, by disease (columns) and withheld source (rows), among cases containing that information source; cells require at least 5 cases, with the number printed below each value.
    \textbf{B}, Over all diseases, the cases correct only with the source (blue) and only without it (orange), the change in points, the number of cases and the exact McNemar test (* $p < 0.05$, ** $p < 0.01$, *** $p < 0.001$).
    }
    \label{fig:supp-ablation}
\end{figure}

\clearpage

\begin{figure}[p]
    \centering
    \includegraphics[width=\textwidth,height=0.999\textheight,keepaspectratio]{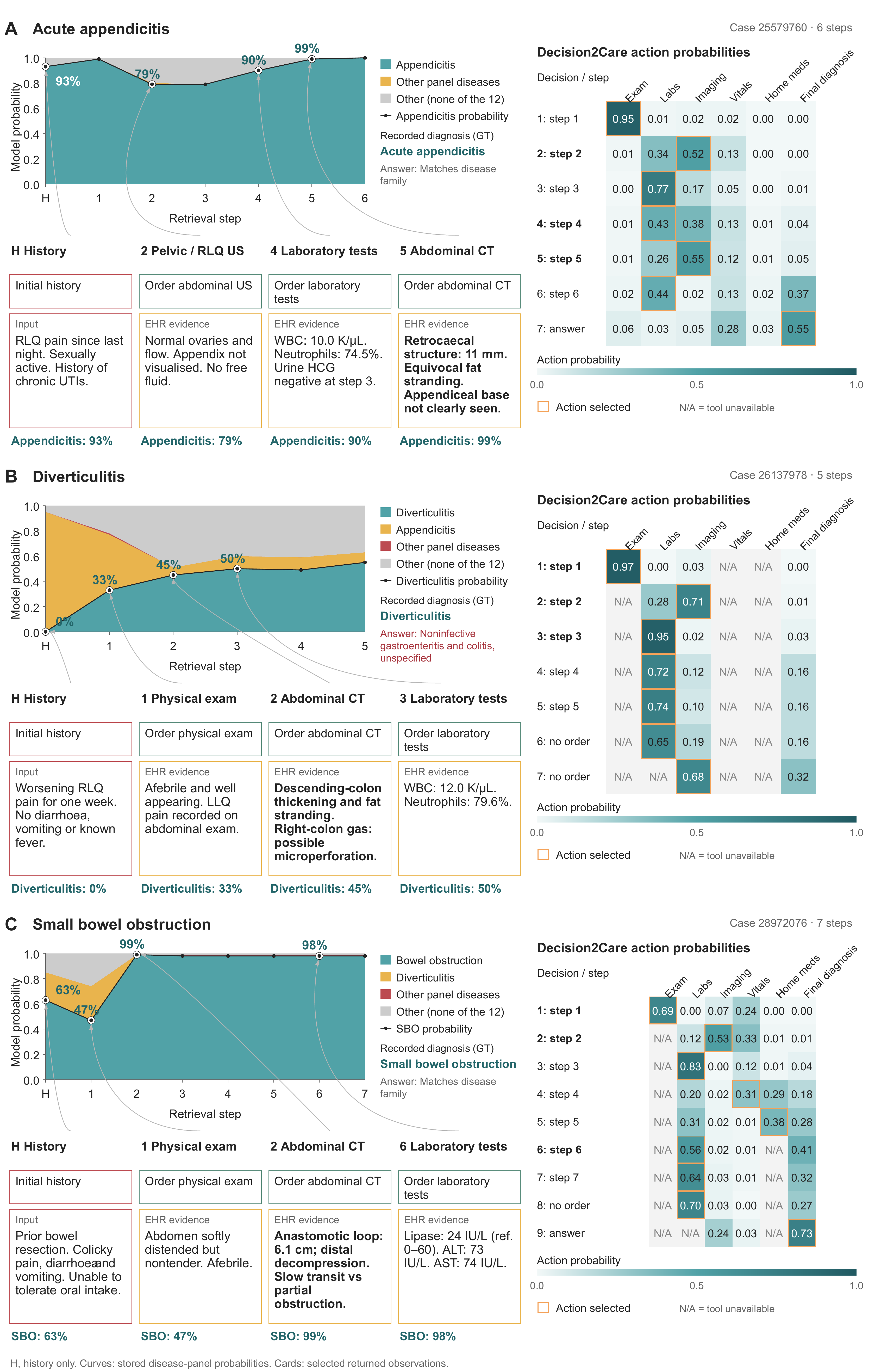}
\end{figure}
\legendnextpage{\textbf{Additional abdominal work-ups.}
    Appendicitis (\textbf{A}), diverticulitis (\textbf{B}) and small bowel obstruction (\textbf{C}).
    Baseline work-ups, displayed as in \figref[, A to~C]{fig:fig3}.
    \label{fig:supp-diagnosis-trajectories}}
\clearpage
\clearpage

\begin{figure}[p]
    \centering
    \includegraphics[width=\textwidth,height=0.999\textheight,keepaspectratio]{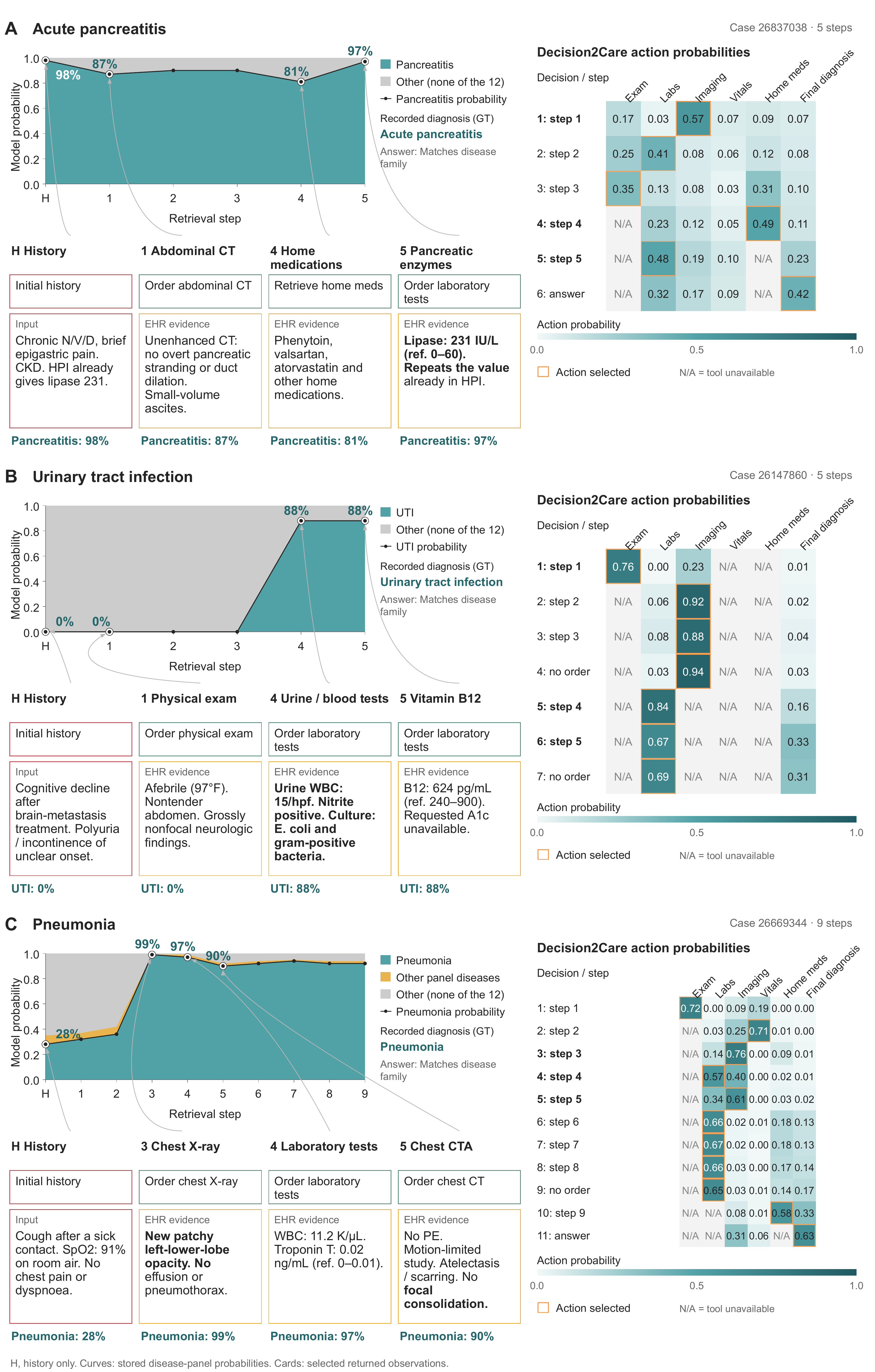}
\end{figure}
\legendnextpage{\textbf{Additional diagnostic work-ups.}
    Pancreatitis (\textbf{A}), urinary tract infection (\textbf{B}) and pneumonia (\textbf{C}).
    Baseline work-ups, displayed as in \figref[, A to~C]{fig:fig3}.
    \label{fig:supp-diagnosis-trajectories-2}}
\clearpage
\clearpage

\begin{figure}[p]
    \centering
    \includegraphics[width=\textwidth,height=0.999\textheight,keepaspectratio]{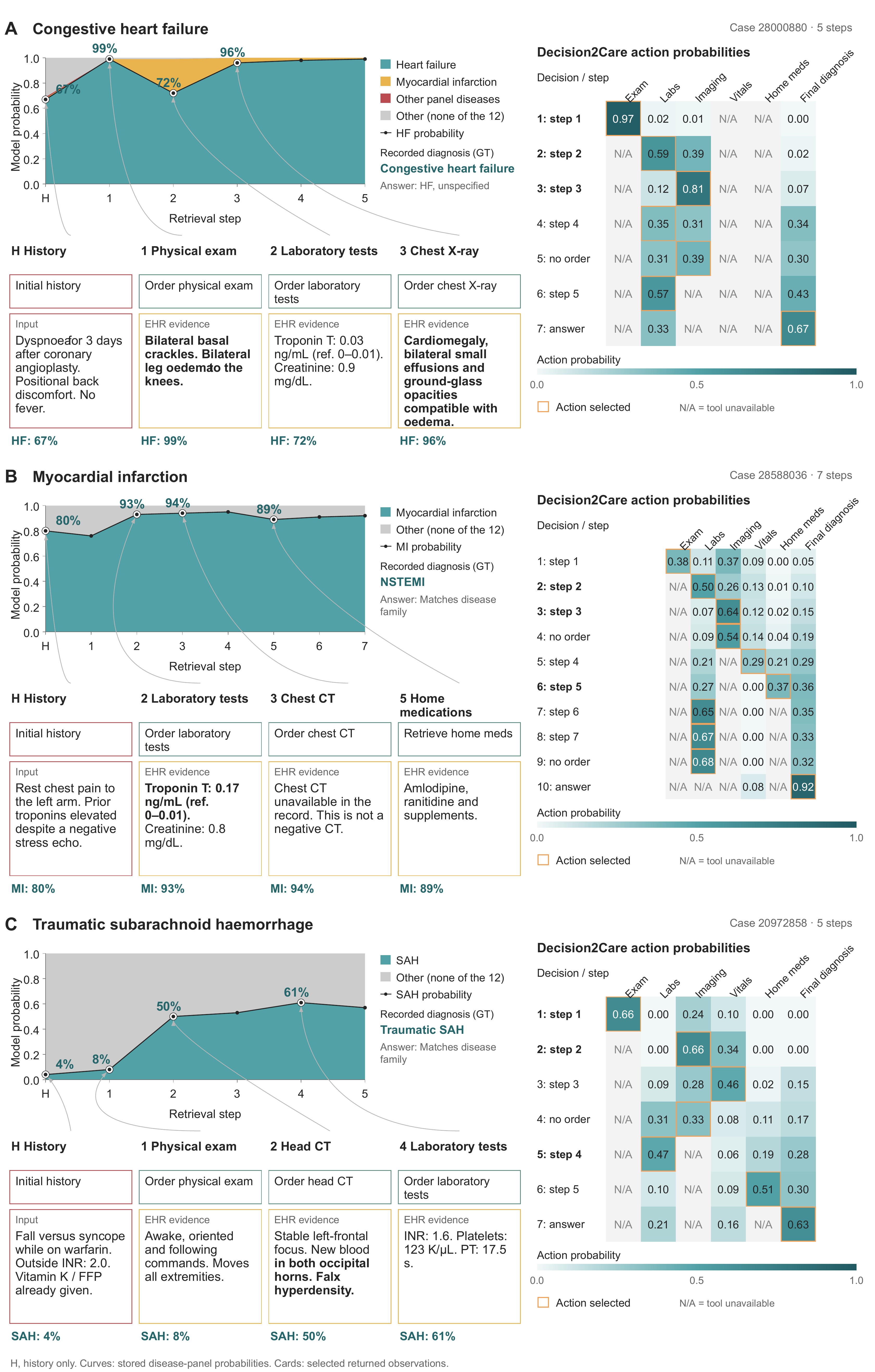}
\end{figure}
\legendnextpage{\textbf{Cardiac and neurological work-ups.}
    Heart failure (\textbf{A}), myocardial infarction (\textbf{B}) and traumatic subarachnoid haemorrhage (\textbf{C}).
    Baseline work-ups, displayed as in \figref[, A to~C]{fig:fig3}.
    \label{fig:supp-diagnosis-trajectories-3}
    }
\clearpage
\clearpage

\begin{figure}[H]
    \centering
    \includegraphics[width=0.55\textwidth]{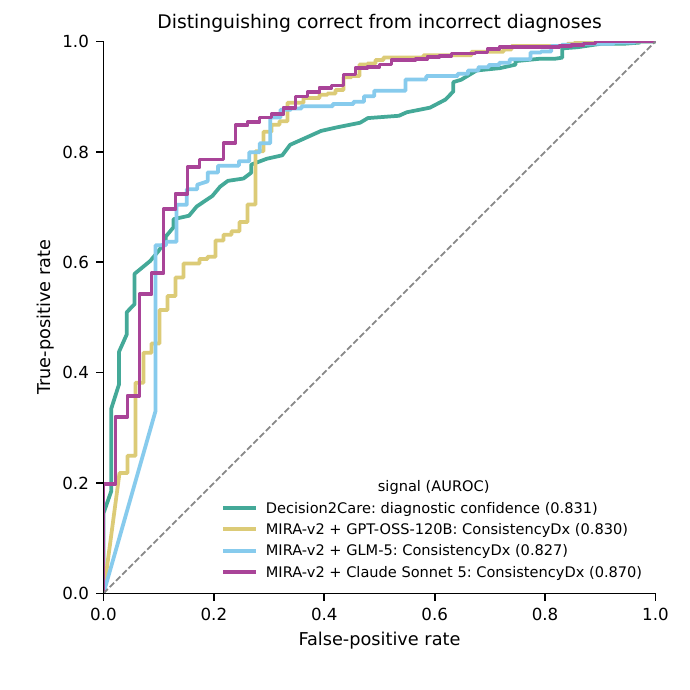}
    \caption{\textbf{Receiver operating characteristic (ROC) curves of the deferral signals.}
    Each system's signal against its own verdicts on the 546 paired admissions, with the area under the ROC curve (AUROC): Decision2Care's confidence in its diagnosis, read once after the work-up in the baseline run, and MIRA-v2's ConsistencyDx, the agreement of the diagnoses of its 5 runs, a case counted as correct when the majority of its runs are correct.}
    \label{fig:supp-deferral}
\end{figure}

\clearpage
\begin{figure}[p]
    \centering
    \includegraphics[width=\textwidth]{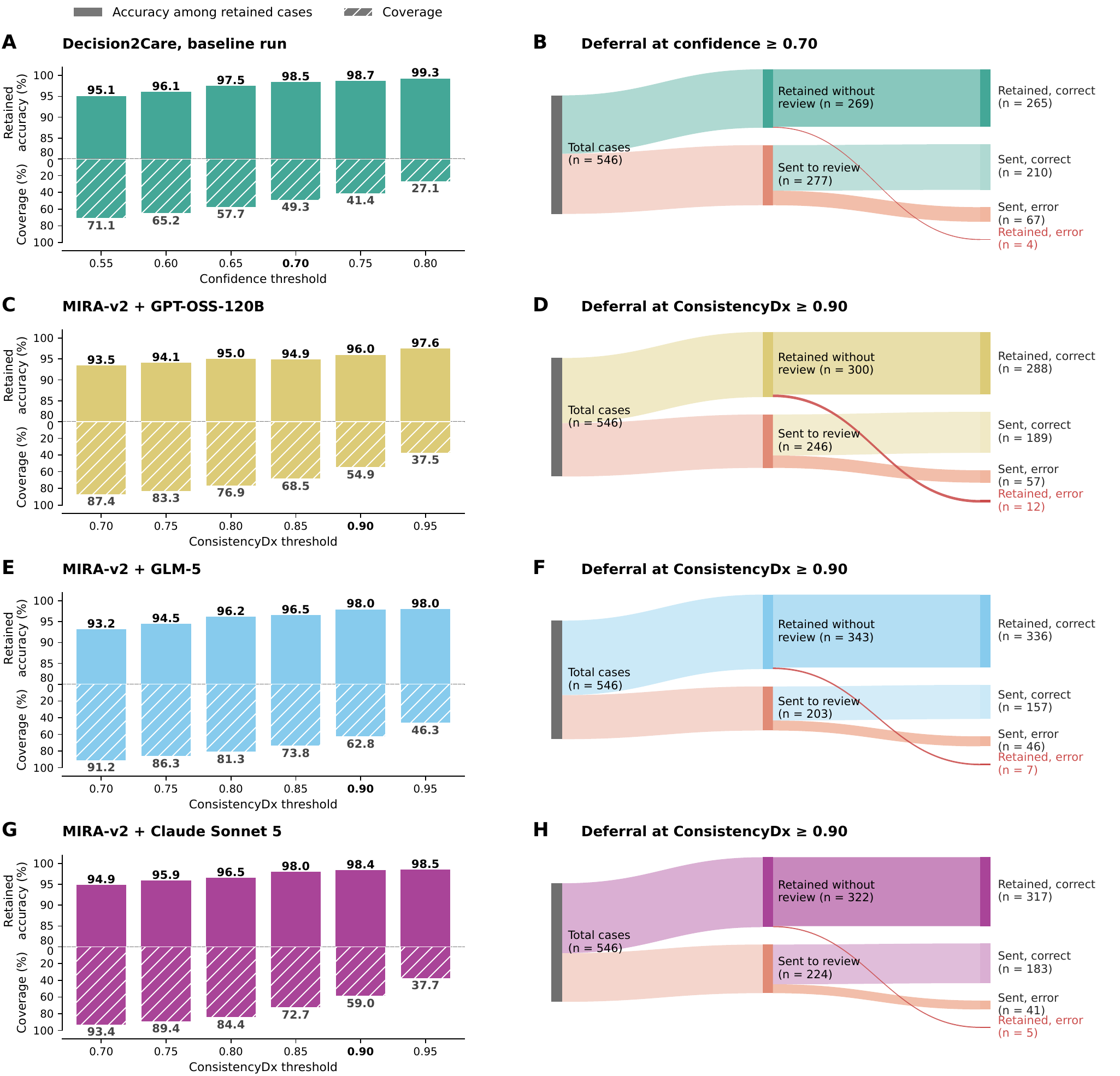}
    \caption{\textbf{Deferral from 1 Decision2Care work-up and 5 MIRA-v2 work-ups.}
    The 546 paired admissions: Decision2Care's baseline run (\textbf{A}, \textbf{B}) and MIRA-v2 with GPT-OSS-120B (\textbf{C}, \textbf{D}), GLM-5 (\textbf{E}, \textbf{F}) and Claude Sonnet 5 (\textbf{G}, \textbf{H}).
    \textbf{A}, \textbf{C}, \textbf{E}, \textbf{G}, Accuracy among retained cases (solid) and coverage, the share retained without review (hatched) as the threshold on Decision2Care's confidence or on ConsistencyDx\cite{zhang2026onprem} rises; a MIRA-v2 case is correct when most of its runs are.
    \textbf{B}, \textbf{D}, \textbf{F}, \textbf{H}, Simulated deferral at a confidence of at least 0.70 and at the published ConsistencyDx threshold of 0.90; counts in table~\ref{tab:supp-deferral}.
    Correct and error describe the agent's diagnosis before review.}
    \label{fig:supp-deferral-comparators}
\end{figure}
\clearpage

\begin{figure}[p]
    \centering
    \includegraphics[width=\textwidth]{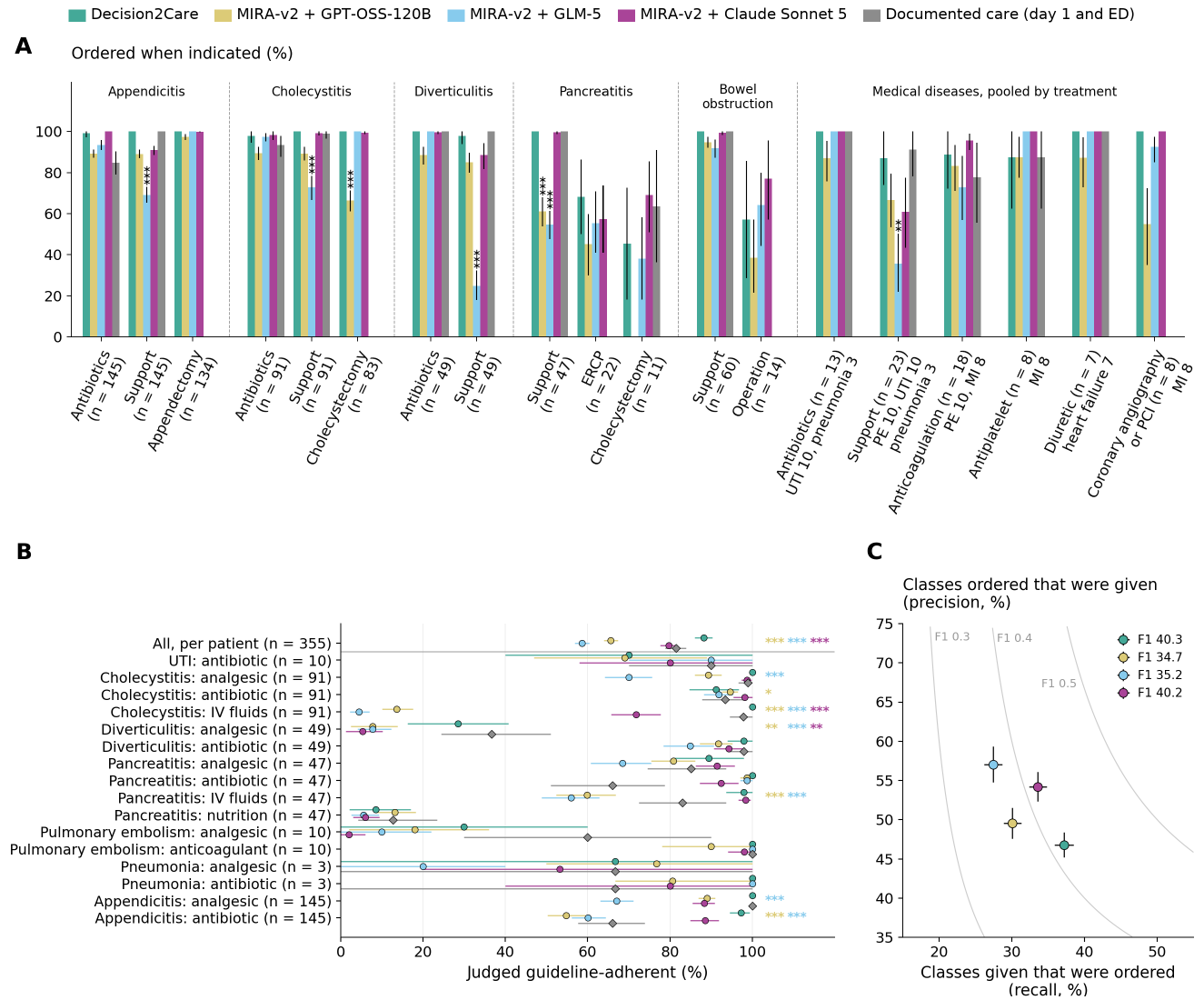}
    \caption{\textbf{Treatment orders against the rubrics, the guidelines and the documented care.}
    The 437 of the 546 paired admissions that every system diagnosed correctly; whiskers, 95\% bootstrap intervals over the admissions.
    \textbf{A}, Share of admissions in which each system ordered a treatment that the disease's rubric indicated, by disease group and item; grey, the care documented on the first day and in the emergency department (ED).
    \textbf{B}, Share of admissions whose drug orders the evaluator of Ferber et~al.\cite{ferber2026mira} judged guideline-adherent, by disease and category and, in the first row, averaged over each patient's categories, on the 355 admissions with one of its 7 diseases; diamonds, documented care.
    \textbf{C}, Drug classes ordered against those given on the first day or in the ED, precision against recall per admission, with the F1 score (the harmonic mean of precision and recall) in the legend.
    Stars, a model against Decision2Care (exact McNemar test, Benjamini--Hochberg corrected; * $q < 0.05$, ** $q < 0.01$, *** $q < 0.001$; the first row of \textbf{B}, Wilcoxon signed-rank test, uncorrected, with the same thresholds for $p$).}
    \label{fig:supp-treatment-comparison}
\end{figure}
\clearpage

\begin{figure}[p]
    \centering
    \captionsetup{skip=4pt}
    \includegraphics[width=\textwidth,trim=0 4 0 10,clip]{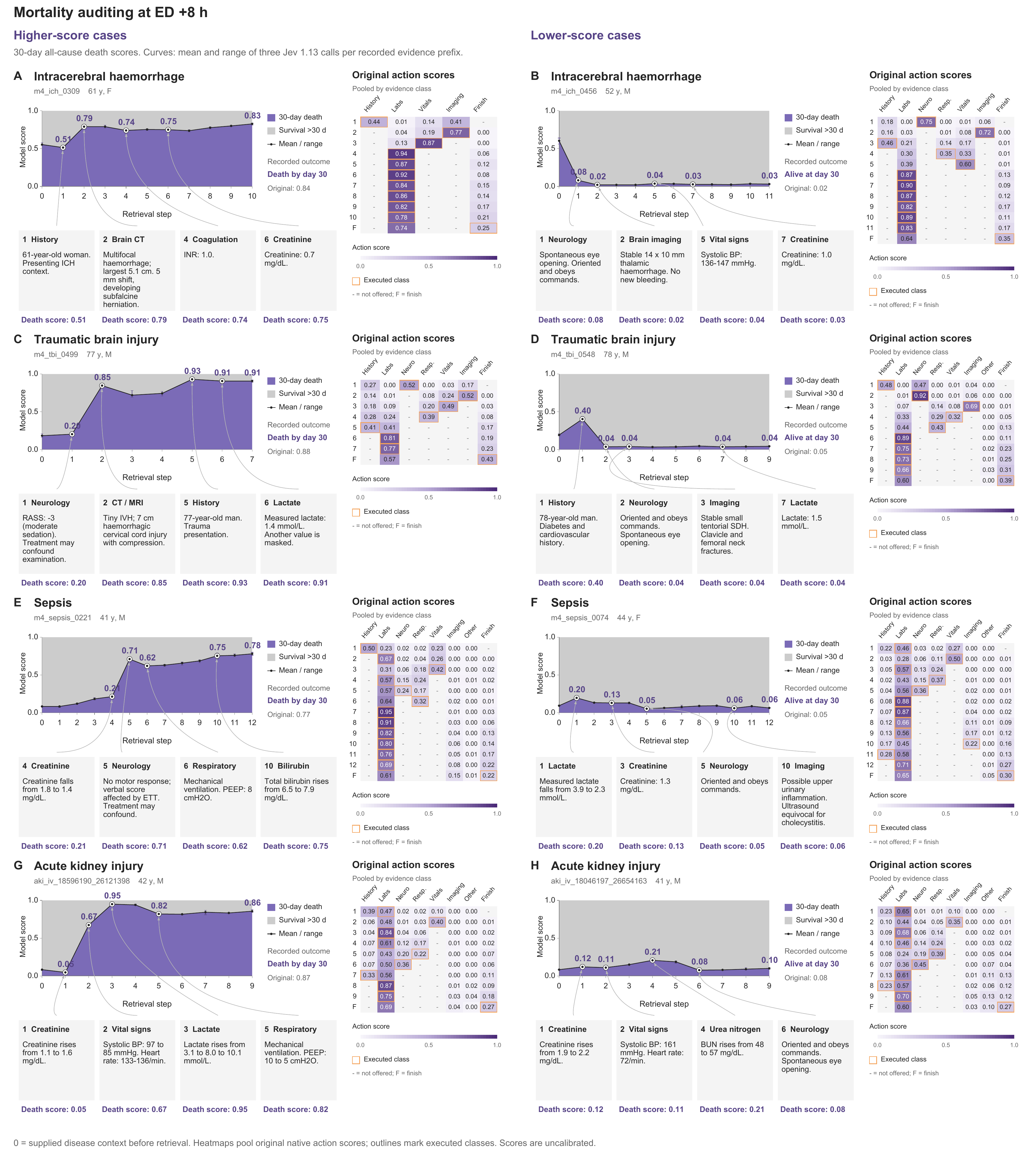}
    \caption{\textbf{Auditing emergency department (ED) mortality predictions along the work-up.}
    ICH, TBI, sepsis and AKI cases at ED +8~h, with higher scores on the left and lower scores on the right.
    Curves show mean 30-day death scores and ranges of 3 queries along fixed retrieval paths; cards show returned evidence; heatmaps show pooled original action scores.
    }
    \label{fig:supp-ed8-audit-1}
\end{figure}
\clearpage

\begin{figure}[p]
    \centering
    \includegraphics[width=\textwidth,height=0.84\textheight,keepaspectratio]{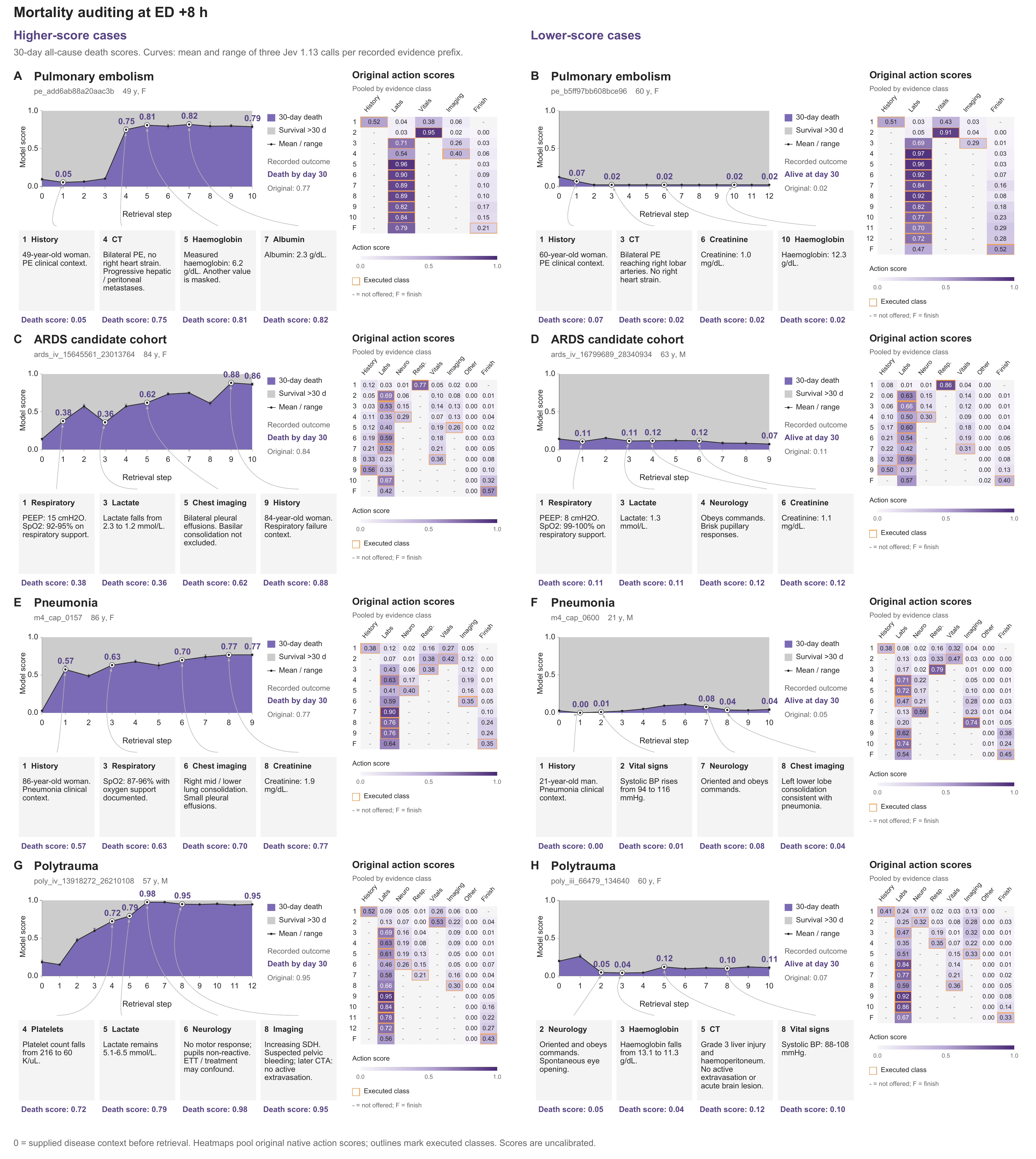}
    \caption{\textbf{Mortality prediction audits in four additional disease cohorts.}
    Examples from PE, the ARDS candidate cohort, pneumonia and polytrauma at ED +8~h; conventions as in \suppfigref{fig:supp-ed8-audit-1}.
    }
    \label{fig:supp-ed8-audit-2}
\end{figure}
\clearpage

\begin{figure}[p]
    \centering
    \includegraphics[width=\textwidth,height=0.84\textheight,keepaspectratio]{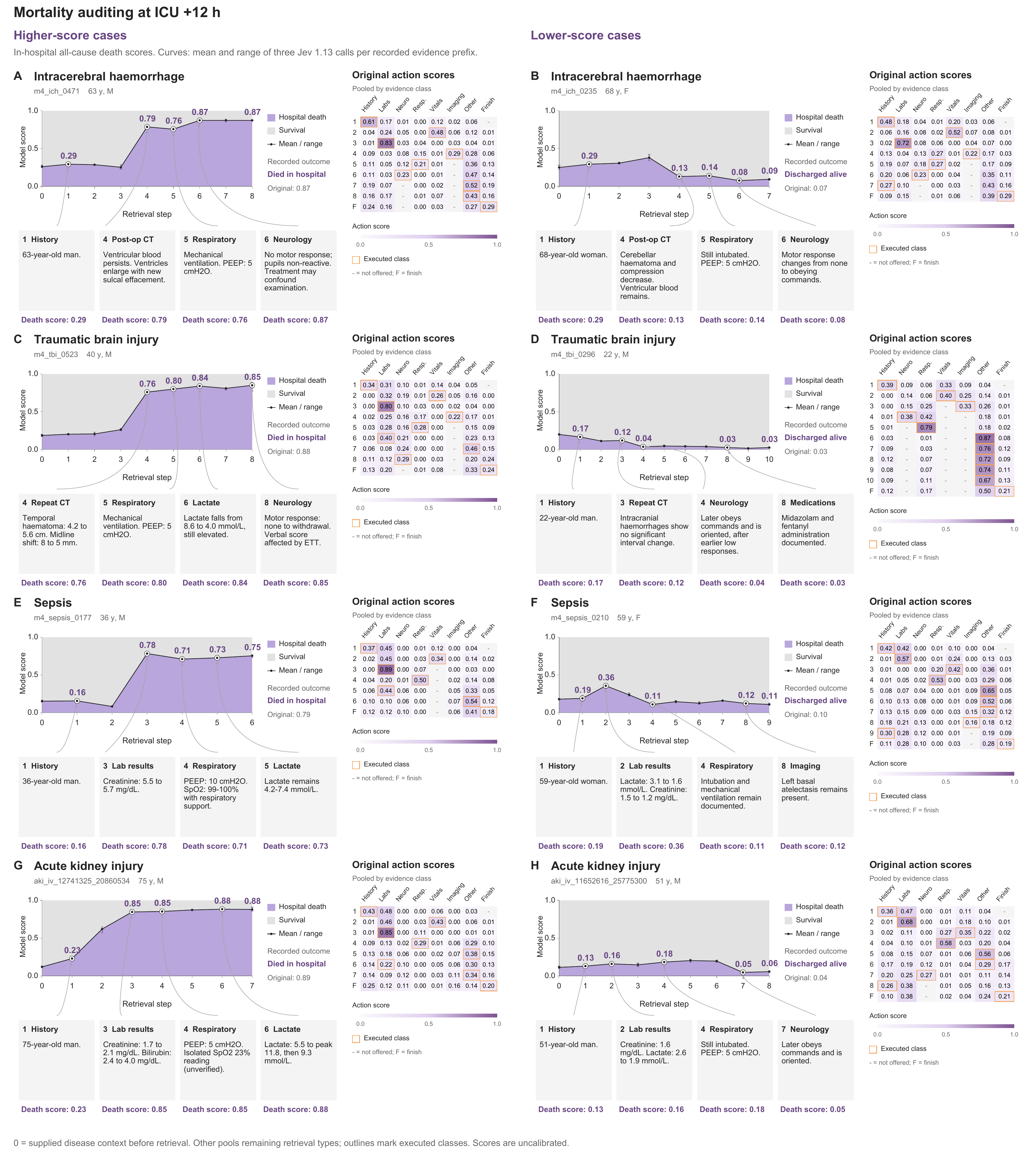}
    \caption{\textbf{Auditing intensive care unit (ICU) mortality predictions.}
    ICH, TBI, sepsis and AKI at ICU +12~h: higher-score cases (left), lower-score cases (right).
    Curves show mean in-hospital death scores and ranges of 3 queries along fixed retrieval paths; cards show returned evidence; heatmaps show pooled original action scores.
    }
    \label{fig:supp-icu12-audit-1}
\end{figure}
\clearpage

\begin{figure}[p]
    \centering
    \includegraphics[width=\textwidth,height=\textheight,keepaspectratio]{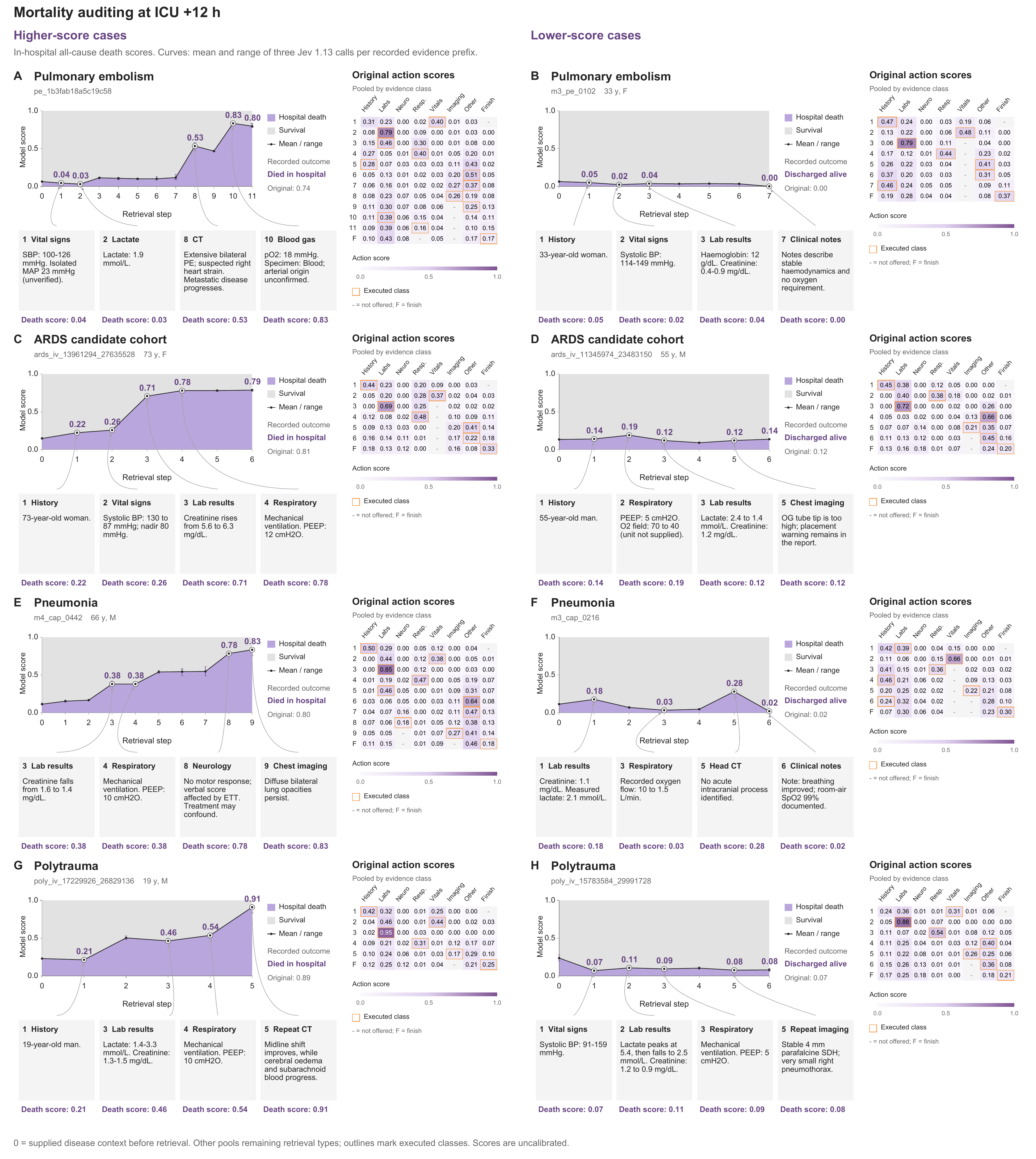}
    \caption{\textbf{ICU mortality audits in four additional disease cohorts.}
    PE, the ARDS candidate cohort, pneumonia and polytrauma at ICU +12~h; conventions as in \suppfigref{fig:supp-icu12-audit-1}.
    }
    \label{fig:supp-icu12-audit-2}
\end{figure}
\clearpage

\begin{figure}[p]
    \centering
    \includegraphics[width=\textwidth,height=0.84\textheight,keepaspectratio]{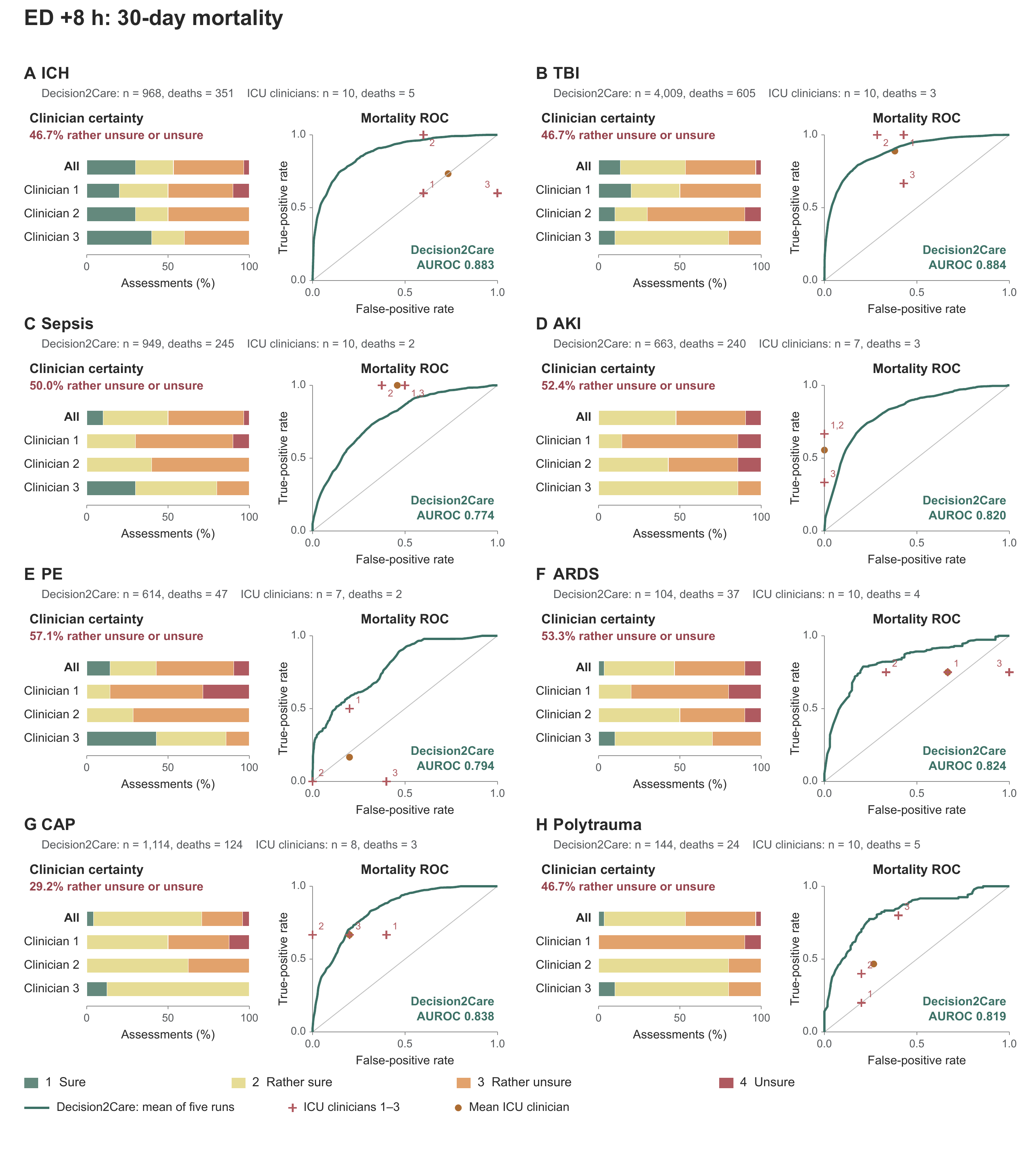}
    \caption{\textbf{30-day mortality predictions by disease.}
    At 8~h after emergency department (ED) arrival, receiver operating characteristic (ROC) curves and areas under those curves (AUROCs) average 5 agent runs in the full MIMIC-III and MIMIC-IV disease cohorts.
    Crosses show the interval predictions of 3 ICU clinicians on 72 selected cases (7--10 per disease); dots, their mean.
    Stacked bars show their certainty ratings, pooled (All) and by ICU clinician.
    }
    \label{fig:supp-mortality-30-disease}
\end{figure}
\clearpage

\begin{figure}[p]
    \centering
    \includegraphics[width=\textwidth,height=0.84\textheight,keepaspectratio]{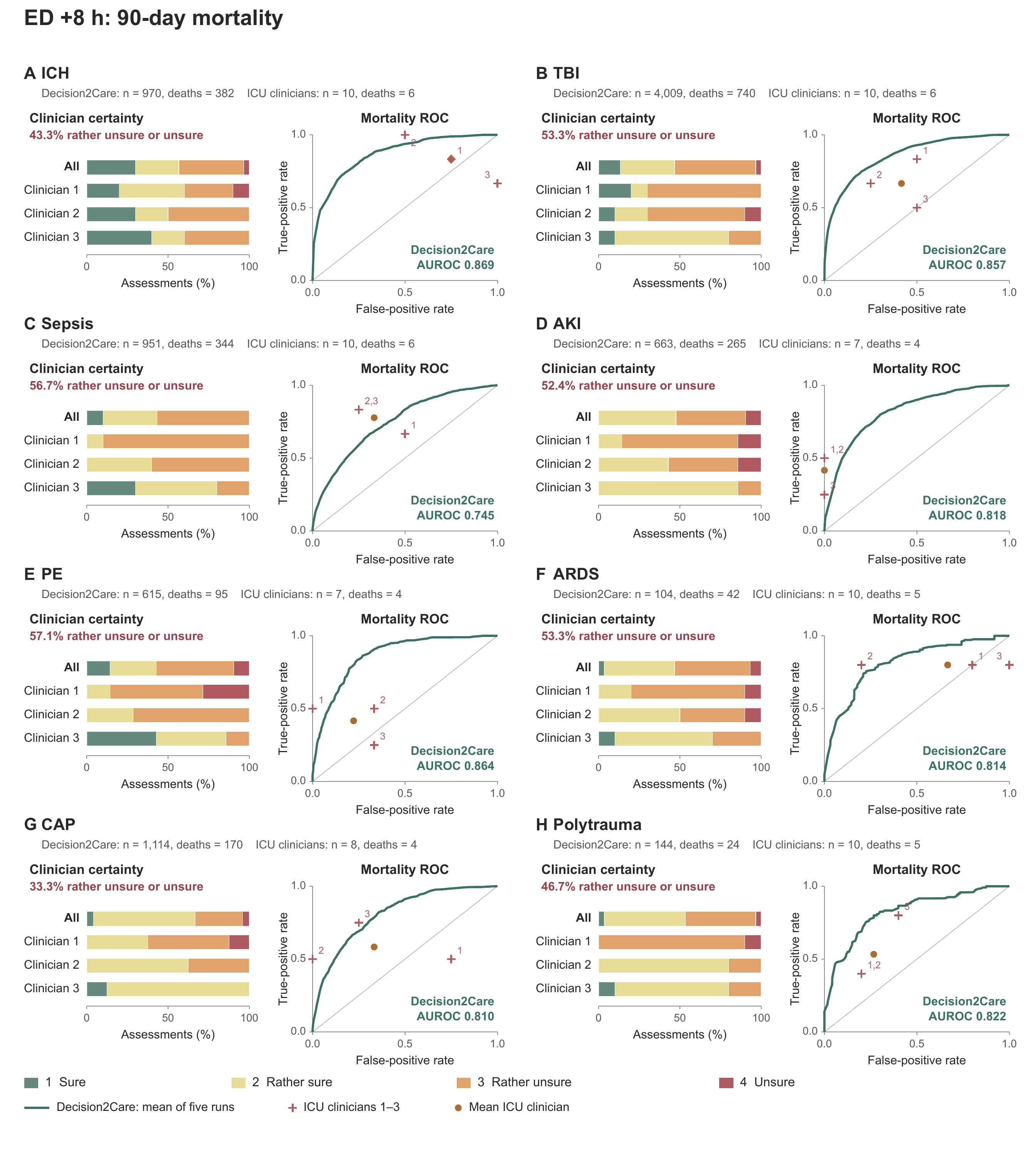}
    \caption{\textbf{90-day mortality predictions by disease.}
    At 8~h after emergency department (ED) arrival, receiver operating characteristic (ROC) curves and areas under those curves (AUROCs) average 5 agent runs in the full MIMIC-III and MIMIC-IV disease cohorts.
    Crosses show the interval predictions of 3 ICU clinicians on 72 selected cases (7--10 per disease); dots, their mean.
    Stacked bars show their certainty ratings, pooled (All) and by ICU clinician.
    }
    \label{fig:supp-mortality-90-disease}
\end{figure}
\clearpage

\begin{figure}[p]
    \centering
    \includegraphics[width=\textwidth]{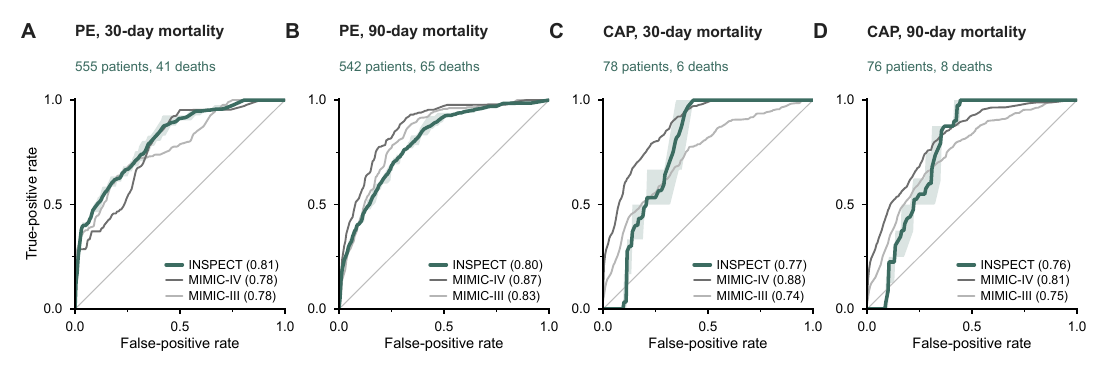}
    \caption{\textbf{Mortality predictions at Stanford Health Care.}
    \textbf{A}--\textbf{D}, Receiver operating characteristic (ROC) curves of the prediction made 8~h after arrival in the emergency department for death within 30~days (\textbf{A}, \textbf{C}) and 90~days (\textbf{B}, \textbf{D}), in pulmonary embolism (PE; \textbf{A}, \textbf{B}) and pneumonia (\textbf{C}, \textbf{D}).
    Green, the patients of INSPECT, a data set of Stanford Health Care, whose number and deaths are given above each panel; grey, the cohorts of MIMIC-III and MIMIC-IV, from Beth Israel Deaconess Medical Center.
    Each curve is the mean of 5 runs, and the band is the range of the 5 runs in INSPECT.
    The key gives the area under each curve.
    In INSPECT the pneumonia group consists of candidates whose community acquisition could not be verified.
    }
    \label{fig:supp-second-centre}
\end{figure}
\clearpage

\begin{figure}[p]
    \centering
    \includegraphics[width=\textwidth]{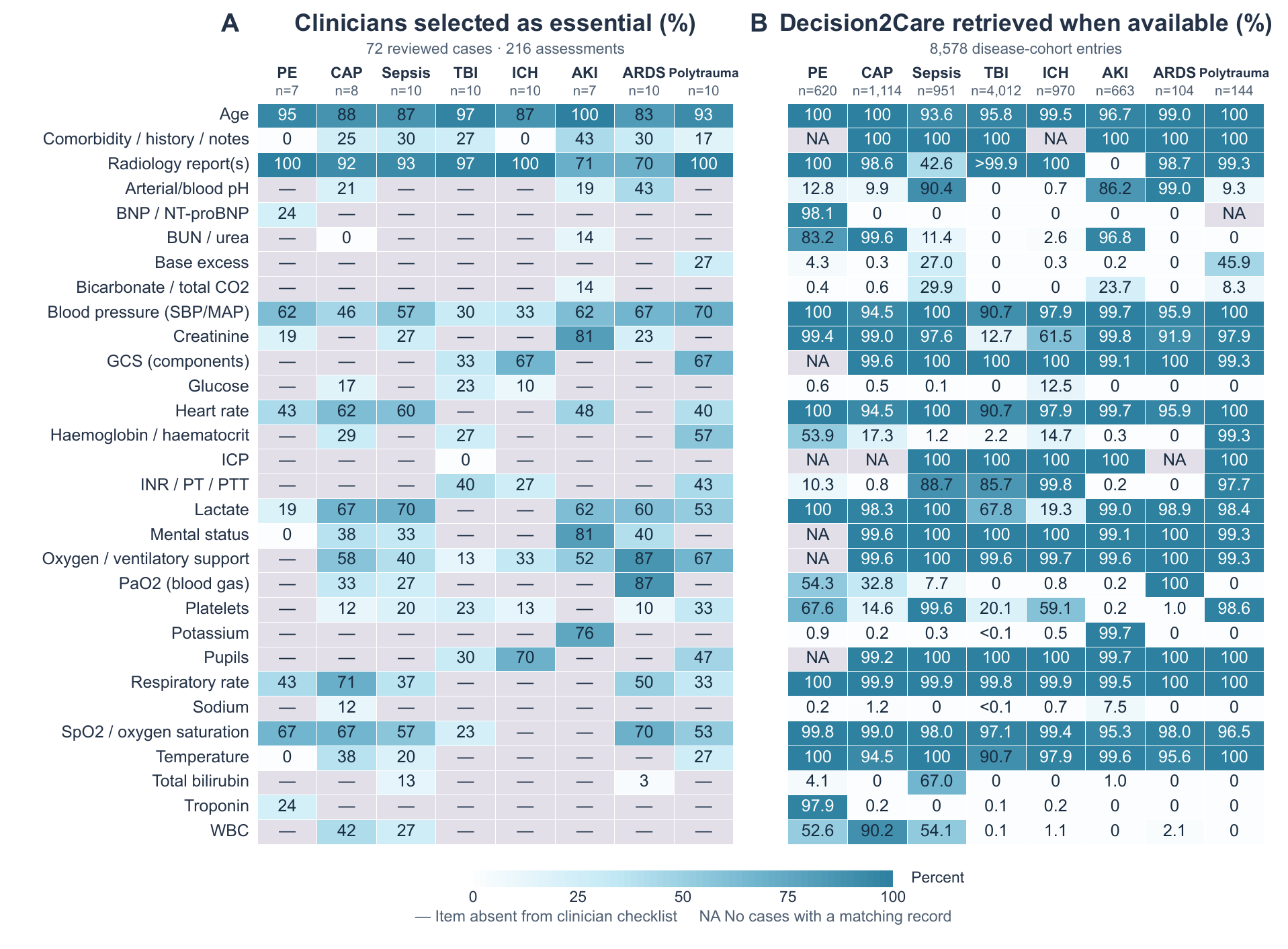}
    \caption{\textbf{The records the ICU clinicians marked essential and the records the agent read.}
    \textbf{A}, Share of the 216 assessments of the 72 review cases (3 ICU clinicians, before seeing the agent's work-up) that marked each kind of record essential, by disease; dash, not on that disease's checklist.
    \textbf{B}, Share of the cases with an available record of each kind at 8~h after arrival in which the agent read one, in the full MIMIC-III and MIMIC-IV cohorts of the 8 diseases (8,578 cohort entries, n above each column; 1 run); NA, no case had an available record of that kind.
    }
    \label{fig:supp-retrieval-audit}
\end{figure}

\clearpage

\begin{figure}[p]
    \centering
    \includegraphics[width=\textwidth]{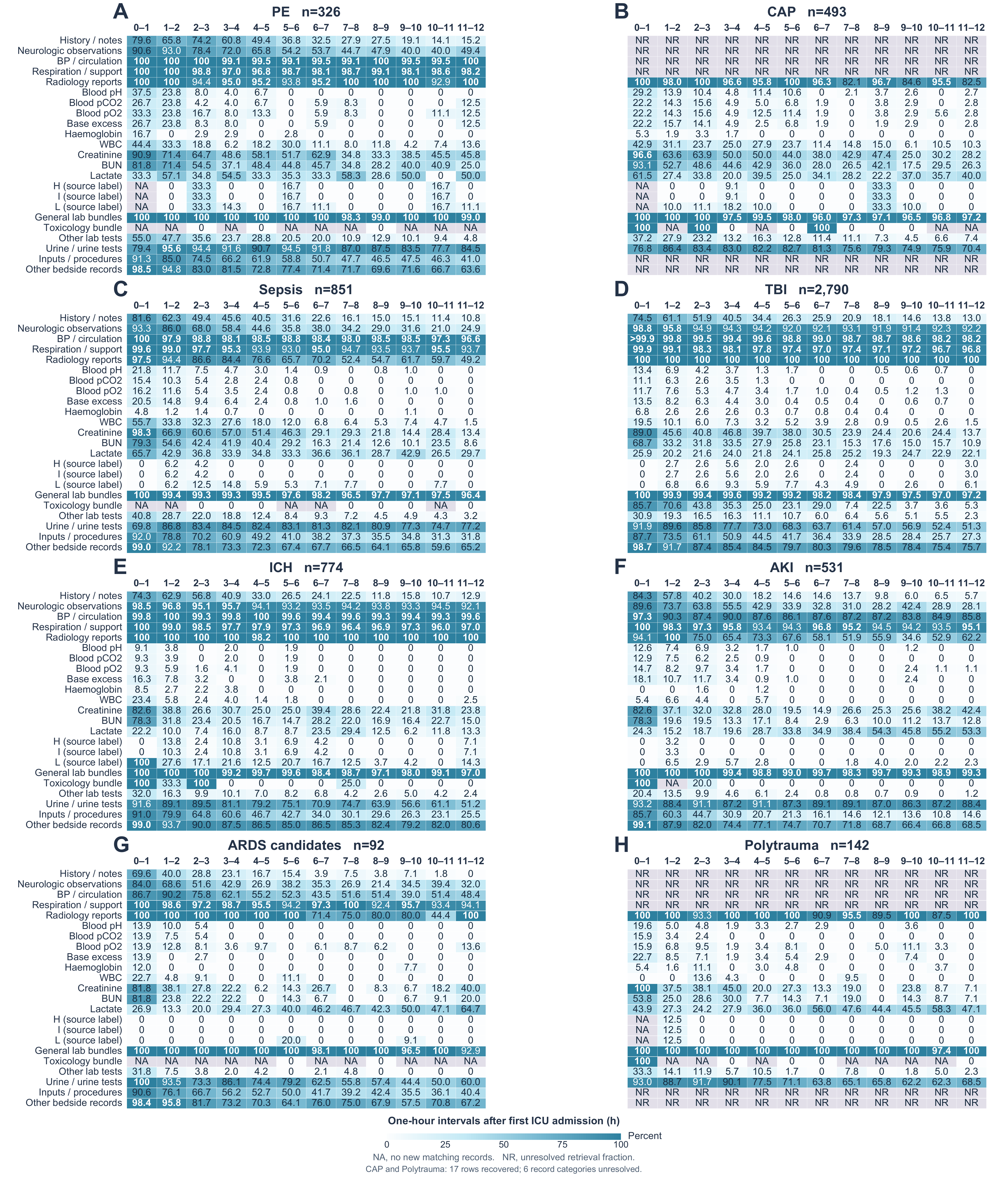}
\end{figure}
\legendnextpage{\textbf{Hourly retrieval of newly available intensive care unit (ICU) records.}
    \textbf{A}--\textbf{H}, Disease cohorts; $n$, cohort entries.
    Among cases with matching new records in each 1-h interval, cells show the percentage in which at least 1 was retrieved.
    Counts are pooled across MIMIC-III and MIMIC-IV.
    Each prediction independently uses cumulative ICU 0--$h$ records.
    Individual test rows exclude bundled returns.
    \label{fig:supp-hourly-retrieval}
    }

\clearpage
\clearpage

\begingroup
\normalsize
\setlength{\tabcolsep}{2pt}
\renewcommand{\arraystretch}{1.0}
\setlength{\LTcapwidth}{\linewidth}
\setlength{\LTleft}{0pt}
\setlength{\LTright}{0pt plus 1fil}
\begin{longtable}{@{\extracolsep{\fill}}L{0.12\linewidth}L{0.56\linewidth}C{0.12\linewidth}C{0.16\linewidth}@{}}
\toprule
\textbf{Code} & \textbf{ICD-10-CM title} & \textbf{Answers} & \textbf{Share of errors (\%)} \\
\midrule
\endfirsthead
\multicolumn{4}{c}{\tablename\ \thetable\ -- continued}\\
\toprule
\textbf{Code} & \textbf{ICD-10-CM title} & \textbf{Answers} & \textbf{Share of errors (\%)} \\
\midrule
\endhead
\midrule
\multicolumn{4}{r}{Continued on next page}\\
\endfoot
\endlastfoot
K80.20 & Calculus of gallbladder without cholecystitis without obstruction & 98 & 68.1 \\
K85.10 & Biliary acute pancreatitis without necrosis or infection & 20 & 13.9 \\
K83.1 & Obstruction of bile duct & 10 & 6.9 \\
K71.2 & Toxic liver disease with acute hepatitis & 5 & 3.5 \\
K66.1 & Hemoperitoneum & 5 & 3.5 \\
K80.71 & Calculus of gallbladder and bile duct without cholecystitis with obstruction & 5 & 3.5 \\
K83.4 & Spasm of sphincter of Oddi & 1 & 0.7 \\
\midrule
\noalign{\unskip\nobreak}
All & Incorrect predictions over the 5 runs & 144 & 100 \\*
\bottomrule
\noalign{\unskip\nobreak}
\caption{\textbf{ICD-10-CM codes of Decision2Care's incorrect cholecystitis predictions.} The 144 incorrect predictions over 5 runs on the 546 paired admissions, by the code given. Every reference code specifies acute cholecystitis. In 88 of the 98 predictions coded K80.20, the reference code lay in the same category, K80. Decision2Care's accuracy in cholecystitis was 77.7\% (95\% confidence interval 70.2--84.3\%; 129 admissions), and of the comparisons by disease with the 3 MIRA-v2 configurations, only that with Claude Sonnet 5 in cholecystitis remained significant after Holm's correction over the 12 diseases ($p < 0.001$).}\label{tab:supp-cholecystitis-errors}\\
\end{longtable}
\endgroup

\clearpage

\begingroup
\footnotesize
\setlength{\tabcolsep}{2pt}
\renewcommand{\arraystretch}{1.0}
\setlength{\LTcapwidth}{\linewidth}
\setlength{\LTleft}{0pt}
\setlength{\LTright}{0pt plus 1fil}
\begin{longtable}{@{\extracolsep{\fill}}L{0.225\linewidth}C{0.105\linewidth}C{0.14\linewidth}C{0.135\linewidth}C{0.14\linewidth}C{0.18\linewidth}@{}}
\toprule
\textbf{Disease} & \makecell{\textbf{Physicians,}\\\textbf{correct}} & \makecell{\textbf{Physicians,}\\\textbf{accuracy, \%}} & \makecell{\textbf{Decision2Care,}\\\textbf{admissions}} & \makecell{\textbf{Decision2Care,}\\\textbf{accuracy, \%}} & \makecell{\textbf{Difference,}\\\textbf{points}} \\
\midrule
\endfirsthead
\multicolumn{6}{c}{\tablename\ \thetable\ -- continued}\\
\toprule
\textbf{Disease} & \makecell{\textbf{Physicians,}\\\textbf{correct}} & \makecell{\textbf{Physicians,}\\\textbf{accuracy, \%}} & \makecell{\textbf{Decision2Care,}\\\textbf{admissions}} & \makecell{\textbf{Decision2Care,}\\\textbf{accuracy, \%}} & \makecell{\textbf{Difference,}\\\textbf{points}} \\
\midrule
\endhead
\midrule
\multicolumn{6}{r}{Continued on next page}\\
\endfoot
\endlastfoot
Appendicitis & 38 of 43 & 88.4 & 148 & 99.9 & $+11.5$ \\
Cholecystitis & 36 of 45 & 80.0 & 129 & 77.7 & $-2.3$ \\
Diverticulitis & 36 of 44 & 81.8 & 54 & 88.1 & $+6.3$ \\
Pancreatitis & 33 of 42 & 78.6 & 52 & 91.5 & $+13.0$ \\
Pneumonia & 15 of 26 & 57.7 & 21 & 45.7 & $-12.0$ \\
Pulmonary embolism & 40 of 45 & 88.9 & 15 & 92.0 & $+3.1$ \\
Urinary tract infection & 27 of 45 & 60.0 & 18 & 82.2 & $+22.2$ \\
\midrule
All 7 diseases & 225 of 290 & 77.6 (72.4--82.0) & 437 & 84.6 (80.2--88.7) & $+7.0$ ($+0.7$ to $+13.2$) \\
The 4 abdominal diseases & 143 of 174 & 82.2 (75.8--87.2) & 383 & 89.2 (85.8--92.3) & $+7.0$ ($+0.6$ to $+13.5$) \\
All 7 diseases, history without an emergency department course & 225 of 290 & 77.6 (72.4--82.0) & 321 & 84.6 (79.1--89.7) & $+7.0$ ($-0.2$ to $+13.9$) \\*
\bottomrule
\noalign{\unskip\nobreak}
\caption{\textbf{Diagnostic accuracy of Decision2Care and of board-certified physicians on the MIRA benchmark.} The physicians' results are those published for the 4 board-certified physicians of the MIRA study\cite{ferber2026mira}, who each worked up a random, non-overlapping share of 311 benchmark admissions. The 21 admissions with pancreatic cancer, which MIRA-v2 does not include, are left out. The admissions that the physicians read have not been published, so the 2 results were not compared admission by admission. In the 4 abdominal diseases (appendicitis, cholecystitis, diverticulitis and pancreatitis), the physicians' admissions are a random sample of those on which Decision2Care was evaluated. Decision2Care's accuracy is the mean of 5 runs. The pooled rows weight each disease by the number of admissions the physicians read. 95\% confidence intervals (in parentheses) are Wilson intervals for the physicians and percentile intervals from 20,000 resamples of admissions within each disease for Decision2Care. For the difference, the physicians' correct answers were resampled as well, from a binomial distribution. The physicians took the history from MIRA's patient agent, whose instructions restricted disclosure of the emergency department course. The last row keeps only the admissions whose recorded history holds no such course (materials and methods).}\label{tab:supp-physicians}\\
\end{longtable}
\endgroup

\clearpage

\begingroup
\normalsize
\setlength{\tabcolsep}{2pt}
\renewcommand{\arraystretch}{1.0}
\setlength{\LTcapwidth}{\linewidth}
\setlength{\LTleft}{0pt}
\setlength{\LTright}{0pt plus 1fil}
\begin{longtable}{@{\extracolsep{\fill}}L{0.44\linewidth}C{0.10\linewidth}C{0.10\linewidth}C{0.10\linewidth}C{0.10\linewidth}C{0.10\linewidth}@{}}
\toprule
\textbf{Variant} & \makecell{\textbf{Accuracy}\\\textbf{(\%)}} & \makecell{\textbf{Change}\\\textbf{(points)}} & \makecell{\textbf{95\% half-}\\\textbf{width}} & \textbf{$p$} & \makecell{\textbf{$p$}\\\textbf{(Holm)}} \\
\midrule
\endfirsthead
\multicolumn{6}{c}{\tablename\ \thetable\ -- continued}\\
\toprule
\textbf{Variant} & \makecell{\textbf{Accuracy}\\\textbf{(\%)}} & \makecell{\textbf{Change}\\\textbf{(points)}} & \makecell{\textbf{95\% half-}\\\textbf{width}} & \textbf{$p$} & \makecell{\textbf{$p$}\\\textbf{(Holm)}} \\
\midrule
\endhead
\midrule
\multicolumn{6}{r}{Continued on next page}\\
\endfoot
\endlastfoot
ICD options in a seeded random order (seed 1) & 87.7 & +1.25 & 1.39 & 0.080 & 0.798 \\
ICD options in a seeded random order (seed 2) & 86.8 & +0.33 & 1.17 & 0.579 & 1.000 \\
ICD options in a seeded random order (seed 3) & 86.6 & +0.15 & 1.45 & 0.843 & 1.000 \\
Every ICD Choice asked under 5 random orders, probabilities averaged & 87.5 & +1.06 & 1.04 & 0.046 & 0.510 \\
ICD instructions paraphrased (set 1), routing unchanged & 86.1 & -0.40 & 1.21 & 0.513 & 1.000 \\
ICD instructions paraphrased (set 2), routing unchanged & 87.0 & +0.51 & 0.98 & 0.307 & 1.000 \\
ICD instructions paraphrased (set 3), routing unchanged & 86.8 & +0.33 & 0.92 & 0.481 & 1.000 \\
All instructions paraphrased (set 1) & 85.0 & -1.50 & 1.20 & 0.014 & 0.172 \\
All instructions paraphrased (set 2) & 86.4 & -0.04 & 1.39 & 0.959 & 1.000 \\
All instructions paraphrased (set 3) & 84.6 & -1.87 & 1.38 & 0.008 & 0.103 \\
Laboratory results as raw values, no flags or reference ranges & 84.4 & -2.05 & 1.35 & 0.003 & 0.040 \\
Laboratory results with flags, no reference ranges & 86.3 & -0.22 & 0.84 & 0.610 & 1.000 \\
State fields in reverse order & 84.8 & -1.68 & 1.97 & 0.094 & 0.848 \\
State rendered as one text note instead of JSON & 85.3 & -1.14 & 1.42 & 0.117 & 0.939 \\*
\bottomrule
\noalign{\unskip\nobreak}
\caption{\textbf{Decision2Care's accuracy under variants of option order, wording and record format.} Each variant run once on the 546 paired admissions and compared case by case with the mean of the 5 runs as recorded (86.48\%). Change: the mean per-case difference in strict accuracy, in points, with the half-width of its 95\% interval and the two-sided $p$ value, uncorrected and after Holm's correction over the 14 variants. JSON, JavaScript Object Notation.}\label{tab:supp-robustness-variants}\\
\end{longtable}
\endgroup

\clearpage

\begingroup
\small
\setlength{\tabcolsep}{2pt}
\renewcommand{\arraystretch}{1.0}
\setlength{\LTcapwidth}{\linewidth}
\setlength{\LTleft}{0pt}
\setlength{\LTright}{0pt plus 1fil}
\begin{longtable}{@{\extracolsep{\fill}}L{0.28\linewidth}C{0.07\linewidth}C{0.2\linewidth}C{0.2\linewidth}C{0.2\linewidth}@{}}
\toprule
\textbf{Disease} & \textbf{n} & \makecell{\textbf{Jev 1.13.0}\\\textbf{(5 runs)}} & \makecell{\textbf{PPLX decider v1.2}\\\textbf{(5 runs)}} & \makecell{\textbf{Qwen3.8-Flash-Next}\\\textbf{readout (5 runs)}} \\
\midrule
\endfirsthead
\multicolumn{5}{c}{\tablename\ \thetable\ -- continued}\\
\toprule
\textbf{Disease} & \textbf{n} & \makecell{\textbf{Jev 1.13.0}\\\textbf{(5 runs)}} & \makecell{\textbf{PPLX decider v1.2}\\\textbf{(5 runs)}} & \makecell{\textbf{Qwen3.8-Flash-Next}\\\textbf{readout (5 runs)}} \\
\midrule
\endhead
\midrule
\multicolumn{5}{r}{Continued on next page}\\
\endfoot
\endlastfoot
Appendicitis & 148 & 99.9 & 98.6 & 98.8 \\
Cholecystitis & 129 & 77.7 & 80.6 & 87.8 \\
Diverticulitis & 54 & 88.1 & 90.7 & 92.6 \\
Pancreatitis & 52 & 91.5 & 85.8 & 82.7 \\
Pneumonia & 21 & 45.7 & 41.9 & 52.4 \\
Pulmonary embolism & 15 & 92.0 & 86.7 & 70.7 \\
Urinary tract infection & 18 & 82.2 & 61.1 & 52.2 \\
Bowel obstruction & 67 & 91.6 & 92.2 & 91.3 \\
Heart failure & 13 & 64.6 & 61.5 & 73.8 \\
Subarachnoid haemorrhage & 7 & 42.9 & 28.6 & 28.6 \\
Myocardial infarction & 13 & 69.2 & 90.8 & 84.6 \\
Meningitis & 9 & 100.0 & 100.0 & 97.8 \\
\midrule
MIRA-v2 benchmark cases & 437 & 87.3 & 86.1 & 87.7 \\
Added cases & 109 & 83.3 & 85.0 & 85.0 \\
\midrule
All admissions (95\% CI) & 546 & 86.5 (83.7--89.2) & 85.9 (83.0--88.8) & 87.2 (84.5--89.9) \\
Difference from Jev 1.13.0, points (95\% CI) & & & $-0.6$ ($-2.9$ to $+1.8$) & $+0.7$ ($-1.7$ to $+3.1$) \\*
\bottomrule
\noalign{\unskip\nobreak}
\caption{\textbf{Diagnostic accuracy of Decision2Care with 3 decision models on the 546 paired admissions.} Each model drove the same controller, tools and typed questions over 5 runs: Jev 1.13.0, the primary decision model; local PPLX decider v1.2; and Qwen3.8-Flash-Next, a general large language model read through its answer-token probabilities (temperature 1, thinking disabled). Accuracy is the share of admissions whose top-ranked ICD-10-CM code title matched the recorded disease under the strict MIMIC-CDM rule, the mean over runs; 95\% confidence intervals (CIs) from 5,000 admission-bootstrap resamples with all runs of an admission resampled together, and the paired differences from Jev 1.13.0 likewise. Intervals condition on the saved runs. Qwen minus PPLX, $+1.3$ points ($-1.1$ to $+3.6$). Jev 1.14.0 reached 85.6\% over 5 runs ($-0.8$ from 1.13.0, $-2.5$ to $+0.8$) and PPLX's hosted service 86.3\% in 1 run. PPLX used its checkpoint's decision readout and question splitting; Qwen used independent answer-token readouts (materials and methods).}\label{tab:supp-decision-models}\\
\end{longtable}
\endgroup

\clearpage

\begingroup
\normalsize
\setlength{\tabcolsep}{2pt}
\renewcommand{\arraystretch}{1.0}
\setlength{\LTcapwidth}{\linewidth}
\setlength{\LTleft}{0pt}
\setlength{\LTright}{0pt plus 1fil}
\begin{longtable}{@{\extracolsep{\fill}}L{0.17\linewidth}L{\dimexpr0.48\linewidth-4\tabcolsep\relax}L{0.35\linewidth}@{}}
\toprule
\textbf{Disease} & \textbf{Guideline} & \textbf{Tests counted} \\
\midrule
\endfirsthead
\multicolumn{3}{c}{\tablename\ \thetable\ -- continued}\\
\toprule
\textbf{Disease} & \textbf{Guideline} & \textbf{Tests counted} \\
\midrule
\endhead
\midrule
\multicolumn{3}{r}{Continued on next page}\\
\endfoot
\endlastfoot
Appendicitis & WSES Jerusalem guidelines 2020 (Di Saverio et al., World J Emerg Surg 15:27)\cite{disaverio2020appendicitis} & ultrasound, CT or MRI of the abdomen \\
Cholecystitis & WSES 2020 (Pisano et al., World J Emerg Surg 15:61)\cite{pisano2020cholecystitis} & ultrasound, MRI or HIDA scan of the abdomen \\
Bowel obstruction & WSES position paper on small bowel obstruction 2021 (Amara et al., World J Emerg Surg, doi:10.1186/s13017-021-00379-8)\cite{amara2021sbo} & CT or radiograph of the abdomen \\
Diverticulitis & ASCRS 2020 (Hall et al., Dis Colon Rectum 63:728)\cite{hall2020diverticulitis} & ultrasound, CT or MRI of the abdomen \\
Pancreatitis & WSES 2019 (Lepp\"aniemi et al., World J Emerg Surg 14:27)\cite{leppaniemi2019pancreatitis} & ultrasound or CT of the abdomen; lipase or amylase \\
Pneumonia & ATS/IDSA 2019 (Metlay et al., Am J Respir Crit Care Med 200:e45)\cite{metlay2019pneumonia} & CT or radiograph of the chest \\
UTI & EAU Guidelines on Urological Infections 2024 (Bonkat et al.)\cite{bonkat2024eau} & urinalysis or urine culture \\
Pulmonary embolism & ESC 2019 (Konstantinides et al., Eur Heart J 41:543)\cite{konstantinides2020pe} & CT of the chest; venous ultrasound; D-dimer \\
Heart failure & NICE CG187, acute heart failure\cite{nice2014cg187} & radiograph of the chest; natriuretic peptide; ECG; echocardiography \\
Myocardial infarction & NICE CG95, recent-onset chest pain\cite{nice2010cg95} & troponin; ECG \\
Meningitis & ESCMID 2016 (van de Beek et al., Clin Microbiol Infect 22 Suppl 3:S37)\cite{vandebeek2016meningitis} & cerebrospinal fluid analysis \\
Subarachnoid haemorrhage & NICE NG228\cite{nice2022ng228} & CT of the head; cerebrospinal fluid analysis \\*
\bottomrule
\noalign{\unskip\nobreak}
\caption{\textbf{The guideline test of each disease.} The disease-specific test categories used in the trajectory analyses and their source guidelines: MIMIC-IV-Ext Clinical Decision Making (MIMIC-CDM) for its 4 diseases\cite{hager2024cdm}, Medical Intelligence for Reasoning and Action (MIRA) for the 3 it added\cite{ferber2026mira}, and for the 5 added here a guideline of the clinical-guidelines corpus from the École polytechnique fédérale de Lausanne (EPFL)\cite{chen2023meditron} or ESCMID. Examination, blood count and inflammatory markers were excluded from these mappings; for cholecystitis, CT was excluded. The bowel-obstruction source addresses a virgin abdomen and the meningitis source addresses bacterial disease; these documents define the operational mappings rather than universal diagnostic requirements. 
}\label{tab:supp-guideline-tests}\\
\end{longtable}
\endgroup

\begingroup
\normalsize
\setlength{\tabcolsep}{2pt}
\renewcommand{\arraystretch}{1.0}
\setlength{\LTcapwidth}{\linewidth}
\setlength{\LTleft}{0pt}
\setlength{\LTright}{0pt plus 1fil}
\begin{longtable}{@{\extracolsep{\fill}}L{\dimexpr0.225\linewidth-12\tabcolsep\relax}C{0.075\linewidth}C{0.13\linewidth}C{0.15\linewidth}C{0.10\linewidth}C{0.16\linewidth}C{0.16\linewidth}@{}}
\toprule
\textbf{Disease} & \makecell{\textbf{Case-}\\\textbf{runs}} & \makecell{\textbf{Decisive}\\\textbf{step, n (\%)}} & \makecell{\textbf{At a guideline}\\\textbf{test, n (\%)}} & \makecell{\textbf{Received}\\\textbf{(\%)}} & \makecell{\textbf{Change at a}\\\textbf{guideline test}} & \makecell{\textbf{Change at}\\\textbf{other actions}} \\
\midrule
\endfirsthead
\multicolumn{7}{c}{\tablename\ \thetable\ -- continued}\\
\toprule
\textbf{Disease} & \makecell{\textbf{Case-}\\\textbf{runs}} & \makecell{\textbf{Decisive}\\\textbf{step, n (\%)}} & \makecell{\textbf{At a guideline}\\\textbf{test, n (\%)}} & \makecell{\textbf{Received}\\\textbf{(\%)}} & \makecell{\textbf{Change at a}\\\textbf{guideline test}} & \makecell{\textbf{Change at}\\\textbf{other actions}} \\
\midrule
\endhead
\midrule
\multicolumn{7}{r}{Continued on next page}\\
\endfoot
\endlastfoot
Appendicitis & 740 & 179 (24) & 104 (58) & 98 & +0.11 (155) & +0.03 (238) \\
Cholecystitis & 645 & 279 (43) & 171 (61) & 85 & +0.15 (119) & +0.04 (376) \\
Bowel obstruction & 335 & 119 (36) & 114 (96) & 99 & +0.15 (84) & -0.01 (221) \\
Diverticulitis & 270 & 176 (65) & 135 (77) & 96 & +0.43 (53) & +0.04 (105) \\
Pancreatitis & 260 & 102 (39) & 97 (95) & 71 & +0.24 (59) & +0.01 (148) \\
Pneumonia & 105 & 37 (35) & 28 (76) & 79 & +0.12 (19) & +0.00 (123) \\
UTI & 90 & 35 (39) & 35 (100) & 91 & +0.20 (19) & -0.01 (62) \\
Pulmonary embolism & 75 & 45 (60) & 45 (100) & 87 & +0.31 (19) & -0.02 (34) \\
Heart failure & 65 & 5 (8) & 1 (20) & 49 & +0.02 (9) & +0.00 (60) \\
Myocardial infarction & 65 & 21 (32) & 21 (100) & 52 & +0.22 (7) & -0.01 (27) \\
Meningitis & 45 & 5 (11) & 3 (60) & 78 & +0.06 (12) & +0.00 (54) \\
Subarachnoid haemorrhage & 35 & 6 (17) & 6 (100) & 71 & +0.10 (5) & +0.01 (21) \\
\midrule
\noalign{\unskip\nobreak}
All & 2730 & 1009 (37) & 760 (75.3) & 88 & +0.18 (559) & +0.02 (1471) \\*
\bottomrule
\noalign{\unskip\nobreak}
\caption{\textbf{The decisive step and the guideline test by disease.} Decision2Care's 5 runs on the 546 paired admissions (2,730 case-runs). Decisive step: the action with the largest rise in the probability of the true disease when that rise is at least 0.20, as a share of the disease's case-runs; at a guideline test, as a share of the decisive steps. Received: case-runs with at least 1 guideline test of the disease. Change: the mean rise in the probability of the true disease per step, with steps per run in parentheses.}\label{tab:supp-decisive-steps}\\
\end{longtable}
\endgroup

\begingroup
\small
\setlength{\tabcolsep}{2pt}
\renewcommand{\arraystretch}{1.0}
\setlength{\LTcapwidth}{\linewidth}
\setlength{\LTleft}{0pt}
\setlength{\LTright}{0pt plus 1fil}
\begin{longtable}{@{\extracolsep{\fill}}L{0.12\linewidth}L{0.16\linewidth}C{0.165\linewidth}C{0.17\linewidth}C{0.165\linewidth}C{0.17\linewidth}@{}}
\toprule
\textbf{Domain} & \textbf{Disease} & \makecell{\textbf{Decision2Care}\\\textbf{(this work)}} & \makecell{\textbf{MIRA-v2 +}\\\textbf{GPT-OSS-120B}} & \makecell{\textbf{MIRA-v2 +}\\\textbf{GLM-5}} & \makecell{\textbf{MIRA-v2 +}\\\textbf{Claude Sonnet 5}} \\
\midrule
\endfirsthead
\multicolumn{6}{c}{\tablename\ \thetable\ -- continued}\\
\toprule
\textbf{Domain} & \textbf{Disease} & \makecell{\textbf{Decision2Care}\\\textbf{(this work)}} & \makecell{\textbf{MIRA-v2 +}\\\textbf{GPT-OSS-120B}} & \makecell{\textbf{MIRA-v2 +}\\\textbf{GLM-5}} & \makecell{\textbf{MIRA-v2 +}\\\textbf{Claude Sonnet 5}} \\
\midrule
\endhead
\midrule
\multicolumn{6}{r}{Continued on next page}\\
\endfoot
\endlastfoot
Physical examination & \textbf{Overall} & \mbox{96.4 (94.5, 97.7)} & \mbox{74.8 (71.0, 78.3)} & \mbox{100.0 (99.3, 100.0)} & \mbox{100.0 (99.3, 100.0)} \\
Physical examination & Appendicitis & \mbox{98.6 (95.2, 99.6)} & \mbox{90.4 (84.6, 94.2)} & \mbox{100.0 (97.5, 100.0)} & \mbox{100.0 (97.5, 100.0)} \\
Physical examination & Cholecystitis & \mbox{95.8 (90.8, 98.1)} & \mbox{49.9 (41.4, 58.4)} & \mbox{100.0 (97.1, 100.0)} & \mbox{100.0 (97.1, 100.0)} \\
Physical examination & Diverticulitis & \mbox{98.1 (90.2, 99.7)} & \mbox{85.9 (74.3, 92.8)} & \mbox{100.0 (93.4, 100.0)} & \mbox{100.0 (93.4, 100.0)} \\
Physical examination & Pancreatitis & \mbox{90.0 (78.9, 95.6)} & \mbox{51.2 (38.0, 64.2)} & \mbox{100.0 (93.1, 100.0)} & \mbox{100.0 (93.1, 100.0)} \\
Physical examination & Pneumonia & \mbox{100.0 (84.5, 100.0)} & \mbox{81.0 (60.0, 92.3)} & \mbox{100.0 (84.5, 100.0)} & \mbox{100.0 (84.5, 100.0)} \\
Physical examination & Pulmonary embolism & \mbox{93.3 (70.2, 98.8)} & \mbox{89.3 (65.2, 97.4)} & \mbox{100.0 (79.6, 100.0)} & \mbox{100.0 (79.6, 100.0)} \\
Physical examination & UTI & \mbox{100.0 (82.4, 100.0)} & \mbox{73.3 (50.2, 88.2)} & \mbox{100.0 (82.4, 100.0)} & \mbox{100.0 (82.4, 100.0)} \\
Physical examination & Bowel obstruction & \mbox{100.0 (94.6, 100.0)} & \mbox{86.3 (76.0, 92.6)} & \mbox{100.0 (94.6, 100.0)} & \mbox{100.0 (94.6, 100.0)} \\
Physical examination & Heart failure & \mbox{93.8 (68.6, 99.1)} & \mbox{90.8 (64.8, 98.1)} & \mbox{100.0 (77.2, 100.0)} & \mbox{100.0 (77.2, 100.0)} \\
Physical examination & Subarachnoid haemorrhage & \mbox{100.0 (64.6, 100.0)} & \mbox{82.9 (45.9, 96.5)} & \mbox{100.0 (64.6, 100.0)} & \mbox{100.0 (64.6, 100.0)} \\
Physical examination & Myocardial infarction & \mbox{69.2 (42.4, 87.3)} & \mbox{87.7 (61.2, 97.0)} & \mbox{100.0 (77.2, 100.0)} & \mbox{100.0 (77.2, 100.0)} \\
Physical examination & Meningitis & \mbox{100.0 (70.1, 100.0)} & \mbox{77.8 (45.3, 93.7)} & \mbox{100.0 (70.1, 100.0)} & \mbox{100.0 (70.1, 100.0)} \\
\midrule
Blood tests & \textbf{Overall} & \mbox{49.7 (49.0, 50.4)} & \mbox{38.6 (37.9, 39.2)} & \mbox{35.7 (35.0, 36.4)} & \mbox{43.9 (43.2, 44.6)} \\
Blood tests & Appendicitis & \mbox{16.6 (15.5, 17.6)} & \mbox{36.8 (35.4, 38.2)} & \mbox{31.9 (30.6, 33.2)} & \mbox{40.1 (38.8, 41.5)} \\
Blood tests & Cholecystitis & \mbox{69.8 (68.5, 71.0)} & \mbox{39.6 (38.3, 41.0)} & \mbox{39.0 (37.7, 40.4)} & \mbox{46.1 (44.7, 47.5)} \\
Blood tests & Diverticulitis & \mbox{36.2 (34.0, 38.4)} & \mbox{41.2 (39.0, 43.5)} & \mbox{32.5 (30.4, 34.7)} & \mbox{40.5 (38.2, 42.7)} \\
Blood tests & Pancreatitis & \mbox{57.4 (55.3, 59.5)} & \mbox{41.4 (39.3, 43.6)} & \mbox{42.5 (40.4, 44.6)} & \mbox{49.1 (47.0, 51.3)} \\
Blood tests & Pneumonia & \mbox{79.7 (77.0, 82.2)} & \mbox{30.9 (27.9, 33.9)} & \mbox{34.2 (31.2, 37.4)} & \mbox{39.4 (36.2, 42.6)} \\
Blood tests & Pulmonary embolism & \mbox{50.1 (46.1, 54.2)} & \mbox{35.4 (31.6, 39.4)} & \mbox{36.0 (32.2, 40.0)} & \mbox{43.0 (39.0, 47.1)} \\
Blood tests & UTI & \mbox{67.8 (63.8, 71.5)} & \mbox{39.3 (35.3, 43.3)} & \mbox{39.3 (35.4, 43.4)} & \mbox{46.1 (42.0, 50.2)} \\
Blood tests & Bowel obstruction & \mbox{55.3 (53.4, 57.2)} & \mbox{42.8 (40.9, 44.7)} & \mbox{33.8 (32.0, 35.7)} & \mbox{47.9 (45.9, 49.8)} \\
Blood tests & Heart failure & \mbox{61.7 (57.6, 65.6)} & \mbox{33.4 (29.7, 37.5)} & \mbox{34.6 (30.8, 38.7)} & \mbox{44.9 (40.8, 49.0)} \\
Blood tests & Subarachnoid haemorrhage & \mbox{59.7 (53.2, 65.8)} & \mbox{32.1 (26.3, 38.4)} & \mbox{35.7 (29.8, 42.1)} & \mbox{41.3 (35.1, 47.8)} \\
Blood tests & Myocardial infarction & \mbox{45.4 (41.3, 49.6)} & \mbox{32.3 (28.5, 36.4)} & \mbox{33.8 (29.9, 37.8)} & \mbox{43.2 (39.1, 47.4)} \\
Blood tests & Meningitis & \mbox{75.1 (70.5, 79.2)} & \mbox{31.3 (26.9, 36.2)} & \mbox{33.9 (29.3, 38.9)} & \mbox{34.4 (29.8, 39.4)} \\
Blood tests & \textit{Tversky distance per case, mean} & 0.713 & 0.627 & 0.641 & 0.580 \\
\midrule
Microbiology & \textbf{Overall} & \mbox{12.8 (10.5, 15.4)} & \mbox{7.3 (5.6, 9.5)} & \mbox{9.9 (8.0, 12.4)} & \mbox{14.9 (12.5, 17.8)} \\
Microbiology & Appendicitis & \mbox{1.0 (0.2, 4.1)} & \mbox{0.6 (0.1, 3.5)} & \mbox{0.4 (0.0, 3.1)} & \mbox{1.3 (0.4, 4.5)} \\
Microbiology & Cholecystitis & \mbox{5.8 (3.2, 10.5)} & \mbox{3.2 (1.4, 7.1)} & \mbox{1.4 (0.4, 4.6)} & \mbox{6.1 (3.4, 10.8)} \\
Microbiology & Diverticulitis & \mbox{5.6 (2.2, 13.3)} & \mbox{4.3 (1.5, 11.5)} & \mbox{3.7 (1.2, 10.7)} & \mbox{11.5 (6.1, 20.6)} \\
Microbiology & Pancreatitis & \mbox{0.0 (0.0, 8.6)} & \mbox{1.0 (0.1, 10.3)} & \mbox{2.0 (0.3, 11.8)} & \mbox{3.9 (0.9, 14.8)} \\
Microbiology & Pneumonia & \mbox{38.0 (26.7, 50.7)} & \mbox{26.4 (16.9, 38.9)} & \mbox{38.0 (26.7, 50.7)} & \mbox{48.8 (36.5, 61.2)} \\
Microbiology & Pulmonary embolism & \mbox{10.0 (2.8, 30.1)} & \mbox{0.0 (0.0, 16.1)} & \mbox{4.0 (0.6, 22.2)} & \mbox{4.0 (0.6, 22.2)} \\
Microbiology & UTI & \mbox{62.6 (46.8, 76.2)} & \mbox{33.2 (20.4, 49.1)} & \mbox{57.9 (42.2, 72.1)} & \mbox{74.7 (59.1, 85.8)} \\
Microbiology & Bowel obstruction & \mbox{10.0 (4.9, 19.2)} & \mbox{2.3 (0.6, 9.0)} & \mbox{2.6 (0.7, 9.4)} & \mbox{4.0 (1.3, 11.5)} \\
Microbiology & Heart failure & \mbox{16.5 (6.5, 36.2)} & \mbox{13.0 (4.5, 32.1)} & \mbox{14.8 (5.5, 34.2)} & \mbox{26.1 (12.5, 46.5)} \\
Microbiology & Subarachnoid haemorrhage & \mbox{0.0 (0.0, 49.0)} & \mbox{0.0 (0.0, 49.0)} & \mbox{15.0 (1.8, 62.5)} & \mbox{0.0 (0.0, 49.0)} \\
Microbiology & Myocardial infarction & \mbox{0.0 (0.0, 35.4)} & \mbox{5.7 (0.5, 42.3)} & \mbox{14.3 (2.6, 51.3)} & \mbox{14.3 (2.6, 51.3)} \\
Microbiology & Meningitis & \mbox{32.4 (21.1, 46.2)} & \mbox{18.0 (9.8, 30.8)} & \mbox{24.4 (14.6, 37.8)} & \mbox{32.4 (21.1, 46.2)} \\
Microbiology & \textit{Tversky distance per case, mean} & 0.696 & 0.703 & 0.692 & 0.678 \\
\midrule
Radiology & \textbf{Overall} & \mbox{65.9 (62.5, 69.1)} & \mbox{65.5 (62.1, 68.7)} & \mbox{67.4 (64.1, 70.6)} & \mbox{68.2 (64.9, 71.3)} \\
Radiology & Appendicitis & \mbox{89.1 (83.6, 92.9)} & \mbox{87.6 (81.9, 91.7)} & \mbox{86.3 (80.4, 90.6)} & \mbox{87.5 (81.7, 91.6)} \\
Radiology & Cholecystitis & \mbox{76.2 (69.5, 81.9)} & \mbox{70.7 (63.7, 76.9)} & \mbox{73.8 (67.0, 79.7)} & \mbox{76.9 (70.2, 82.4)} \\
Radiology & Diverticulitis & \mbox{90.7 (80.5, 95.8)} & \mbox{92.4 (82.7, 96.9)} & \mbox{92.4 (82.7, 96.9)} & \mbox{93.4 (84.0, 97.5)} \\
Radiology & Pancreatitis & \mbox{49.1 (37.4, 60.9)} & \mbox{69.7 (57.8, 79.4)} & \mbox{64.2 (52.2, 74.7)} & \mbox{73.3 (61.6, 82.5)} \\
Radiology & Pneumonia & \mbox{68.8 (51.4, 82.0)} & \mbox{60.0 (42.9, 75.0)} & \mbox{72.5 (55.3, 84.9)} & \mbox{64.4 (47.1, 78.6)} \\
Radiology & Pulmonary embolism & \mbox{47.9 (32.0, 64.2)} & \mbox{49.7 (33.6, 65.9)} & \mbox{50.9 (34.7, 67.0)} & \mbox{49.7 (33.6, 65.9)} \\
Radiology & UTI & \mbox{18.3 (7.7, 37.7)} & \mbox{34.2 (18.6, 54.1)} & \mbox{30.8 (16.1, 50.8)} & \mbox{26.7 (13.1, 46.6)} \\
Radiology & Bowel obstruction & \mbox{52.2 (44.5, 59.8)} & \mbox{41.1 (33.8, 48.9)} & \mbox{43.6 (36.2, 51.4)} & \mbox{41.6 (34.3, 49.4)} \\
Radiology & Heart failure & \mbox{31.4 (15.8, 52.8)} & \mbox{58.1 (37.4, 76.3)} & \mbox{62.9 (41.8, 80.0)} & \mbox{66.7 (45.4, 82.8)} \\
Radiology & Subarachnoid haemorrhage & \mbox{41.2 (21.6, 64.0)} & \mbox{45.9 (25.2, 68.0)} & \mbox{55.3 (33.0, 75.7)} & \mbox{63.5 (40.2, 81.9)} \\
Radiology & Myocardial infarction & \mbox{23.6 (7.8, 53.1)} & \mbox{45.5 (21.3, 72.0)} & \mbox{90.9 (62.3, 98.4)} & \mbox{69.1 (40.1, 88.2)} \\
Radiology & Meningitis & \mbox{16.2 (5.3, 40.3)} & \mbox{30.0 (13.3, 54.4)} & \mbox{32.5 (15.0, 56.8)} & \mbox{27.5 (11.7, 52.0)} \\
Radiology & \textit{Tversky distance per case, mean} & 0.431 & 0.379 & 0.345 & 0.353 \\*
\bottomrule
\noalign{\unskip\nobreak}
\caption{\textbf{Test selection against the record.} The share of each admission's recorded blood tests, microbiology tests and imaging studies that each system requested on the 546 paired cases, pooled over cases and averaged over 5 runs, and the share of cases in which the physical examination was requested, with Wilson 95\% intervals. Tversky distance per case: $1 - \text{both} / (\text{both} + 2 \times \text{extra} + \text{missed})$, 0 when the requests equal the record.}\label{tab:supp-test-selection}\\
\end{longtable}
\endgroup

\begingroup
\small
\setlength{\tabcolsep}{2pt}
\renewcommand{\arraystretch}{0.95}
\setlength{\LTcapwidth}{\linewidth}
\setlength{\LTleft}{0pt}
\setlength{\LTright}{0pt plus 1fil}
\begin{longtable}{@{\extracolsep{\fill}}L{0.22\linewidth}L{\dimexpr0.40\linewidth-8\tabcolsep\relax}C{0.11\linewidth}C{0.16\linewidth}C{0.11\linewidth}@{}}
\toprule
\textbf{System} & \textbf{Rule} & \makecell{\textbf{Retained,}\\\textbf{n (\%)}} & \makecell{\textbf{Retained}\\\textbf{accuracy (\%)}} & \makecell{\textbf{Retained}\\\textbf{errors / all}} \\
\midrule
\endfirsthead
\multicolumn{5}{c}{\tablename\ \thetable\ -- continued}\\
\toprule
\textbf{System} & \textbf{Rule} & \makecell{\textbf{Retained,}\\\textbf{n (\%)}} & \makecell{\textbf{Retained}\\\textbf{accuracy (\%)}} & \makecell{\textbf{Retained}\\\textbf{errors / all}} \\
\midrule
\endhead
\midrule
\multicolumn{5}{r}{Continued on next page}\\
\endfoot
\endlastfoot
Decision2Care & confidence $\geq$ 0.70 (fig.~\ref{fig:supp-deferral-comparators}B) & 269 (49.3) & 98.5 [96.2, 99.4] & 4 / 71 \\
Decision2Care & confidence $\geq$ 0.70, each of the 5 runs & 269--273 (49.3--50.0) & 97.8--98.5 & 4--6 / 71--77 \\
Decision2Care & confidence $\geq$ 0.80 & 148 (27.1) & 99.3 [96.3, 99.9] & 1 / 71 \\
MIRA-v2 + GPT-OSS-120B & ConsistencyDx $\geq$ 0.90, the published threshold (fig.~\ref{fig:supp-deferral-comparators}, D, F and~H) & 300 (54.9) & 96.0 [93.1, 97.7] & 12 / 69 \\
MIRA-v2 + GPT-OSS-120B & ConsistencyDx $\geq$ 0.95 & 205 (37.5) & 97.6 [94.4, 99.0] & 5 / 69 \\
MIRA-v2 + GLM-5 & ConsistencyDx $\geq$ 0.90, the published threshold (fig.~\ref{fig:supp-deferral-comparators}, D, F and~H) & 343 (62.8) & 98.0 [95.8, 99.0] & 7 / 53 \\
MIRA-v2 + GLM-5 & ConsistencyDx $\geq$ 0.95 & 253 (46.3) & 98.0 [95.5, 99.2] & 5 / 53 \\
MIRA-v2 + Claude Sonnet 5 & ConsistencyDx $\geq$ 0.90, the published threshold (fig.~\ref{fig:supp-deferral-comparators}, D, F and~H) & 322 (59.0) & 98.4 [96.4, 99.3] & 5 / 46 \\
MIRA-v2 + Claude Sonnet 5 & ConsistencyDx $\geq$ 0.95 & 206 (37.7) & 98.5 [95.8, 99.5] & 3 / 46 \\
\midrule
Decision2Care & the 269 highest-confidence cases, Decision2Care's coverage at 0.70 & 269 (49.3) & 98.5 [96.2, 99.4] & 4 / 71 \\
Decision2Care & the highest-confidence 70\% of cases & 382 (70.0) & 95.0 [92.4, 96.8] & 19 / 71 \\
MIRA-v2 + GPT-OSS-120B & the 269 highest-confidence cases, Decision2Care's coverage at 0.70 & 269 (49.3) & 96.7 [93.8, 98.2] & 9 / 69 \\
MIRA-v2 + GPT-OSS-120B & the highest-confidence 70\% of cases & 382 (70.0) & 95.0 [92.4, 96.8] & 19 / 69 \\
MIRA-v2 + GLM-5 & the 269 highest-confidence cases, Decision2Care's coverage at 0.70 & 269 (49.3) & 98.1 [95.7, 99.2] & 5 / 53 \\
MIRA-v2 + GLM-5 & the highest-confidence 70\% of cases & 382 (70.0) & 97.4 [95.2, 98.6] & 10 / 53 \\
MIRA-v2 + Claude Sonnet 5 & the 269 highest-confidence cases, Decision2Care's coverage at 0.70 & 269 (49.3) & 98.9 [96.8, 99.6] & 3 / 46 \\
MIRA-v2 + Claude Sonnet 5 & the highest-confidence 70\% of cases & 382 (70.0) & 98.2 [96.3, 99.1] & 7 / 46 \\
\midrule
Decision2Care & the largest set retained with no error & 69 (12.6) & 100.0 [94.7, 100.0] & 0 / 71 \\
MIRA-v2 + GPT-OSS-120B & the largest set retained with no error & 0 (0.0) & -- & -- / 69 \\
MIRA-v2 + GLM-5 & the largest set retained with no error & 0 (0.0) & -- & -- / 53 \\
MIRA-v2 + Claude Sonnet 5 & the largest set retained with no error & 99 (18.1) & 100.0 [96.3, 100.0] & 0 / 46 \\
Decision2Care & the largest set retained with at most 2 errors & 210 (38.5) & 99.0 [96.6, 99.7] & 2 / 71 \\
MIRA-v2 + GPT-OSS-120B & the largest set retained with at most 2 errors & 106 (19.4) & 98.1 [93.4, 99.5] & 2 / 69 \\
MIRA-v2 + GLM-5 & the largest set retained with at most 2 errors & 0 (0.0) & -- & -- / 53 \\
MIRA-v2 + Claude Sonnet 5 & the largest set retained with at most 2 errors & 181 (33.2) & 98.9 [96.1, 99.7] & 2 / 46 \\
Decision2Care & the largest set retained with at most 6 errors & 292 (53.5) & 97.9 [95.6, 99.1] & 6 / 71 \\
MIRA-v2 + GPT-OSS-120B & the largest set retained with at most 6 errors & 222 (40.7) & 97.3 [94.2, 98.8] & 6 / 69 \\
MIRA-v2 + GLM-5 & the largest set retained with at most 6 errors & 320 (58.6) & 98.1 [96.0, 99.1] & 6 / 53 \\
MIRA-v2 + Claude Sonnet 5 & the largest set retained with at most 6 errors & 368 (67.4) & 98.4 [96.5, 99.3] & 6 / 46 \\
\midrule
Decision2Care & confidence $\geq$ 0.70, appendicitis (148 admissions) & 130 (87.8) & 100.0 [97.1, 100.0] & 0 / 1 \\
Decision2Care & confidence $\geq$ 0.70, cholecystitis (129 admissions) & 59 (45.7) & 94.9 [86.1, 98.3] & 3 / 28 \\
Decision2Care & confidence $\geq$ 0.70, bowel obstruction (67 admissions) & 29 (43.3) & 100.0 [88.3, 100.0] & 0 / 5 \\
Decision2Care & confidence $\geq$ 0.70, diverticulitis (54 admissions) & 19 (35.2) & 100.0 [83.2, 100.0] & 0 / 5 \\
Decision2Care & confidence $\geq$ 0.70, pancreatitis (52 admissions) & 14 (26.9) & 92.9 [68.5, 98.7] & 1 / 4 \\
Decision2Care & confidence $\geq$ 0.70, pneumonia (21 admissions) & 1 (4.8) & 100.0 [20.7, 100.0] & 0 / 11 \\
Decision2Care & confidence $\geq$ 0.70, urinary tract infection (18 admissions) & 1 (5.6) & 100.0 [20.7, 100.0] & 0 / 3 \\
Decision2Care & confidence $\geq$ 0.70, pulmonary embolism (15 admissions) & 5 (33.3) & 100.0 [56.6, 100.0] & 0 / 2 \\
Decision2Care & confidence $\geq$ 0.70, myocardial infarction (13 admissions) & 7 (53.8) & 100.0 [64.6, 100.0] & 0 / 4 \\
Decision2Care & confidence $\geq$ 0.70, heart failure (13 admissions) & 0 (0.0) & -- & 0 / 4 \\
Decision2Care & confidence $\geq$ 0.70, meningitis (9 admissions) & 3 (33.3) & 100.0 [43.8, 100.0] & 0 / 0 \\
Decision2Care & confidence $\geq$ 0.70, subarachnoid haemorrhage (7 admissions) & 1 (14.3) & 100.0 [20.7, 100.0] & 0 / 4 \\*
\bottomrule
\noalign{\unskip\nobreak}
\caption{\textbf{Operating points of the 4 systems on the 546 paired admissions.} Retained: cases not deferred for review. Coverage is their share of all cases; retained accuracy is the share diagnosed correctly, with a Wilson 95\% interval, and retained errors are reported over all errors. Decision2Care: its confidence in its diagnosis, read once after the work-up, baseline run unless the row says otherwise. MIRA-v2: ConsistencyDx, the agreement of the diagnoses of its 5 runs\cite{zhang2026onprem}, a case counted as correct when the majority of its runs are correct. At the published threshold of 0.90, MIRA-v2 retained more cases than Decision2Care did at 0.70 (exact McNemar tests, $p = 0.016$ with GPT-OSS-120B and $p < 0.001$ with GLM-5 and with Claude Sonnet 5). Below the rules, the highest-confidence cases for each system at 2 coverage levels, the largest set each signal retains at a given number of errors, and Decision2Care's retained cases by disease. ConsistencyDx computed on Decision2Care's own 5 runs is at chance (area under the receiver operating characteristic curve [AUROC], 0.51): 451 of the 546 cases had the same diagnosis in all 5 runs.}\label{tab:supp-deferral}\\
\end{longtable}
\endgroup

\begingroup
\normalsize
\setlength{\tabcolsep}{2pt}
\renewcommand{\arraystretch}{1.0}
\setlength{\LTcapwidth}{\linewidth}
\setlength{\LTleft}{0pt}
\setlength{\LTright}{0pt plus 1fil}
\begin{longtable}{@{\extracolsep{\fill}}L{0.18\linewidth}C{0.13\linewidth}C{0.12\linewidth}C{0.16\linewidth}C{0.10\linewidth}C{0.13\linewidth}C{0.13\linewidth}@{}}
\toprule
\textbf{Condition} & \makecell{\textbf{Accuracy,}\\\textbf{all cases (\%)}} & \makecell{\textbf{Coverage}\\\textbf{(\%)}} & \makecell{\textbf{Retained}\\\textbf{accuracy (\%)}} & \makecell{\textbf{Retained}\\\textbf{errors}} & \textbf{AUROC} & \makecell{\textbf{Mean}\\\textbf{confidence}} \\
\midrule
\endfirsthead
\multicolumn{7}{c}{\tablename\ \thetable\ -- continued}\\
\toprule
\textbf{Condition} & \makecell{\textbf{Accuracy,}\\\textbf{all cases (\%)}} & \makecell{\textbf{Coverage}\\\textbf{(\%)}} & \makecell{\textbf{Retained}\\\textbf{accuracy (\%)}} & \makecell{\textbf{Retained}\\\textbf{errors}} & \textbf{AUROC} & \makecell{\textbf{Mean}\\\textbf{confidence}} \\
\midrule
\endhead
\midrule
\multicolumn{7}{r}{Continued on next page}\\
\endfoot
\endlastfoot
As recorded & 86.5 (85.9--87.0) & 49.7 (49.3--50.0) & 98.4 (97.8--98.5) & 4.4 (4.0--6.0) & 0.834 (0.831--0.839) & 0.650 (0.649--0.652) \\
History withheld & 59.4 (58.6--60.3) & 35.1 (34.2--36.4) & 97.5 (97.3--98.0) & 4.8 (4.0--5.0) & 0.910 (0.908--0.912) & 0.538 (0.535--0.542) \\
Decisive results made normal & 22.2 (22.0--22.5) & 4.2 (3.7--4.6) & 84.3 (80.0--87.0) & 3.6 (3.0--5.0) & 0.569 (0.555--0.580) & 0.379 (0.376--0.383) \\*
\bottomrule
\noalign{\unskip\nobreak}
\caption{\textbf{Decision2Care's confidence after work-ups with the history withheld or the decisive results made normal.} Results from 5 runs of each condition on the 546 paired admissions (materials and methods), the mean with the lowest and highest run; as recorded, the 5 runs of the Results. Accuracy over all cases by the strict rule; the share retained at a confidence of at least 0.70 and accuracy among retained cases, with the errors among them; the AUROC of the confidence for a correct diagnosis; and its mean.}\label{tab:supp-stress}\\
\end{longtable}
\endgroup

\begingroup
\footnotesize
\setlength{\tabcolsep}{2pt}
\renewcommand{\arraystretch}{1.0}
\setlength{\LTcapwidth}{\linewidth}
\setlength{\LTleft}{0pt}
\setlength{\LTright}{0pt plus 1fil}
\begin{longtable}{@{\extracolsep{\fill}}L{0.12\linewidth}L{0.15\linewidth}C{0.045\linewidth}C{0.12\linewidth}C{0.128\linewidth}C{0.12\linewidth}C{0.134\linewidth}C{0.12\linewidth}@{}}
\toprule
\textbf{Disease group} & \textbf{Treatment} & \textbf{n} & \textbf{Decision2Care} & \makecell{\textbf{MIRA-v2 +}\\\textbf{GPT-OSS-120B}} & \makecell{\textbf{MIRA-v2 +}\\\textbf{GLM-5}} & \makecell{\textbf{MIRA-v2 +}\\\textbf{Claude Sonnet 5}} & \makecell{\textbf{Documented}\\\textbf{care}} \\
\midrule
\endfirsthead
\multicolumn{8}{c}{\tablename\ \thetable\ -- continued}\\
\toprule
\textbf{Disease group} & \textbf{Treatment} & \textbf{n} & \textbf{Decision2Care} & \makecell{\textbf{MIRA-v2 +}\\\textbf{GPT-OSS-120B}} & \makecell{\textbf{MIRA-v2 +}\\\textbf{GLM-5}} & \makecell{\textbf{MIRA-v2 +}\\\textbf{Claude Sonnet 5}} & \makecell{\textbf{Documented}\\\textbf{care}} \\
\midrule
\endhead
\midrule
\multicolumn{8}{r}{Continued on next page}\\
\endfoot
\endlastfoot
Appendicitis & Antibiotics & 145 & \makecell{99.3\\{}[97.2, 100.0]} & \makecell{89.2\\{}[87.1, 91.3]} & \makecell{93.5\\{}[91.0, 95.9]} & \makecell{100.0\\{}[100.0, 100.0]} & \makecell{84.7***\\{}[79.0, 90.3]} \\
Appendicitis & Supportive care & 145 & \makecell{100.0\\{}[100.0, 100.0]} & \makecell{89.1\\{}[87.0, 91.2]} & \makecell{69.1***\\{}[65.3, 73.1]} & \makecell{90.9\\{}[88.6, 93.1]} & \makecell{100.0\\{}[100.0, 100.0]} \\
Appendicitis & Appendectomy & 134 & \makecell{100.0\\{}[100.0, 100.0]} & \makecell{97.5\\{}[96.0, 98.8]} & \makecell{100.0\\{}[100.0, 100.0]} & \makecell{99.9\\{}[99.6, 100.0]} & -- \\
Cholecystitis & Antibiotics & 91 & \makecell{97.8\\{}[94.5, 100.0]} & \makecell{89.5\\{}[86.2, 92.6]} & \makecell{97.4\\{}[95.1, 99.2]} & \makecell{98.4\\{}[95.9, 100.0]} & \makecell{93.3\\{}[87.8, 97.8]} \\
Cholecystitis & Supportive care & 91 & \makecell{100.0\\{}[100.0, 100.0]} & \makecell{89.2\\{}[85.8, 92.5]} & \makecell{72.8***\\{}[66.8, 78.4]} & \makecell{99.1\\{}[98.0, 100.0]} & \makecell{98.9\\{}[96.6, 100.0]} \\
Cholecystitis & Cholecystectomy & 83 & \makecell{100.0\\{}[100.0, 100.0]} & \makecell{66.5***\\{}[61.2, 71.2]} & \makecell{100.0\\{}[100.0, 100.0]} & \makecell{99.5\\{}[98.8, 100.0]} & -- \\
Diverticulitis & Antibiotics & 49 & \makecell{100.0\\{}[100.0, 100.0]} & \makecell{88.5\\{}[83.9, 92.4]} & \makecell{100.0\\{}[100.0, 100.0]} & \makecell{99.6\\{}[98.8, 100.0]} & \makecell{100.0\\{}[100.0, 100.0]} \\
Diverticulitis & Supportive care & 49 & \makecell{98.0\\{}[93.9, 100.0]} & \makecell{85.0\\{}[80.0, 89.7]} & \makecell{24.9***\\{}[17.9, 32.2]} & \makecell{88.6\\{}[81.6, 94.3]} & \makecell{100.0\\{}[100.0, 100.0]} \\
Diverticulitis & Drainage & 2 & \makecell{50.0\\{}[0.0, 100.0]} & \makecell{30.0\\{}[0.0, 60.0]} & \makecell{20.0\\{}[0.0, 40.0]} & \makecell{40.0\\{}[0.0, 80.0]} & -- \\
Pancreatitis & Supportive care & 47 & \makecell{100.0\\{}[100.0, 100.0]} & \makecell{61.1***\\{}[53.8, 68.0]} & \makecell{54.8***\\{}[47.6, 61.3]} & \makecell{99.6\\{}[98.7, 100.0]} & \makecell{100.0\\{}[100.0, 100.0]} \\
Pancreatitis & ERCP & 22 & \makecell{68.2\\{}[50.0, 86.4]} & \makecell{45.2\\{}[30.0, 59.8]} & \makecell{55.5\\{}[40.9, 70.9]} & \makecell{57.3\\{}[40.9, 73.6]} & -- \\
Pancreatitis & Cholecystectomy & 11 & \makecell{45.5\\{}[18.2, 72.7]} & \makecell{0.0\\{}[0.0, 0.0]} & \makecell{38.2\\{}[18.2, 58.2]} & \makecell{69.1\\{}[50.9, 85.5]} & \makecell{63.6\\{}[36.4, 90.9]} \\
Pancreatitis & Drainage & 1 & \makecell{0.0\\{}[0.0, 0.0]} & \makecell{0.0\\{}[0.0, 0.0]} & \makecell{0.0\\{}[0.0, 0.0]} & \makecell{0.0\\{}[0.0, 0.0]} & -- \\
Bowel obstruction & Supportive care & 60 & \makecell{100.0\\{}[100.0, 100.0]} & \makecell{94.8\\{}[92.2, 97.3]} & \makecell{91.9\\{}[87.3, 96.0]} & \makecell{99.3\\{}[98.3, 100.0]} & \makecell{100.0\\{}[100.0, 100.0]} \\
Bowel obstruction & Operation & 14 & \makecell{57.1\\{}[28.6, 85.7]} & \makecell{38.6\\{}[21.4, 57.1]} & \makecell{64.3\\{}[44.3, 80.0]} & \makecell{77.1\\{}[57.1, 95.7]} & -- \\
Bowel obstruction & Nasogastric decompression & 3 & \makecell{100.0\\{}[100.0, 100.0]} & \makecell{78.3\\{}[60.0, 100.0]} & \makecell{73.3\\{}[40.0, 100.0]} & \makecell{93.3\\{}[80.0, 100.0]} & -- \\
Medical diseases & Antibiotics (UTI 10, pneumonia 3) & 13 & \makecell{100.0\\{}[100.0, 100.0]} & \makecell{86.9\\{}[75.8, 95.4]} & \makecell{100.0\\{}[100.0, 100.0]} & \makecell{100.0\\{}[100.0, 100.0]} & \makecell{100.0\\{}[100.0, 100.0]} \\
Medical diseases & Supportive care (UTI 10, PE 10, pneumonia 3) & 23 & \makecell{87.0\\{}[73.9, 100.0]} & \makecell{66.7\\{}[53.3, 79.6]} & \makecell{35.7**\\{}[22.0, 50.2]} & \makecell{60.9\\{}[43.5, 77.4]} & \makecell{91.3\\{}[78.3, 100.0]} \\
Medical diseases & Anticoagulation (PE 10, MI 8) & 18 & \makecell{88.9\\{}[72.2, 100.0]} & \makecell{83.3\\{}[71.1, 93.3]} & \makecell{72.8\\{}[56.9, 88.1]} & \makecell{95.6\\{}[91.1, 98.9]} & \makecell{77.8\\{}[55.6, 94.4]} \\
Medical diseases & Antiplatelet therapy (MI 8) & 8 & \makecell{87.5\\{}[62.5, 100.0]} & \makecell{87.5\\{}[77.5, 97.5]} & \makecell{100.0\\{}[100.0, 100.0]} & \makecell{100.0\\{}[100.0, 100.0]} & \makecell{87.5\\{}[62.5, 100.0]} \\
Medical diseases & Diuretic (heart failure 7) & 7 & \makecell{100.0\\{}[100.0, 100.0]} & \makecell{87.1\\{}[72.9, 97.1]} & \makecell{100.0\\{}[100.0, 100.0]} & \makecell{100.0\\{}[100.0, 100.0]} & \makecell{100.0\\{}[100.0, 100.0]} \\
Medical diseases & Coronary angiography or PCI (MI 8) & 8 & \makecell{100.0\\{}[100.0, 100.0]} & \makecell{55.0\\{}[35.0, 72.5]} & \makecell{92.5\\{}[85.0, 97.5]} & \makecell{100.0\\{}[100.0, 100.0]} & -- \\
\midrule
\multicolumn{8}{@{}p{0.93\linewidth}@{}}{\textit{All 546 paired admissions, every run, incorrect diagnoses included}} \\*
\midrule
\noalign{\unskip\nobreak}
Appendicitis & Antibiotics & 148 & \makecell{98.6\\{}[96.6, 100.0]} & \makecell{88.5\\{}[86.1, 90.7]} & \makecell{93.0\\{}[90.4, 95.3]} & \makecell{99.5\\{}[98.5, 100.0]} & \makecell{84.4***\\{}[78.2, 90.4]} \\
Appendicitis & Supportive care & 148 & \makecell{100.0\\{}[100.0, 100.0]} & \makecell{88.9\\{}[86.9, 90.8]} & \makecell{68.8***\\{}[65.1, 72.6]} & \makecell{90.8\\{}[88.5, 93.0]} & \makecell{100.0\\{}[100.0, 100.0]} \\
Appendicitis & Appendectomy & 137 & \makecell{99.3\\{}[97.8, 100.0]} & \makecell{95.5\\{}[92.7, 97.8]} & \makecell{98.4\\{}[96.4, 99.9]} & \makecell{99.0\\{}[97.2, 100.0]} & -- \\
Cholecystitis & Antibiotics & 129 & \makecell{80.6\\{}[73.6, 86.8]} & \makecell{74.9\\{}[69.3, 80.0]} & \makecell{83.6\\{}[78.1, 88.5]} & \makecell{91.9***\\{}[87.9, 95.2]} & \makecell{86.7\\{}[80.6, 92.2]} \\
Cholecystitis & Supportive care & 129 & \makecell{100.0\\{}[100.0, 100.0]} & \makecell{86.8\\{}[83.6, 89.8]} & \makecell{75.3***\\{}[70.5, 80.0]} & \makecell{98.0\\{}[96.0, 99.5]} & \makecell{99.2\\{}[97.6, 100.0]} \\
Cholecystitis & Cholecystectomy & 119 & \makecell{89.1\\{}[83.2, 94.1]} & \makecell{53.1***\\{}[47.4, 59.0]} & \makecell{92.6\\{}[88.7, 96.0]} & \makecell{96.8*\\{}[94.3, 98.7]} & -- \\
Diverticulitis & Antibiotics & 54 & \makecell{92.6\\{}[85.2, 98.1]} & \makecell{87.0\\{}[81.5, 91.9]} & \makecell{98.1\\{}[94.4, 100.0]} & \makecell{97.4\\{}[93.0, 100.0]} & \makecell{100.0\\{}[100.0, 100.0]} \\
Diverticulitis & Supportive care & 54 & \makecell{98.1\\{}[94.4, 100.0]} & \makecell{83.3\\{}[78.1, 88.2]} & \makecell{27.0***\\{}[20.0, 34.8]} & \makecell{88.9\\{}[82.6, 94.1]} & \makecell{100.0\\{}[100.0, 100.0]} \\
Diverticulitis & Drainage & 2 & \makecell{50.0\\{}[0.0, 100.0]} & \makecell{30.0\\{}[0.0, 60.0]} & \makecell{20.0\\{}[0.0, 40.0]} & \makecell{40.0\\{}[0.0, 80.0]} & -- \\
Pancreatitis & Supportive care & 52 & \makecell{96.2\\{}[90.4, 100.0]} & \makecell{57.3***\\{}[49.6, 64.6]} & \makecell{51.5***\\{}[44.6, 58.5]} & \makecell{98.5\\{}[96.5, 100.0]} & \makecell{98.1\\{}[94.2, 100.0]} \\
Pancreatitis & ERCP & 23 & \makecell{69.6\\{}[52.1, 87.0]} & \makecell{47.0\\{}[32.2, 60.9]} & \makecell{55.7\\{}[40.9, 69.6]} & \makecell{58.3\\{}[41.7, 72.2]} & -- \\
Pancreatitis & Cholecystectomy & 12 & \makecell{50.0\\{}[25.0, 75.0]} & \makecell{0.0\\{}[0.0, 0.0]} & \makecell{38.3\\{}[21.7, 56.7]} & \makecell{71.7\\{}[53.3, 86.7]} & \makecell{58.3\\{}[33.3, 83.3]} \\
Pancreatitis & Drainage & 1 & \makecell{0.0\\{}[0.0, 0.0]} & \makecell{0.0\\{}[0.0, 0.0]} & \makecell{0.0\\{}[0.0, 0.0]} & \makecell{0.0\\{}[0.0, 0.0]} & -- \\
Bowel obstruction & Supportive care & 67 & \makecell{100.0\\{}[100.0, 100.0]} & \makecell{91.0\\{}[86.9, 94.6]} & \makecell{87.2\\{}[81.2, 92.2]} & \makecell{97.9\\{}[95.5, 99.7]} & \makecell{100.0\\{}[100.0, 100.0]} \\
Bowel obstruction & Operation & 15 & \makecell{60.0\\{}[33.3, 80.0]} & \makecell{37.3\\{}[21.3, 54.7]} & \makecell{62.7\\{}[45.3, 78.7]} & \makecell{74.7\\{}[56.0, 92.0]} & -- \\
Bowel obstruction & Nasogastric decompression & 3 & \makecell{100.0\\{}[100.0, 100.0]} & \makecell{73.3\\{}[60.0, 100.0]} & \makecell{73.3\\{}[40.0, 100.0]} & \makecell{93.3\\{}[80.0, 100.0]} & -- \\
Medical diseases & Antibiotics (pneumonia 21, UTI 18) & 39 & \makecell{92.3\\{}[82.1, 100.0]} & \makecell{68.2\\{}[57.4, 77.9]} & \makecell{84.6\\{}[75.9, 92.3]} & \makecell{87.2\\{}[77.9, 94.9]} & \makecell{94.9\\{}[87.2, 100.0]} \\
Medical diseases & Supportive care (pneumonia 21, UTI 18, PE 15) & 54 & \makecell{79.6\\{}[68.5, 88.9]} & \makecell{64.1\\{}[55.2, 73.0]} & \makecell{34.8***\\{}[24.8, 45.2]} & \makecell{67.4\\{}[56.7, 77.8]} & \makecell{86.8\\{}[77.4, 94.4]} \\
Medical diseases & Anticoagulation (PE 15, MI 13) & 28 & \makecell{82.1\\{}[67.9, 96.4]} & \makecell{65.7\\{}[51.4, 79.3]} & \makecell{64.3\\{}[50.7, 77.9]} & \makecell{75.0\\{}[59.3, 87.9]} & \makecell{85.2\\{}[71.4, 96.4]} \\
Medical diseases & Antiplatelet therapy (MI 13) & 13 & \makecell{76.9\\{}[53.8, 100.0]} & \makecell{80.0\\{}[66.2, 92.3]} & \makecell{86.2\\{}[66.1, 100.0]} & \makecell{84.6\\{}[61.5, 100.0]} & \makecell{92.3\\{}[76.9, 100.0]} \\
Medical diseases & Diuretic (heart failure 13) & 13 & \makecell{84.6\\{}[61.5, 100.0]} & \makecell{75.4\\{}[55.4, 92.3]} & \makecell{84.6\\{}[61.5, 100.0]} & \makecell{83.1\\{}[61.5, 100.0]} & \makecell{92.3\\{}[76.9, 100.0]} \\
Medical diseases & Coronary angiography or PCI (MI 11) & 11 & \makecell{81.8\\{}[54.5, 100.0]} & \makecell{41.8\\{}[23.6, 60.0]} & \makecell{81.8\\{}[63.6, 94.5]} & \makecell{87.3\\{}[69.1, 100.0]} & -- \\
Medical diseases & Nimodipine (SAH 1) & 1 & \makecell{100.0\\{}[100.0, 100.0]} & \makecell{20.0\\{}[20.0, 20.0]} & \makecell{20.0\\{}[20.0, 20.0]} & \makecell{40.0\\{}[40.0, 40.0]} & \makecell{0.0\\{}[0.0, 0.0]} \\*
\bottomrule
\noalign{\unskip\nobreak}
\caption{\textbf{Treatments ordered when indicated.} Per disease group and treatment, the number of admissions whose rubric indicated it (materials and methods) and the share in which each system ordered it, a model's runs averaged per admission, with 95\% bootstrap intervals: above the rule, the 437 of the 546 paired admissions that every system diagnosed correctly, correctly diagnosed runs only (fig.~\ref{fig:supp-treatment-comparison}A); below it, all 546 admissions and every run. Documented care: the first day's prescriptions and the emergency department's dispensing, for the drug items and for cholecystectomy in biliary pancreatitis. Stars: a model, or the documented care, against Decision2Care by an exact McNemar test, Benjamini--Hochberg corrected; * $q < 0.05$, ** $q < 0.01$, *** $q < 0.001$. 
}\label{tab:supp-treatment-indicated}\\
\end{longtable}
\endgroup

\begingroup
\footnotesize
\setlength{\tabcolsep}{2pt}
\renewcommand{\arraystretch}{1.0}
\setlength{\LTcapwidth}{\linewidth}
\setlength{\LTleft}{0pt}
\setlength{\LTright}{0pt plus 1fil}
\begin{longtable}{@{\extracolsep{\fill}}L{0.13\linewidth}L{0.13\linewidth}C{0.045\linewidth}C{0.12\linewidth}C{0.128\linewidth}C{0.12\linewidth}C{0.134\linewidth}C{0.12\linewidth}@{}}
\toprule
\textbf{Disease} & \textbf{Category} & \textbf{n} & \textbf{Decision2Care} & \makecell{\textbf{MIRA-v2 +}\\\textbf{GPT-OSS-120B}} & \makecell{\textbf{MIRA-v2 +}\\\textbf{GLM-5}} & \makecell{\textbf{MIRA-v2 +}\\\textbf{Claude Sonnet 5}} & \makecell{\textbf{Documented}\\\textbf{care}} \\
\midrule
\endfirsthead
\multicolumn{8}{c}{\tablename\ \thetable\ -- continued}\\
\toprule
\textbf{Disease} & \textbf{Category} & \textbf{n} & \textbf{Decision2Care} & \makecell{\textbf{MIRA-v2 +}\\\textbf{GPT-OSS-120B}} & \makecell{\textbf{MIRA-v2 +}\\\textbf{GLM-5}} & \makecell{\textbf{MIRA-v2 +}\\\textbf{Claude Sonnet 5}} & \makecell{\textbf{Documented}\\\textbf{care}} \\
\midrule
\endhead
\midrule
\multicolumn{8}{r}{Continued on next page}\\
\endfoot
\endlastfoot
\textbf{All, per patient} & mean over the categories & 355 & \makecell{88.2\\{}[86.0, 90.3]} & \makecell{65.6***\\{}[63.9, 67.3]} & \makecell{58.7***\\{}[57.0, 60.4]} & \makecell{79.7***\\{}[77.6, 81.7]} & \makecell{81.4***\\{}[78.8, 83.9]} \\
Urinary tract infection & antibiotic & 10 & \makecell{70.0\\{}[40.0, 100.0]} & \makecell{69.0\\{}[47.0, 89.0]} & \makecell{90.0\\{}[70.0, 100.0]} & \makecell{80.0\\{}[58.0, 100.0]} & \makecell{90.0\\{}[70.0, 100.0]} \\
Cholecystitis & analgesic & 91 & \makecell{100.0\\{}[100.0, 100.0]} & \makecell{89.2\\{}[85.9, 92.5]} & \makecell{70.0***\\{}[64.1, 75.6]} & \makecell{98.7\\{}[97.4, 99.8]} & \makecell{98.9\\{}[96.6, 100.0]} \\
Cholecystitis & antibiotic & 91 & \makecell{91.2\\{}[84.6, 96.7]} & \makecell{94.6*\\{}[92.6, 96.5]} & \makecell{91.8\\{}[88.2, 95.2]} & \makecell{98.1\\{}[95.4, 100.0]} & \makecell{93.3\\{}[87.8, 97.8]} \\
Cholecystitis & intravenous fluids & 91 & \makecell{100.0\\{}[100.0, 100.0]} & \makecell{13.7***\\{}[10.1, 17.6]} & \makecell{4.5***\\{}[2.3, 7.0]} & \makecell{71.8***\\{}[65.7, 77.7]} & \makecell{97.8\\{}[94.4, 100.0]} \\
Diverticulitis & analgesic & 49 & \makecell{28.6\\{}[16.3, 40.8]} & \makecell{7.8**\\{}[2.4, 13.9]} & \makecell{7.8***\\{}[4.1, 12.2]} & \makecell{5.3**\\{}[1.2, 10.2]} & \makecell{36.7\\{}[24.5, 51.0]} \\
Diverticulitis & antibiotic & 49 & \makecell{98.0\\{}[93.9, 100.0]} & \makecell{91.7\\{}[87.2, 95.1]} & \makecell{84.9\\{}[78.4, 90.6]} & \makecell{94.3\\{}[90.6, 97.6]} & \makecell{98.0\\{}[93.9, 100.0]} \\
Pancreatitis & analgesic & 47 & \makecell{89.4\\{}[80.9, 97.9]} & \makecell{80.8\\{}[75.4, 86.1]} & \makecell{68.4\\{}[60.7, 75.3]} & \makecell{91.4\\{}[86.2, 95.7]} & \makecell{85.1\\{}[74.5, 93.6]} \\
Pancreatitis & antibiotic & 47 & \makecell{100.0\\{}[100.0, 100.0]} & \makecell{98.7\\{}[97.0, 100.0]} & \makecell{98.7\\{}[97.0, 100.0]} & \makecell{92.3\\{}[87.2, 96.6]} & \makecell{66.0***\\{}[51.1, 78.7]} \\
Pancreatitis & intravenous fluids & 47 & \makecell{97.9\\{}[93.6, 100.0]} & \makecell{59.9***\\{}[52.4, 66.8]} & \makecell{56.1***\\{}[48.8, 62.9]} & \makecell{98.3\\{}[96.6, 99.6]} & \makecell{83.0\\{}[72.3, 93.6]} \\
Pancreatitis & nutrition & 47 & \makecell{8.5\\{}[2.1, 17.0]} & \makecell{13.2\\{}[8.5, 18.3]} & \makecell{5.5\\{}[2.6, 9.4]} & \makecell{6.0\\{}[3.0, 9.4]} & \makecell{12.8\\{}[4.3, 23.4]} \\
Pulmonary embolism & analgesic & 10 & \makecell{30.0\\{}[0.0, 60.0]} & \makecell{18.0\\{}[2.0, 36.0]} & \makecell{10.0\\{}[0.0, 22.0]} & \makecell{2.0\\{}[0.0, 6.0]} & \makecell{60.0\\{}[30.0, 90.0]} \\
Pulmonary embolism & anticoagulant & 10 & \makecell{100.0\\{}[100.0, 100.0]} & \makecell{90.0\\{}[78.0, 100.0]} & \makecell{100.0\\{}[100.0, 100.0]} & \makecell{98.0\\{}[94.0, 100.0]} & \makecell{100.0\\{}[100.0, 100.0]} \\
Pneumonia & analgesic & 3 & \makecell{66.7\\{}[0.0, 100.0]} & \makecell{76.7\\{}[50.0, 100.0]} & \makecell{20.0\\{}[0.0, 40.0]} & \makecell{53.3\\{}[20.0, 100.0]} & \makecell{66.7\\{}[0.0, 100.0]} \\
Pneumonia & antibiotic & 3 & \makecell{100.0\\{}[100.0, 100.0]} & \makecell{80.6\\{}[66.7, 100.0]} & \makecell{100.0\\{}[100.0, 100.0]} & \makecell{80.0\\{}[40.0, 100.0]} & \makecell{66.7\\{}[0.0, 100.0]} \\
Appendicitis & analgesic & 145 & \makecell{100.0\\{}[100.0, 100.0]} & \makecell{88.9\\{}[86.9, 91.0]} & \makecell{66.9***\\{}[63.1, 71.0]} & \makecell{88.2\\{}[85.4, 90.9]} & \makecell{100.0\\{}[100.0, 100.0]} \\
Appendicitis & antibiotic & 145 & \makecell{97.2\\{}[94.5, 99.3]} & \makecell{54.8***\\{}[50.3, 59.4]} & \makecell{60.1***\\{}[56.1, 64.3]} & \makecell{88.6\\{}[84.8, 91.9]} & \makecell{66.0***\\{}[57.7, 73.8]} \\*
\bottomrule
\noalign{\unskip\nobreak}
\caption{\textbf{Guideline adherence of the drug orders by disease and category.} The share of admissions whose orders the evaluator of Ferber et~al.\cite{ferber2026mira} judged adherent (materials and methods), on the 355 of the 437 admissions every system diagnosed correctly that have one of its 7 diseases, a model's correctly diagnosed runs averaged per admission, with 95\% bootstrap intervals (fig.~\ref{fig:supp-treatment-comparison}B); the first row averages each patient's categories. Documented care: the first day's prescriptions and the emergency department's dispensing, judged the same way, on 353 admissions. Stars: against Decision2Care, an exact McNemar test for the categories, Benjamini--Hochberg corrected (* $q < 0.05$, ** $q < 0.01$, *** $q < 0.001$), and a Wilcoxon signed-rank test for the first row, uncorrected (the same thresholds for $p$).}\label{tab:supp-guideline-categories}\\
\end{longtable}
\endgroup

\begingroup
\footnotesize
\setlength{\tabcolsep}{2pt}
\renewcommand{\arraystretch}{1.0}
\setlength{\LTcapwidth}{\linewidth}
\setlength{\LTleft}{0pt}
\setlength{\LTright}{0pt plus 1fil}
\begin{longtable}{@{\extracolsep{\fill}}L{0.35\linewidth}C{0.145\linewidth}C{0.145\linewidth}C{0.145\linewidth}C{0.145\linewidth}@{}}
\toprule
\textbf{Measure} & \textbf{Decision2Care} & \makecell{\textbf{MIRA-v2 +}\\\textbf{GPT-OSS-120B}} & \makecell{\textbf{MIRA-v2 +}\\\textbf{GLM-5}} & \makecell{\textbf{MIRA-v2 +}\\\textbf{Claude Sonnet 5}} \\
\midrule
\endfirsthead
\multicolumn{5}{c}{\tablename\ \thetable\ -- continued}\\
\toprule
\textbf{Measure} & \textbf{Decision2Care} & \makecell{\textbf{MIRA-v2 +}\\\textbf{GPT-OSS-120B}} & \makecell{\textbf{MIRA-v2 +}\\\textbf{GLM-5}} & \makecell{\textbf{MIRA-v2 +}\\\textbf{Claude Sonnet 5}} \\
\midrule
\endhead
\midrule
\multicolumn{5}{r}{Continued on next page}\\
\endfoot
\endlastfoot
\multicolumn{5}{@{}p{0.93\linewidth}@{}}{\textit{Drug classes against those given on the first day or in the emergency department (435 admissions with care recorded; mean per admission)}} \\*
\midrule
\noalign{\unskip\nobreak}
F1 & \makecell{40.3\\{}[39.1, 41.6]} & \makecell{34.7***\\{}[33.4, 36.1]} & \makecell{35.2***\\{}[33.8, 36.7]} & \makecell{40.2\\{}[39.0, 41.5]} \\
Precision: classes ordered that were given (\%) & \makecell{46.8\\{}[45.2, 48.4]} & \makecell{49.5**\\{}[47.6, 51.5]} & \makecell{57.0***\\{}[54.7, 59.4]} & \makecell{54.2***\\{}[52.3, 56.1]} \\
Recall: classes given that were ordered (\%) & \makecell{37.2\\{}[35.9, 38.5]} & \makecell{30.1***\\{}[28.8, 31.3]} & \makecell{27.4***\\{}[26.2, 28.7]} & \makecell{33.6***\\{}[32.4, 34.8]} \\
Classes ordered per admission (given: 8.9) & \makecell{6.90\\{}[6.70, 7.11]} & \makecell{5.61***\\{}[5.37, 5.85]} & \makecell{4.37***\\{}[4.15, 4.59]} & \makecell{5.56***\\{}[5.35, 5.78]} \\
Antibiotic decision as given (\%) & \makecell{81.8\\{}[78.0, 85.3]} & \makecell{79.3\\{}[76.4, 82.2]} & \makecell{83.5*\\{}[80.3, 86.6]} & \makecell{84.7\\{}[81.6, 88.0]} \\
F1, the 2 heparins as 1 class & \makecell{50.7\\{}[49.3, 52.1]} & \makecell{36.9***\\{}[35.6, 38.3]} & \makecell{35.4***\\{}[33.9, 36.9]} & \makecell{40.3***\\{}[39.0, 41.6]} \\
Precision, the 2 heparins as 1 class (\%) & \makecell{58.7\\{}[57.0, 60.5]} & \makecell{52.4***\\{}[50.4, 54.4]} & \makecell{57.2\\{}[54.9, 59.5]} & \makecell{54.2***\\{}[52.4, 56.2]} \\
Recall, the 2 heparins as 1 class (\%) & \makecell{46.9\\{}[45.3, 48.5]} & \makecell{32.0***\\{}[30.8, 33.3]} & \makecell{27.6***\\{}[26.3, 28.9]} & \makecell{33.6***\\{}[32.5, 34.9]} \\
\midrule
\multicolumn{5}{@{}p{0.93\linewidth}@{}}{\textit{Anticoagulation against the orders of the first day (437 admissions)}} \\*
\midrule
\noalign{\unskip\nobreak}
Therapeutic anticoagulation, sensitivity (given in 20) (\%) & \makecell{65.0\\{}[45.0, 85.0]} & \makecell{74.5\\{}[59.5, 88.2]} & \makecell{76.2\\{}[60.2, 91.2]} & \makecell{86.0\\{}[74.0, 97.0]} \\
Therapeutic anticoagulation, specificity (not given in 417) (\%) & \makecell{99.0\\{}[98.1, 99.8]} & \makecell{98.9\\{}[97.9, 99.7]} & \makecell{99.2\\{}[98.4, 99.8]} & \makecell{98.8\\{}[97.6, 99.7]} \\
Venous thromboembolism prophylaxis, sensitivity (given in 377) (\%) & \makecell{92.3\\{}[89.7, 95.0]} & \makecell{18.2***\\{}[16.0, 20.6]} & \makecell{0.2***\\{}[0.1, 0.5]} & \makecell{1.2***\\{}[0.6, 1.8]} \\
Venous thromboembolism prophylaxis, specificity (not given in 60) (\%) & \makecell{36.7\\{}[25.0, 48.4]} & \makecell{89.3***\\{}[84.7, 93.4]} & \makecell{97.6***\\{}[94.2, 99.7]} & \makecell{99.3***\\{}[98.0, 100.0]} \\
\midrule
\multicolumn{5}{@{}p{0.93\linewidth}@{}}{\textit{Procedures of the admission (349 coded in 244 of the 437 admissions)}} \\*
\midrule
\noalign{\unskip\nobreak}
Requested, a shared ICD-9-CM code (\%) & \makecell{47.3\\{}[41.4, 53.7]} & \makecell{42.1\\{}[36.2, 48.7]} & \makecell{52.6\\{}[46.6, 59.7]} & \makecell{56.2\\{}[50.3, 62.9]} \\
Requested, an equivalent code (\%) & \makecell{14.3\\{}[10.7, 18.1]} & \makecell{3.8\\{}[2.2, 5.6]} & \makecell{5.6\\{}[3.5, 7.9]} & \makecell{6.2\\{}[4.1, 8.6]} \\
Requested, total (\%) & \makecell{61.6\\{}[55.9, 68.0]} & \makecell{45.9\\{}[40.2, 52.2]} & \makecell{58.2\\{}[52.5, 65.2]} & \makecell{62.4\\{}[56.8, 68.9]} \\
Precision: requests that matched a coded procedure (\%) & \makecell{42.7\\{}[38.2, 47.2]} & \makecell{54.9\\{}[49.4, 60.2]} & \makecell{58.1\\{}[53.0, 63.2]} & \makecell{56.7\\{}[51.9, 61.4]} \\
Requests per admission & \makecell{1.14\\{}[1.05, 1.24]} & \makecell{0.67\\{}[0.63, 0.71]} & \makecell{0.82\\{}[0.77, 0.86]} & \makecell{0.88\\{}[0.83, 0.93]} \\
Tversky distance (0, identical sets) & \makecell{0.54\\{}[0.49, 0.59]} & \makecell{0.60\\{}[0.55, 0.64]} & \makecell{0.52\\{}[0.47, 0.56]} & \makecell{0.50\\{}[0.45, 0.55]} \\
Requested, total, appendicitis (106 procedures in 145 admissions) (\%) & \makecell{92.5\\{}[86.9, 97.2]} & \makecell{89.8\\{}[84.0, 94.7]} & \makecell{92.5\\{}[86.9, 97.2]} & \makecell{92.3\\{}[86.7, 97.1]} \\
Requested, total, cholecystitis (98 procedures in 91 admissions) (\%) & \makecell{79.6\\{}[70.1, 89.8]} & \makecell{52.5\\{}[45.2, 60.7]} & \makecell{79.4\\{}[69.9, 89.5]} & \makecell{79.2\\{}[69.8, 89.4]} \\
Requested, total, pancreatitis (59 procedures in 47 admissions) (\%) & \makecell{18.6\\{}[9.0, 31.3]} & \makecell{1.7\\{}[0.0, 4.7]} & \makecell{15.9\\{}[10.1, 24.1]} & \makecell{34.6\\{}[24.8, 47.4]} \\
Requested, total, bowel obstruction (37 procedures in 60 admissions) (\%) & \makecell{24.3\\{}[11.6, 41.2]} & \makecell{8.5\\{}[1.5, 17.5]} & \makecell{20.5\\{}[12.6, 30.0]} & \makecell{22.7\\{}[13.5, 32.6]} \\
\midrule
\multicolumn{5}{@{}p{0.93\linewidth}@{}}{\textit{Decision2Care's procedure requests at a higher probability for the procedure group (the figures use 0.5)}} \\*
\midrule
\noalign{\unskip\nobreak}
Probability $\geq$ 0.50: requests per admission; requested (\%); precision (\%) & 1.14; 61.6; 42.7 & -- & -- & -- \\
Probability $\geq$ 0.70: requests per admission; requested (\%); precision (\%) & 0.94; 56.2; 47.4 & -- & -- & -- \\
Probability $\geq$ 0.80: requests per admission; requested (\%); precision (\%) & 0.76; 49.9; 51.5 & -- & -- & -- \\
Probability $\geq$ 0.85: requests per admission; requested (\%); precision (\%) & 0.62; 42.4; 54.1 & -- & -- & -- \\
Probability $\geq$ 0.90: requests per admission; requested (\%); precision (\%) & 0.41; 30.4; 58.1 & -- & -- & -- \\*
\bottomrule
\noalign{\unskip\nobreak}
\caption{\textbf{Drug classes, anticoagulation and procedures against the admission's record.} The 437 of the 546 paired admissions that every system diagnosed correctly; a model's correctly diagnosed runs averaged per admission; 95\% bootstrap intervals. Drug classes: a system's orders against the classes given on the first day or in the emergency department (materials and methods), and with the 2 heparins counted as 1 class (fig.~\ref{fig:supp-treatment-comparison}C). Anticoagulation: against the first day's orders at treatment dose and at prophylactic dose. Procedures: the measures of Ferber et~al.\cite{ferber2026mira}, pooled and then by disease; last, Decision2Care's requests as the probability required for a procedure group rises above 0.5. Stars: a model against Decision2Care, Wilcoxon signed-rank test on the per-admission values, the antibiotic decision and anticoagulation by an exact McNemar test, uncorrected; * $p < 0.05$, ** $p < 0.01$, *** $p < 0.001$. F1 is the harmonic mean of precision and recall.}\label{tab:supp-treatment-measures}\\
\end{longtable}
\endgroup

\clearpage

\begingroup
\footnotesize
\setlength{\tabcolsep}{2pt}
\renewcommand{\arraystretch}{1.0}
\setlength{\LTcapwidth}{\linewidth}
\setlength{\LTleft}{0pt}
\setlength{\LTright}{0pt plus 1fil}
\begin{longtable}{@{\extracolsep{\fill}}L{0.19\linewidth}L{0.12\linewidth}C{0.12\linewidth}C{0.12\linewidth}C{0.19\linewidth}C{0.19\linewidth}@{}}
\toprule
\textbf{Disease} & \textbf{Database} & \makecell{\textbf{Patients}\\\textbf{(30/90 d)}} & \makecell{\textbf{Deaths}\\\textbf{(30/90 d)}} & \makecell{\textbf{AUROC, 30 d,}\\\textbf{mean}\\\textbf{(range)}} & \makecell{\textbf{AUROC, 90 d,}\\\textbf{mean}\\\textbf{(range)}} \\
\midrule
\endfirsthead
\multicolumn{6}{c}{\tablename\ \thetable\ -- continued}\\
\toprule
\textbf{Disease} & \textbf{Database} & \makecell{\textbf{Patients}\\\textbf{(30/90 d)}} & \makecell{\textbf{Deaths}\\\textbf{(30/90 d)}} & \makecell{\textbf{AUROC, 30 d,}\\\textbf{mean}\\\textbf{(range)}} & \makecell{\textbf{AUROC, 90 d,}\\\textbf{mean}\\\textbf{(range)}} \\
\midrule
\endhead
\midrule
\multicolumn{6}{r}{Continued on next page}\\
\endfoot
\endlastfoot
Intracerebral haemorrhage & MIMIC-III & 316 / 317 & 131 / 140 & 0.864 (0.862--0.865) & 0.854 (0.851--0.857) \\
Intracerebral haemorrhage & MIMIC-IV & 652 / 653 & 220 / 242 & 0.895 (0.893--0.899) & 0.879 (0.877--0.880) \\
Traumatic brain injury & MIMIC-III & 1,528 & 271 / 319 & 0.871 (0.870--0.873) & 0.855 (0.855--0.856) \\
Traumatic brain injury & MIMIC-IV & 2,481 & 334 / 421 & 0.891 (0.889--0.893) & 0.856 (0.854--0.858) \\
Sepsis & MIMIC-III & 614 / 615 & 164 / 225 & 0.779 (0.774--0.783) & 0.762 (0.759--0.766) \\
Sepsis & MIMIC-IV & 335 / 336 & 81 / 119 & 0.768 (0.757--0.776) & 0.717 (0.709--0.726) \\
Acute kidney injury & MIMIC-III & 122 & 42 / 44 & 0.874 (0.871--0.878) & 0.872 (0.868--0.877) \\
Acute kidney injury & MIMIC-IV & 541 & 198 / 221 & 0.807 (0.804--0.812) & 0.806 (0.803--0.810) \\
Pulmonary embolism & MIMIC-III & 209 & 26 / 52 & 0.779 (0.768--0.786) & 0.831 (0.829--0.835) \\
Pulmonary embolism & MIMIC-IV & 405 / 406 & 21 / 43 & 0.777 (0.772--0.782) & 0.869 (0.866--0.873) \\
ARDS candidates & MIMIC-III & 38 & 8 / 11 & 0.722 (0.713--0.746) & 0.713 (0.699--0.724) \\
ARDS candidates & MIMIC-IV & 66 & 29 / 31 & 0.848 (0.840--0.859) & 0.849 (0.842--0.861) \\
Pneumonia (CAP) & MIMIC-III & 344 & 68 / 86 & 0.737 (0.734--0.739) & 0.747 (0.745--0.749) \\
Pneumonia (CAP) & MIMIC-IV & 770 & 56 / 84 & 0.875 (0.873--0.880) & 0.815 (0.813--0.817) \\
Polytrauma (matched) & MIMIC-III & 72 & 13 / 13 & 0.892 (0.874--0.899) & 0.894 (0.876--0.901) \\
Polytrauma (matched) & MIMIC-IV & 72 & 11 / 11 & 0.736 (0.722--0.747) & 0.742 (0.730--0.750) \\
\midrule
Pooled, 8 cohorts & MIMIC-III and IV & 8,565 / 8,570 & 1,673 / 2,062 & 0.871 (0.870--0.872) & 0.849 (0.848--0.850) \\
\midrule
Pulmonary embolism & INSPECT & 555 / 542 & 41 / 65 & 0.813 (0.803--0.825) & 0.800 (0.794--0.810) \\
Pneumonia candidates & INSPECT & 78 / 76 & 6 / 8 & 0.770 (0.749--0.786) & 0.762 (0.748--0.784) \\*
\bottomrule
\noalign{\unskip\nobreak}
\caption{\textbf{The prediction of death at 8 h after arrival in the emergency department, by cohort.} Results from 5 runs on each fixed cohort; the AUROC of the prediction for all-cause death within 30 and 90 days of the prediction, the mean with the lowest and highest run. Patients and deaths are those with a known outcome at each horizon. The pooled curves of Fig.~\ref{fig:fig6}a,b concatenate the 8 disease cohorts from MIMIC-III and MIMIC-IV within each run. Each disease-task membership is retained when an admission belongs to multiple cohorts (materials and methods). Clinical screening defined the acute respiratory distress syndrome (ARDS) candidate cohorts. CAP, community-acquired pneumonia. The polytrauma cohorts are matched between the 2 databases. The last 2 rows are from INSPECT, a data set of Stanford Health Care (materials and methods; fig.~\ref{fig:supp-second-centre}), and are not part of the pooled curves; all other rows are from Beth Israel Deaconess Medical Center. Its pneumonia group consists of candidates whose community acquisition could not be verified. The 95\% bootstrap intervals over patients at 30 and 90 days were 0.746--0.873 and 0.743--0.851 in pulmonary embolism, and 0.658--0.883 and 0.646--0.871 in the pneumonia candidates.}\label{tab:supp-prognosis-cohorts}\\
\end{longtable}
\endgroup

\begingroup
\footnotesize
\setlength{\tabcolsep}{2pt}
\renewcommand{\arraystretch}{1.0}
\setlength{\LTcapwidth}{\linewidth}
\setlength{\LTleft}{0pt}
\setlength{\LTright}{0pt plus 1fil}
\begin{longtable}{@{\extracolsep{\fill}}L{0.14\linewidth}C{0.07\linewidth}C{0.07\linewidth}C{0.165\linewidth}C{0.09\linewidth}C{0.07\linewidth}C{0.07\linewidth}C{0.165\linewidth}C{0.09\linewidth}@{}}
\toprule
 & \makecell{\textbf{Sens.,}\\\textbf{30 d}} & \makecell{\textbf{Spec.,}\\\textbf{30 d}} & \makecell{\textbf{AUROC, 30 d}\\\textbf{(agent minus}\\\textbf{ICU clinician)}} & \makecell{\textbf{Unsure,}\\\textbf{30 d}} & \makecell{\textbf{Sens.,}\\\textbf{90 d}} & \makecell{\textbf{Spec.,}\\\textbf{90 d}} & \makecell{\textbf{AUROC, 90 d}\\\textbf{(agent minus}\\\textbf{ICU clinician)}} & \makecell{\textbf{Unsure,}\\\textbf{90 d}} \\
\midrule
\endfirsthead
\multicolumn{9}{c}{\tablename\ \thetable\ -- continued}\\
\toprule
 & \makecell{\textbf{Sens.,}\\\textbf{30 d}} & \makecell{\textbf{Spec.,}\\\textbf{30 d}} & \makecell{\textbf{AUROC, 30 d}\\\textbf{(agent minus}\\\textbf{ICU clinician)}} & \makecell{\textbf{Unsure,}\\\textbf{30 d}} & \makecell{\textbf{Sens.,}\\\textbf{90 d}} & \makecell{\textbf{Spec.,}\\\textbf{90 d}} & \makecell{\textbf{AUROC, 90 d}\\\textbf{(agent minus}\\\textbf{ICU clinician)}} & \makecell{\textbf{Unsure,}\\\textbf{90 d}} \\
\midrule
\endhead
\midrule
\multicolumn{9}{r}{Continued on next page}\\
\endfoot
\endlastfoot
ICU clinician 1 & 0.63 & 0.60 & 0.720 (-0.15 to +0.13) & 51 (71\%) & 0.65 & 0.53 & 0.655 (-0.17 to +0.12) & 55 (76\%) \\
ICU clinician 2 & 0.70 & 0.76 & 0.755 (-0.15 to +0.05) & 37 (51\%) & 0.68 & 0.78 & 0.741 (-0.23 to -0.01) & 37 (51\%) \\
ICU clinician 3 & 0.63 & 0.49 & 0.670 (-0.06 to +0.15) & 15 (21\%) & 0.62 & 0.50 & 0.637 (-0.13 to +0.11) & 15 (21\%) \\
Mean of the 3 & 0.65 & 0.61 & 0.727 (-0.12 to +0.08) & 103 (48\%) & 0.65 & 0.60 & 0.696 (-0.19 to +0.05) & 107 (50\%) \\
\midrule
Agent, 1 run & & & 0.707 (0.565--0.836) & & & & 0.628 (0.496--0.757) & \\
\midrule
\multicolumn{9}{@{}l}{\textit{Weighted back to the cohorts}} \\
ICU clinician 1 & & & 0.927 (-0.13 to +0.06) & & & & 0.867 (-0.17 to +0.11) & \\
ICU clinician 2 & & & 0.933 (-0.12 to +0.03) & & & & 0.888 (-0.14 to +0.03) & \\
ICU clinician 3 & & & 0.875 (-0.04 to +0.11) & & & & 0.822 (-0.08 to +0.11) & \\
Mean of the 3 & & & 0.923 (-0.11 to +0.05) & & & & 0.883 (-0.15 to +0.05) & \\
Agent, 1 run & & & 0.903 & & & & 0.844 & \\*
\bottomrule
\noalign{\unskip\nobreak}
\caption{\textbf{Three ICU clinicians' predictions and the agent's on the same 72 cases.} The 72 review cases of Fig.~\ref{fig:fig6}a,b, 7--10 from each of 8 diseases (35 from MIMIC-III and 37 from MIMIC-IV), drawn within disease and database from strata of 90-day mortality crossed with the correctness of the agent's 90-day prediction (21 true positive, 19 false negative, 18 true negative and 14 false positive); 27 and 40 ended in death by 30 and 90 days of the prediction made 8 h after arrival. Wrong agent predictions of 90-day mortality were thus over-represented (33 of 72, 46\%, compared with 2,247 of the 8,570 cohort entries, 26\%), so the AUROCs of the review cases are lower than those of the cohorts (0.871 and 0.849). The ICU clinicians were given every record offered to the agent (a median of 124 per case), of which the agent had read a median of 54\%. Each ICU clinician predicted the interval of death (within 7 days, days 8--30, days 31--90, alive at 90 days) and the probability of death by 30 and by 90 days and rated their certainty (1, sure, to 4, unsure) before seeing the agent's work-up (materials and methods); sensitivity (Sens.) and specificity (Spec.) are those of the interval prediction for death by each horizon, the crosses of Fig.~\ref{fig:fig6}a,b; the AUROC is that of the probability, with the 95\% interval of the agent's AUROC minus the ICU clinician's from 2,000 paired case-bootstrap resamples; unsure, the predictions rated rather unsure or unsure. The agent's AUROC comes from 1 run on the same cases, with its 95\% bootstrap interval. In the weighted rows each case is weighted by the number of cohort entries in its stratum divided by the number of review cases drawn from that stratum (effective sample size 14.6; 77 cohort entries lay in 2 strata without a review case), and the intervals come from 2,000 case-bootstrap resamples with the weights fixed (materials and methods). Of the 33 cases in which the agent's 90-day prediction was wrong, the interval predictions of at least 2 of the 3 ICU clinicians were also wrong in 20 and those of all 3 in 10.}\label{tab:supp-readers}\\
\end{longtable}
\endgroup

\begingroup
\footnotesize
\setlength{\tabcolsep}{2pt}
\renewcommand{\arraystretch}{1.0}
\setlength{\LTcapwidth}{\linewidth}
\setlength{\LTleft}{0pt}
\setlength{\LTright}{0pt plus 1fil}
\begin{longtable}{@{\extracolsep{\fill}}L{0.21\linewidth}C{0.07\linewidth}C{0.09\linewidth}C{0.11\linewidth}C{0.10\linewidth}C{0.09\linewidth}C{0.10\linewidth}C{0.09\linewidth}@{}}
\toprule
 & \textbf{Reviews} & \makecell{\textbf{Essential}\\\textbf{record}\\\textbf{unread}} & \makecell{\textbf{Step}\\\textbf{unnecessary}} & \makecell{\textbf{Prediction}\\\textbf{at least}\\\textbf{defensible}} & \makecell{\textbf{Prediction}\\\textbf{plausible}} & \makecell{\textbf{Plausibility}\\\textbf{(mean)}} & \makecell{\textbf{Interval}\\\textbf{agreed}} \\
\midrule
\endfirsthead
\multicolumn{8}{c}{\tablename\ \thetable\ -- continued}\\
\toprule
 & \textbf{Reviews} & \makecell{\textbf{Essential}\\\textbf{record}\\\textbf{unread}} & \makecell{\textbf{Step}\\\textbf{unnecessary}} & \makecell{\textbf{Prediction}\\\textbf{at least}\\\textbf{defensible}} & \makecell{\textbf{Prediction}\\\textbf{plausible}} & \makecell{\textbf{Plausibility}\\\textbf{(mean)}} & \makecell{\textbf{Interval}\\\textbf{agreed}} \\
\midrule
\endhead
\midrule
\multicolumn{8}{r}{Continued on next page}\\
\endfoot
\endlastfoot
ICU clinician 1 & 72 & 5 (7\%) & 55 (76\%) & 67 (93\%) & 44 (61\%) & 3.90 & 50 (69\%) \\
ICU clinician 2 & 72 & 12 (17\%) & 27 (38\%) & 50 (69\%) & 30 (42\%) & 3.21 & 40 (56\%) \\
ICU clinician 3 & 72 & 3 (4\%) & 40 (56\%) & 65 (90\%) & 51 (71\%) & 4.10 & 53 (74\%) \\
\midrule
Intracerebral haemorrhage & 30 & 0 (0\%) & 16 (53\%) & 27 (90\%) & 19 (63\%) & 3.97 & 21 (70\%) \\
Traumatic brain injury & 30 & 0 (0\%) & 8 (27\%) & 25 (83\%) & 18 (60\%) & 3.87 & 18 (60\%) \\
Sepsis & 30 & 8 (27\%) & 16 (53\%) & 26 (87\%) & 15 (50\%) & 3.60 & 18 (60\%) \\
Acute kidney injury & 21 & 7 (33\%) & 9 (43\%) & 15 (71\%) & 11 (52\%) & 3.24 & 14 (67\%) \\
Pulmonary embolism & 21 & 1 (5\%) & 21 (100\%) & 18 (86\%) & 9 (43\%) & 3.57 & 13 (62\%) \\
ARDS candidates & 30 & 2 (7\%) & 11 (37\%) & 23 (77\%) & 16 (53\%) & 3.57 & 18 (60\%) \\
Pneumonia (CAP) & 24 & 2 (8\%) & 21 (88\%) & 21 (88\%) & 14 (58\%) & 3.88 & 17 (71\%) \\
Polytrauma (matched) & 30 & 0 (0\%) & 20 (67\%) & 27 (90\%) & 23 (77\%) & 4.03 & 24 (80\%) \\
\midrule
All & 216 & 20 (9\%) & 122 (56\%) & 182 (84\%) & 125 (58\%) & 3.74 & 143 (66\%) \\*
\bottomrule
\noalign{\unskip\nobreak}
\caption{\textbf{The ICU clinicians' review of the agent's work-ups and predictions.} After making their own prediction for each of the 72 cases, the 3 ICU clinicians were shown the records the agent retrieved, in order, and its prediction, and recorded whether an essential record had been left unread, whether a step had been unnecessary, how plausible the prediction was on a scale of 1 to 5 and whether they agreed with the interval it predicted (materials and methods). On the scale, 1 was labelled `implausible given the data', 3 `defensible, I would differ' and 5 `fully consistent with the data'; at least defensible, a rating of 3 or more; plausible, a rating of 4 or 5. Counts of reviews by ICU clinician (72 each), by disease (21 to 30) and in all (216).}\label{tab:supp-readers-review}\\
\end{longtable}
\endgroup

\begingroup
\footnotesize
\setlength{\tabcolsep}{2pt}
\renewcommand{\arraystretch}{1.0}
\setlength{\LTcapwidth}{\linewidth}
\setlength{\LTleft}{0pt}
\setlength{\LTright}{0pt plus 1fil}
\begin{longtable}{@{\extracolsep{\fill}}L{0.13\linewidth}C{0.08\linewidth}C{0.05\linewidth}C{0.05\linewidth}C{0.05\linewidth}C{0.05\linewidth}C{0.05\linewidth}C{0.05\linewidth}C{0.05\linewidth}C{0.05\linewidth}C{0.05\linewidth}C{0.05\linewidth}C{0.05\linewidth}C{0.05\linewidth}C{0.06\linewidth}@{}}
\toprule
\textbf{Disease} & \makecell{\textbf{Patients}\\\textbf{(deaths)}} & \textbf{1 h} & \textbf{2 h} & \textbf{3 h} & \textbf{4 h} & \textbf{5 h} & \textbf{6 h} & \textbf{7 h} & \textbf{8 h} & \textbf{9 h} & \textbf{10 h} & \textbf{11 h} & \textbf{12 h} & \makecell{\textbf{First}\\\textbf{hour}} \\
\midrule
\endfirsthead
\multicolumn{15}{c}{\tablename\ \thetable\ -- continued}\\
\toprule
\textbf{Disease} & \makecell{\textbf{Patients}\\\textbf{(deaths)}} & \textbf{1 h} & \textbf{2 h} & \textbf{3 h} & \textbf{4 h} & \textbf{5 h} & \textbf{6 h} & \textbf{7 h} & \textbf{8 h} & \textbf{9 h} & \textbf{10 h} & \textbf{11 h} & \textbf{12 h} & \makecell{\textbf{First}\\\textbf{hour}} \\
\midrule
\endhead
\midrule
\multicolumn{15}{r}{Continued on next page}\\
\endfoot
\endlastfoot
Intracerebral haemorrhage & 774 (196) & 0.678 & 0.787 & 0.825 & 0.839 & 0.845 & 0.854 & 0.842 & 0.862 & 0.863 & 0.871 & 0.875 & 0.874 & 2 \\
Traumatic brain injury & 2,790 (358) & 0.724 & 0.790 & 0.819 & 0.835 & 0.847 & 0.854 & 0.857 & 0.865 & 0.868 & 0.866 & 0.871 & 0.872 & 2 \\
Sepsis & 851 (170) & 0.611 & 0.659 & 0.698 & 0.688 & 0.698 & 0.720 & 0.701 & 0.724 & 0.726 & 0.744 & 0.734 & 0.750 & none \\
Acute kidney injury & 531 (159) & 0.692 & 0.766 & 0.763 & 0.737 & 0.736 & 0.738 & 0.742 & 0.735 & 0.702 & 0.711 & 0.733 & 0.721 & 2 \\
Pulmonary embolism & 326 (29) & 0.702 & 0.707 & 0.751 & 0.710 & 0.784 & 0.814 & 0.796 & 0.820 & 0.814 & 0.818 & 0.829 & 0.801 & none \\
ARDS candidates & 92 (24) & 0.624 & 0.704 & 0.722 & 0.706 & 0.748 & 0.742 & 0.760 & 0.723 & 0.724 & 0.723 & 0.762 & 0.737 & 3 \\
Pneumonia (CAP) & 493 (69) & 0.669 & 0.695 & 0.702 & 0.700 & 0.694 & 0.716 & 0.740 & 0.730 & 0.734 & 0.772 & 0.765 & 0.752 & none \\
Polytrauma (matched) & 142 (21) & 0.551 & 0.646 & 0.689 & 0.667 & 0.681 & 0.702 & 0.689 & 0.671 & 0.766 & 0.758 & 0.762 & 0.783 & 6 \\*
\bottomrule
\noalign{\unskip\nobreak}
\caption{\textbf{The prediction of death in hospital from the records of the first 1 to 12 h in the intensive care unit (ICU).} AUROC by disease, patients from MIMIC-III and MIMIC-IV pooled, each hour an independent prediction from the records of the first $h$ hours after admission to the ICU, in patients alive and in hospital at 24 h (materials and methods); 95\% bootstrap intervals are drawn in Fig.~\ref{fig:fig6}c and given for 12 h in table~\ref{tab:supp-icu-scores}. First hour, the earliest hour at which the AUROC reached the highest of the 3 24-h scores (a descriptive mark, not a test).}\label{tab:supp-icu-hourly}\\
\end{longtable}
\endgroup

\begingroup
\small
\setlength{\tabcolsep}{2pt}
\renewcommand{\arraystretch}{1.0}
\setlength{\LTcapwidth}{\linewidth}
\setlength{\LTleft}{0pt}
\setlength{\LTright}{0pt plus 1fil}
\begin{longtable}{@{\extracolsep{\fill}}L{0.12\linewidth}L{0.07\linewidth}C{0.15\linewidth}C{0.095\linewidth}C{0.18\linewidth}C{0.15\linewidth}C{0.165\linewidth}@{}}
\toprule
\textbf{Disease} & \textbf{Score} & \makecell{\textbf{Score at 24 h,}\\\textbf{AUROC}\\\textbf{(95\% CI)}} & \makecell{\textbf{Prediction}\\\textbf{at 2 h}} & \makecell{\textbf{Difference at 2 h}\\\textbf{(95\% CI)}} & \makecell{\textbf{Prediction}\\\textbf{at 12 h}\\\textbf{(95\% CI)}} & \makecell{\textbf{Difference at 12 h}\\\textbf{(95\% CI)}} \\
\midrule
\endfirsthead
\multicolumn{7}{c}{\tablename\ \thetable\ -- continued}\\
\toprule
\textbf{Disease} & \textbf{Score} & \makecell{\textbf{Score at 24 h,}\\\textbf{AUROC}\\\textbf{(95\% CI)}} & \makecell{\textbf{Prediction}\\\textbf{at 2 h}} & \makecell{\textbf{Difference at 2 h}\\\textbf{(95\% CI)}} & \makecell{\textbf{Prediction}\\\textbf{at 12 h}\\\textbf{(95\% CI)}} & \makecell{\textbf{Difference at 12 h}\\\textbf{(95\% CI)}} \\
\midrule
\endhead
\midrule
\multicolumn{7}{r}{Continued on next page}\\
\endfoot
\endlastfoot
Intracerebral haemorrhage & APS III & 0.631 (0.582--0.679) & 0.787 & +0.156 (+0.097, +0.214) & 0.874 (0.846--0.899) & +0.244 (+0.190, +0.296) \\
 & SAPS II & 0.713 (0.672--0.751) &  & +0.074 (+0.022, +0.127) &  & +0.161 (+0.120, +0.203) \\
 & OASIS & 0.728 (0.687--0.766) &  & +0.059 (+0.011, +0.109) &  & +0.146 (+0.105, +0.188) \\
Traumatic brain injury & APS III & 0.701 (0.670--0.733) & 0.790 & +0.089 (+0.053, +0.126) & 0.872 (0.854--0.891) & +0.171 (+0.139, +0.205) \\
 & SAPS II & 0.756 (0.730--0.781) &  & +0.034 (+0.004, +0.066) &  & +0.116 (+0.087, +0.147) \\
 & OASIS & 0.763 (0.738--0.789) &  & +0.027 (-0.001, +0.058) &  & +0.109 (+0.084, +0.135) \\
Sepsis & APS III & 0.754 (0.714--0.791) & 0.659 & -0.095 (-0.143, -0.046) & 0.750 (0.707--0.789) & -0.004 (-0.044, +0.037) \\
 & SAPS II & 0.750 (0.713--0.786) &  & -0.092 (-0.137, -0.043) &  & -0.001 (-0.042, +0.043) \\
 & OASIS & 0.708 (0.663--0.752) &  & -0.049 (-0.097, +0.000) &  & +0.042 (-0.005, +0.089) \\
Acute kidney injury & APS III & 0.722 (0.674--0.769) & 0.766 & +0.044 (-0.014, +0.100) & 0.721 (0.677--0.764) & -0.001 (-0.053, +0.055) \\
 & SAPS II & 0.765 (0.721--0.807) &  & +0.001 (-0.052, +0.056) &  & -0.044 (-0.095, +0.009) \\
 & OASIS & 0.743 (0.697--0.786) &  & +0.023 (-0.034, +0.080) &  & -0.022 (-0.080, +0.036) \\
Pulmonary embolism & APS III & 0.842 (0.765--0.914) & 0.707 & -0.135 (-0.216, -0.056) & 0.801 (0.709--0.887) & -0.042 (-0.115, +0.033) \\
 & SAPS II & 0.817 (0.730--0.897) &  & -0.110 (-0.193, -0.025) &  & -0.016 (-0.104, +0.075) \\
 & OASIS & 0.843 (0.769--0.907) &  & -0.136 (-0.223, -0.056) &  & -0.042 (-0.128, +0.043) \\
ARDS candidates & APS III & 0.706 (0.585--0.822) & 0.704 & -0.002 (-0.142, +0.138) & 0.737 (0.618--0.844) & +0.030 (-0.104, +0.165) \\
 & SAPS II & 0.648 (0.521--0.769) &  & +0.056 (-0.117, +0.231) &  & +0.088 (-0.061, +0.247) \\
 & OASIS & 0.650 (0.527--0.772) &  & +0.054 (-0.105, +0.215) &  & +0.086 (-0.071, +0.240) \\
Pneumonia (CAP) & APS III & 0.748 (0.688--0.804) & 0.695 & -0.054 (-0.123, +0.018) & 0.752 (0.691--0.811) & +0.004 (-0.058, +0.066) \\
 & SAPS II & 0.801 (0.745--0.857) &  & -0.107 (-0.174, -0.042) &  & -0.049 (-0.107, +0.009) \\
 & OASIS & 0.736 (0.668--0.798) &  & -0.042 (-0.114, +0.030) &  & +0.016 (-0.054, +0.089) \\
Polytrauma (matched) & APS III & 0.700 (0.581--0.802) & 0.646 & -0.054 (-0.183, +0.071) & 0.783 (0.676--0.883) & +0.083 (-0.024, +0.194) \\
 & SAPS II & 0.640 (0.510--0.754) &  & +0.006 (-0.168, +0.188) &  & +0.143 (-0.002, +0.299) \\
 & OASIS & 0.689 (0.574--0.794) &  & -0.044 (-0.174, +0.082) &  & +0.093 (-0.009, +0.200) \\*
\bottomrule
\noalign{\unskip\nobreak}
\caption{\textbf{The ICU prediction against APS III, SAPS II and OASIS.} The AUROC of each score computed from the first 24 h in the same patients as Fig.~\ref{fig:fig6}c, the AUROC of the prediction with 2 and 12 h of records, and the paired difference, prediction minus score, with 95\% intervals from 2,000 patient bootstrap replicates stratified by database and outcome and shared by all hours and scores (materials and methods). APS III, Acute Physiology Score III. CI, confidence interval. OASIS, Oxford Acute Severity of Illness Score. SAPS II, Simplified Acute Physiology Score II.}\label{tab:supp-icu-scores}\\
\end{longtable}
\endgroup

\clearpage

\endgroup

\supplementend

\end{document}